\documentclass[10pt,twocolumn,letterpaper]{article}

\usepackage[pagenumbers]{cvpr} 

\ifdefined\pdfminorversion
\fi

\newcommand{\bench}{\textit{FireGround}}
\newcommand{\agent}{FireAgents}

\usepackage{array}

\newcommand{\firegroundtitle}{\bench{}: Benchmarking Embodied Decision-Making in Dynamic Residential Fires}

\DeclareMathSizes{8.2}{8.2}{6}{5}
\DeclareMathSizes{8.3}{8.3}{6}{5}

\definecolor{cvprblue}{rgb}{0.21,0.49,0.74}
\usepackage[pagebackref,breaklinks,colorlinks,allcolors=cvprblue]{hyperref}

\def\confName{CVPR}
\def\confYear{2027}

\title{\firegroundtitle}

\author{%
  Weiqin~Zu$^{1,2}$\thanks{These authors contributed equally.}\quad
  Yujie~Lu$^{2}$\footnotemark[1]\quad
  Miao~Li$^{2}$\quad
  Yifan~Chen$^{4,2}$\quad
  Huiying~Li$^{2}$\quad
  Yizhe~Huang$^{3,2}$\\
  Kaiyang~Cai$^{5,2}$\quad
  Hualong~Yan$^{6}$\quad
  Ruihuan~Zhang$^{6}$\quad
  Dengji~Zhao$^{1}$\thanks{Corresponding authors.}\quad
  Zhenliang~Zhang$^{2}$\footnotemark[2]\quad
  Xue~Feng\footnotemark[2]\\[3pt]
  {\small $^{1}$ShanghaiTech University \quad
          $^{2}$State Key Laboratory of General Artificial Intelligence, BIGAI}\\
  {\small $^{3}$Peking University \quad
          $^{4}$Tsinghua University \quad
          $^{5}$University of Science and Technology of China \quad
          $^{6}$China Fire and Rescue Institute}\\
  {\small\ttfamily \{zuwq,zhaodj\}@shanghaitech.edu.cn \quad
                   zlzhang@bigai.ai \quad feng.xue1580@gmail.com}
}

\begin{document}
\maketitle
\begin{abstract}
Dynamic embodied environments such as residential fire scenarios challenge agents to interpret evolving visual observations and act promptly as hazards evolve. We introduce \bench{}, an Unreal Engine~5 benchmark coupling local visual perception, fire suppression, and civilian evacuation in 3D residential environments. Smoke and illumination changes affect observations, suppression alters fire progression, and civilian health deteriorates with exposure, requiring agents to balance information gathering with timely intervention. Its five tasks vary observation scope, rescue objectives, and robot teams, with metrics tailored to these challenges. We also introduce \agent{}, a standardized agent architecture supporting symbolic and visual observations under single-robot and centralized or decentralized multi-robot control. Evaluations of proprietary and open-source foundation models reveal lower performance under visual observations in several settings and difficulties translating broader exploration into better outcomes. Benefits from perception assistance and additional robots depend on task objectives and control structure. Together, \bench{} and \agent{} support systematic evaluation of perception, reasoning, and action in dynamic embodied environments. The project website is \href{https://zuwq.github.io/FireGround/}{FireGround}. \end{abstract}

\section{Introduction}

Dynamic embodied environments continually change what agents can observe, which tasks require immediate attention, and when actions can be effective. Agents must interpret incomplete and evolving observations, anticipate environmental changes, and assess the consequences of their actions to decide where and when to intervene. Residential fire response provides a demanding setting for studying these capabilities, as fires spread and civilian conditions worsen while agents perceive, deliberate, and act~\cite{queralta2020collaborative,saputra2025rescue}.

Residential fire response presents three coupled challenges. Limited viewpoints and indoor occlusions restrict available information, while smoke, illumination changes, and combustion continually alter visual observations. Acquiring and updating observations takes time, requiring agents to balance the value of new information against the costs of delayed intervention and the risks of acting on outdated observations. Suppression and rescue are also interdependent: extinguishing a nearby fire can reduce civilian exposure but delay evacuation, requiring agents to jointly consider suppression priorities, rescue timing, and evacuation routes.

Existing benchmarks address aspects of these challenges. HAZARD studies object evacuation in evolving disasters~\cite{zhou2024hazard}, CREW-Wildfire combines wildfire suppression with civilian rescue~\cite{hyun2025crew}, and FireNav examines cooperative navigation in indoor fires~\cite{anonymousfirenav}. Building on these directions, we investigate how foundation models use evolving local visual information to coordinate active suppression and evacuation while accounting for civilian health.

\begin{figure*}[t]
  \centering
  \resizebox{\textwidth}{!}{\includegraphics{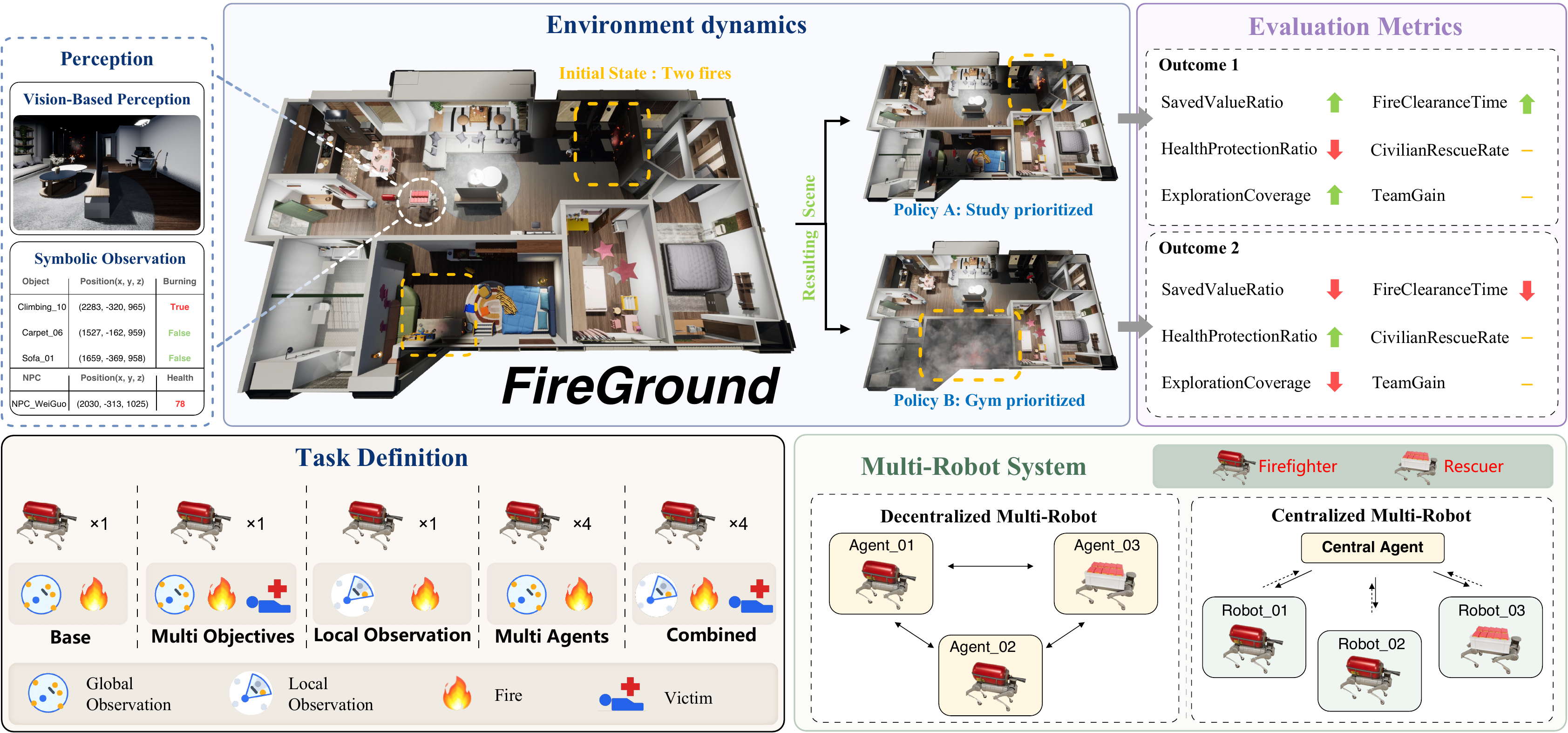}}  
\caption{Overview of \bench{}.
Starting from the same initial fire scenario, different response priorities lead to distinct fire trajectories and outcomes, illustrating trade-offs between property preservation and human safety.
Visual or symbolic observations provide decision inputs, while task configurations define objectives, observation scope, and team size.
Heterogeneous firefighting and rescue robots translate decisions into actions through decentralized or centralized coordination, thereby shaping the evolving environment.
Multidimensional evaluation assesses the resulting outcomes, linking response priorities and coordination choices to performance across task objectives.}

\label{fig:ema_bench_overview}
\end{figure*}

We introduce \bench{}, a benchmark built in Unreal Engine~5 featuring 3D residential environments with spreading fires and first-person visual observations affected by smoke and lighting changes (Fig.~\ref{fig:ema_bench_overview}). Suppression alters subsequent fire progression, while exposure-dependent health loss connects rescue timing and evacuation routes to civilian outcomes. We propose five tasks that systematically vary observation scope, rescue objectives, and robot teams under shared fire dynamics. Complementary symbolic observations support comparisons across information conditions, and evaluation metrics are tailored to the dynamics and decision-making challenges of residential fire response.

We further introduce \agent{}, a standardized agent architecture for evaluating widely used foundation models across observation interfaces and single-robot, centralized multi-robot, and decentralized multi-robot control. Our experiments reveal performance declines under visual observations relative to symbolic inputs in several task settings, while greater exploration coverage does not consistently translate into better intervention outcomes. These findings suggest that models struggle to balance information acquisition with timely action under evolving hazards. Benefits from perception assistance and additional robots also depend on task objectives and control structure.

Our contributions are:
\begin{itemize}
    \item \bench{}, a residential fire benchmark coupling dynamic visual perception with active suppression and civilian rescue, supported by evaluation metrics tailored to these challenges.
    \item \agent{}, a standardized architecture for evaluating foundation models across observation interfaces and robot control structures.
    \item An empirical evaluation revealing challenges in translating visual information, exploration, and team resources into timely and effective intervention.
\end{itemize}

\section{Related Work}

\begin{table*}[t]
    \centering
    \footnotesize
    \setlength{\tabcolsep}{1.15pt}
    \renewcommand{\arraystretch}{1.08}
    \newcommand{\benchYes}{\ensuremath{\checkmark}}
    \newcommand{\benchNo}{\ensuremath{\times}}
    \newcommand{\benchUnspecified}{\textendash}
    \caption{Comparison with representative embodied and hazard-response benchmarks. Intervention Coupling means robot actions alter subsequent hazard evolution.}
    \label{tab:fireground-benchmarks}
    \label{tab:sim_compa}
    \begin{tabular*}{\textwidth}{
        @{\extracolsep{\fill}}lccccc
        >{\raggedright\arraybackslash}p{0.15\textwidth}
        @{\hspace{4pt}}>{\raggedright\arraybackslash}p{0.14\textwidth}@{}
    }
        \toprule
        & \multicolumn{2}{c}{Hazard Dynamics}
        & \multicolumn{5}{c}{Evaluation Setup} \\
        \cmidrule(lr){2-3}\cmidrule(lr){4-8}
        Benchmark
        & \shortstack{Cumulative \&\\Irreversible Loss}
        & \shortstack{Intervention\\Coupling}
        & \shortstack{Inference-Time\\Progression}
        & \shortstack{Active\\Exploration}
        & \shortstack{Multi-Robot\\Tasks}
        & \shortstack{Input\\Modality} & Evaluated Tasks \\
        \midrule
        \multicolumn{8}{@{}l}{\textit{General embodied benchmarks}} \\
        MAP-THOR~\cite{nayak2024mapthor}
        & \benchUnspecified & \benchUnspecified & \benchUnspecified & \benchYes
        & \benchYes & Visual + Symbolic & Household tasks \\
        PARTNR~\cite{chang2025partnr}
        & \benchUnspecified & \benchUnspecified & \benchUnspecified & \benchYes
        & \benchYes & Symbolic & Household tasks \\

        \midrule
        \multicolumn{8}{@{}l}{\textit{Dynamic and hazard-response benchmarks}} \\
        FD-Bench~\cite{zhu2025fdbench}
        & \benchUnspecified & \benchNo & \benchNo & \benchNo
        & \benchNo & Not reported & Escape planning \\
        HAZARD~\cite{zhou2024hazard}
        & \benchYes & \benchNo & \benchUnspecified & \benchYes
        & \benchNo & Symbolic & Object rescue \\
        CREW-Wildfire~\cite{hyun2025crew}
        & \benchYes & \benchYes & \benchNo & \benchYes
        & \benchYes & Symbolic & Suppression, rescue \\
        FireNav~\cite{anonymousfirenav}
        & \benchYes & \benchNo & \benchUnspecified & \benchYes
        & \benchYes & Visual + Symbolic & Target search \\
        RescueBench~\cite{wu2026rescuebench}
        & \benchUnspecified & \benchUnspecified & \benchUnspecified & \benchYes
        & \benchYes & Visual + Symbolic & Search and rescue \\
        RT-Safe~\cite{xu2026rtsafe}
        & \benchUnspecified & \benchYes & \benchYes & \benchNo
        & \benchNo & Visual + Symbolic & Safe navigation \\
        ROAR~\cite{wang2026roar}
        & \benchYes & \benchYes & \benchYes & \benchUnspecified
        & \benchYes & Visual + Symbolic & Collection, combat \\

        \midrule
        \textbf{\bench{} (Ours)}
        & \benchYes & \benchYes & \benchYes & \benchYes
        & \benchYes & \textbf{Visual + Symbolic} & \textbf{Suppression, rescue} \\
        \bottomrule
    \end{tabular*}
    \par\smallskip
    \parbox{\textwidth}{\raggedright
        \benchYes{} Yes. \benchNo{} No. \benchUnspecified{} Not reported.}
\end{table*}

\noindent\textbf{Real-Time Decision-Making in Dynamic Environments.}
An environment can evolve without agent intervention. Inference deadlines constrain model selection in embodied control~\cite{song2024modec}. Real-time evaluation also accounts for how the environment changes during deliberation~\cite{ramstedt2019realtime,kang2025winfast}. Recent benchmarks compare live and paused interaction in games and other digital tasks~\cite{tang2026realtimebench,pan2026liveclock}. In embodied settings, RT-Safe and LIBERO-Dynamic account for inference time in urban navigation and moving-object manipulation, respectively, while ROAR and GPTNT evaluate coordination under time pressure in multiplayer games~\cite{xu2026rtsafe,ren2026liberodynamic,wang2026roar,parekh2026gptnt}. In FireGround, delayed intervention can change which objects ignite before a fire is suppressed.

\noindent\textbf{Embodied Multi-Agent Decision-Making.}
Prior work on embodied cooperation has examined tasks such as joint object lifting from egocentric observations~\cite{jain2019twobody}. Foundation models have been applied to embodied navigation, multi-robot dispatch, team adaptation, and collaborative planning~\cite{zu2024language,chen2024dispatch,li2025learn,zu2025cots}. Interactive 3D platforms support a range of embodied tasks~\cite{kolve2017ai2,puig2018virtualhome,wang2024grutopia,zhou2025virtual}, including long-horizon instruction following in ALFRED~\cite{shridhar2020alfred}. MAP-THOR evaluates long-horizon planning under partial observability, Habitat-MAS and RoCoBench study task allocation and execution across robots, and VIKI-Bench tests visually grounded coordination~\cite{nayak2024mapthor,chen2024emos,mandi2024roco,kang2025viki}. PARTNR, CoEmbod, and CoCoBench provide tasks with explicit dependencies such as spatial allocation, ordering, and synchronized execution~\cite{chang2025partnr,dong2026coembod,chen2026cocobench}. MECoBench compares team structures and communication, while MAGMA-Bench and SimWorld-Robotics examine execution disruptions and navigation in dynamic urban environments, respectively~\cite{liu2026mecobench,bernat2026magma,zhuang2025simworldrobotics}. In fire response, a team's assignments can change as fires spread or teammates contain them.

\noindent\textbf{Embodied Environments for Hazard Response.}
FD-Bench evaluates evacuation reasoning in grid-based fires, while FireWorldBench uses multimodal questions to assess predictions of fire evolution and intervention outcomes~\cite{zhu2025fdbench,chen2026fireworldbench}. HAZARD studies object evacuation during evolving disasters; RescueBench and ESARBench focus on embodied search and rescue~\cite{zhou2024hazard,wu2026rescuebench,zhang2026esarbench}. VULCAN and FireNav study cooperative navigation in indoor fires~\cite{liu2026vulcan,anonymousfirenav}. RoboCup Rescue, RMASBench, and FireCommander offer earlier settings for coordinated suppression, while CREW-Wildfire evaluates large heterogeneous teams suppressing wildfires and rescuing civilians~\cite{skinner2010robocup,maffioletti2013rmasbench,seraj2020firecommander,hyun2025crew}. FireGround models object-level fire propagation in residential interiors, starting with single-agent suppression under global observations and examining the individual and combined effects of local observations, rescue objectives, and multi-agent coordination on intervention decisions under shared fire dynamics. Table~\ref{tab:fireground-benchmarks} compares representative benchmarks.


\section{\bench{} Benchmark}
\bench{} evaluates embodied decision-making under fire dynamics that continue during deliberation and action. Five tasks share these dynamics, varying rescue objectives, observation scope, and team size to examine their individual and combined effects on intervention outcomes.

\subsection{Environment and Dynamics}
\label{sec:environment}

\noindent\textbf{Scene and Episode Generation.}
Built in Unreal Engine 5, \bench{} constructs residential scenes from
modular room templates that preserve each room's function while varying
its furnishings and layout (Appendix~\ref{appendix:assets}).
Templates can be selected explicitly or sampled with a fixed random seed.
Initial fire locations and counts, as well as combustion, spread, and
suppression parameters, are configurable.
A configurable preburn period allows fires to develop before robots
intervene, setting the initial conditions for evaluation.
We provide approximately 2,000 predefined configurations covering
different room layouts, fire conditions, and civilian exposure, together
with their simulated fire trajectories\footnote{\url{https://huggingface.co/datasets/ZWQ1103/FireGround}}.
Configuration and data details are provided in the supplementary material (Appendices~\ref{appendix:scene_generation} and~\ref{appendix:dataset_release}).

\noindent\textbf{Object-Level Fire Dynamics.}
\bench{} uses an object-level phenomenological model of fuel consumption,
fire spread, and suppression~\cite{nuessle2005approaching,peacock2015cfast}.
Each combustible object $j$ has a remaining fuel reserve $Q_{j,t}\geq 0$
and a fire intensity $I_{j,t}\geq 0$.
When its intensity first exceeds $\theta_j^{\mathrm{prop}}$, the object
ignites unburned combustible objects within a size-dependent neighborhood.
While burning, its state evolves as
\begin{align}
    Q_{j,t+\Delta t}
    &=
    \max\left(0,\,
        Q_{j,t}-\beta_j\Delta t
    \right),
    \label{eq:fuel_update}
    \\
    I_{j,t+\Delta t}
    &=
    \max\left(0,\,
        I_{j,t}+g_j\Delta t
        -\eta_j N_{j,t}^{\mathrm{sup}}\Delta t
    \right),
    \label{eq:fire_update}
\end{align}
where $\beta_j$ and $g_j$ are time-invariant, object-specific fuel-consumption and fire-growth rates, respectively. This parameterization enables controlled evaluation of agent decisions under evolving fire hazards.
The total suppression rate is $\eta_j N_{j,t}^{\mathrm{sup}}$, where $\eta_j$ is the per-robot rate and $N_{j,t}^{\mathrm{sup}}$ is the number of robots suppressing object $j$ simultaneously, with additive contributions.

A burning object becomes permanently inactive when its fuel is exhausted
or its intensity reaches zero.
Fuel consumption causes irreversible damage, while suppression before
the propagation threshold can prevent ignition of nearby objects.
Combustion also changes the scene's appearance~\cite{zhou2024hazard}.
Fire and smoke affect visibility, changes in lighting alter visual cues, and burning objects darken, shrink, and may disappear.
These effects are reflected in agents' rendered visual observations.
Parameters, state transitions, and visual effects are detailed in
Appendix~\ref{appendix:fire_model}.

\noindent\textbf{Civilian Exposure and Protection.}
Motivated by cumulative exposure models in fire hazard
assessment~\cite{hurley2008fire}, \bench{} represents civilian health
by a phenomenological variable $H_t\in[0,100]$.
Let $d_t^{\min}$ be the distance to the nearest burning object and
$\tau_t^{\mathrm{exp}}$ the exposure time accumulated within radius
$R_{\mathrm{exp}}$ of a burning object.
During exposure, health is updated as
\begin{equation}
    H_{t+\Delta t}
    =
    \max\left(
        0,\,
        H_t-\lambda_H\mu_t^{\mathrm{mask}}
        \frac{\tau_{t+\Delta t}^{\mathrm{exp}}}{d_t^{\min}}
    \right),
    \label{eq:health_update}
\end{equation}
where $\lambda_H$ is a per-update coefficient for the fixed simulation
interval, and $\mu_t^{\mathrm{mask}}$ equals $1$ without a mask and lies
in $(0,1)$ with a mask.
Longer exposure and shorter distances therefore cause greater health loss.
Suppression and escort can end exposure, while masks reduce the damage
accumulated during it.
Health remains constant outside the exposure radius.
Parameters and civilian state transitions are detailed in
Appendix~\ref{appendix:victim_model}.

\subsection{Agent--Environment Interaction}

\noindent\textbf{Observation Interfaces.}
\bench{} provides three observation interfaces.
\emph{Global Symbolic} returns simulator-generated records of
task-relevant objects and civilians throughout the scene, including
their identities, locations, burning status, and health where applicable.
\emph{Local Symbolic} restricts these records to entities observable
from the robot's first-person viewpoint.
\emph{Local Visual} provides first-person RGB images and aligned
depth maps in place of structured entity records.
Agents identify objects and civilians and infer states such as
whether an object is burning from these observations.
For actions specified by image coordinates, aligned depth supports
mapping the selected locations to 3D targets
(Appendix~\ref{appendix:observation_interfaces}).

\noindent\textbf{Actions and Robot Roles.}
\bench{} supports two types of quadrupedal robots.
Only \emph{firefighting robots} (FRs) can perform \emph{fire suppression},
targeting a selected burning object from their current position.
\emph{Rescue robots} (RRs) automatically equip nearby civilians
with protective masks to reduce subsequent health loss.

Both robot types support the following high-level actions.
\emph{Navigation} moves the robot either to a target object or a specified distance in a direction relative to its current heading.
\emph{Room exploration} follows a planned route through a selected room and collects observations along the way.
\emph{Civilian instruction} asks nearby civilians to follow the robot
or releases those already following, allowing either robot type to
escort civilians to a safe zone.
\emph{Waiting} holds the robot in place for a fixed interval while it
awaits task updates or a teammate's intervention, such as suppressing
a fire along an evacuation route.
Definitions and parameters for the high-level actions and low-level
movement controls are provided in Appendices~\ref{appendix:action_space}
and~\ref{appendix:navigation_and_movement}, respectively.

\subsection{Problem Formulation}
\label{sec:problem_formulation}
Let \(x(t)\) denote the state at time \(t\), including object fuel reserves and fire intensities, civilian locations and health, and robot poses and ongoing actions. Robot \(i\) receives observations $o_i(t)=\mathcal{O}_{m,i}(x(t))$ through interface \(m\).
At each decision time, an agent selects an action for robot \(i\) from its role-specific action set \(\mathcal{A}_{r_i}\), where \(r_i\) denotes the robot’s role. Decisions are based on available observations, interaction history, and execution feedback.

Robots execute actions asynchronously, with potentially different start and completion times. Over an interval \([t,t+\Delta t)\), the state evolves as $x(t+\Delta t)
=
\mathcal{F}_{\Delta t}
\!\left(x(t),\mathbf{u}_{[t,t+\Delta t)}\right)$, where \(\mathbf{u}_{[t,t+\Delta t)}\) denotes the actions executed by all robots during that interval. The transition captures both autonomous hazard progression and the effects of robot interventions.
In real-time runs, the environment continues to evolve during deliberation and action execution. Consequently, conditions observed during decision-making may have changed by the time the selected action is executed
(Appendix~\ref{appendix:interaction_timing}).

\subsection{Task Design}
\label{sec:task_design}

\noindent\textbf{Base Task (T1).}
A single firefighting robot determines the order in which to extinguish fires based on \textit{global symbolic} observations.
Because fires continue to spread while the robot travels,
these decisions must account for both travel time and the risk
of further ignition.

\noindent\textbf{Task Variants (T2--T4).}
Each variant changes one aspect of T1.
T2 adds civilian rescue, with civilian safety taking priority.
The robot must decide whether to suppress a nearby fire before
escorting a civilian: suppression can reduce exposure during rescue,
but delays evacuation.
T3 restricts observations to the robot's first-person viewpoint.
Exploring an unvisited room may reveal a more urgent fire,
but delays suppression of fires already found.
T4 extends T1 to a team of firefighting robots.
The team must allocate fires among robots and update assignments
as fires spread or teammates contain them.

\noindent\textbf{Combined Task (T5).}
T5 combines civilian rescue, local observations, and multiple robots.
Finding a civilian near an active fire may require the team
to reassign a robot and coordinate suppression with evacuation,
while other fires continue to develop.
For T3 and T5, local observations are provided through either
the \textit{local symbolic} or \textit{local visual} interface (Appendix~\ref{appendix:task_configurations}).

\subsection{Evaluation Metrics}
\label{Chapter:Bench_Metrics}
\bench{} provides diverse metrics to evaluate the performance of models, including \textbf{Saved Value Ratio (SVR)} and \textbf{Fire Clearance Time (FCT)} for fire suppression, \textbf{Health Protection Ratio (HPR)} and \textbf{Civilian Rescue Rate (CRR)} for civilian rescue, \textbf{Exploration Coverage (EC)} for exploration, and \textbf{Team Gain (TG)} for the improvement in property preservation achieved by a multi-robot team over a matched single-robot baseline. Definitions and details of the metrics are discussed in Appendices~\ref{appendix:evaluation_protocol} and~\ref{appendix:metrics}.

\section{\agent{}: Foundation-Model Baselines}
\label{sec:agent_baselines}
\begin{figure}[tb]
  \centering
  \resizebox{\columnwidth}{!}{\includegraphics{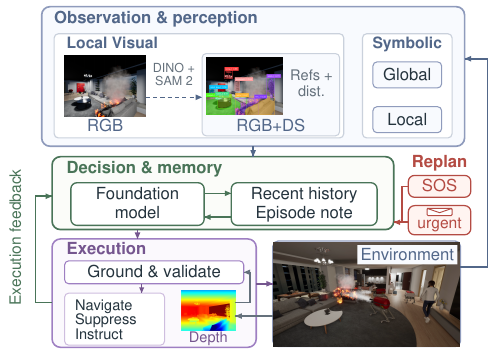}}
\caption{\textbf{Overview of \agent{}.}
Each run uses one fixed configuration: Global Symbolic, Local Symbolic, RGB, or RGB+DS.
 DS augments RGB with DINO/SAM~2 annotations; depth supports internal grounding and distance estimation.
SOS and urgent teammate messages trigger immediate replanning.
In real-time runs, the environment evolves during inference and execution.}
  \label{fig:agent_framework}
\end{figure}

We propose \agent{}, a standardized agent architecture for evaluating foundation models in \bench{} (Fig.~\ref{fig:agent_framework}). It integrates observation processing, decision-making and memory, and action execution. Role-specific instructions and action sets support firefighting and rescue robots, while the shared architecture accommodates different observation interfaces and three control structures: single-robot (SR) control, decentralized multi-robot (DMR) control, and centralized multi-robot (CMR) control.

\subsection{Observation Processing and Action Grounding}
\agent{} resolves model-selected action targets to locations in \bench{} using procedures tailored to each observation interface. Under the Global Symbolic and Local Symbolic interfaces, the model receives structured entity records covering the entire scene or the robot’s current view, respectively. For entity-targeted actions, the model selects an entity reference, which the execution interface resolves to a target location.

Under the Local Visual interface, we evaluate two configurations: RGB and RGB+DS. In the RGB configuration, the model receives first-person RGB images and specifies action targets as image pixels. The execution interface back-projects these pixels into 3D locations using aligned depth maps and camera parameters. RGB+DS augments RGB observations with Grounding DINO~\cite{liu2023groundingdino} detections and SAM~2~\cite{ravi2024sam2} segmentations, providing annotated images, selectable entity references, and approximate target distances. Models can select these references as action targets, while the original RGB images and pixel-based selection remain available when detections are missing or unreliable. Depth supports grounding and RGB+DS distance estimation; depth maps and full target coordinates are not provided to the model(Appendix~\ref{appendix:observation_interfaces}).

\subsection{Decision, Memory, and Execution}
\label{sec:decision_memory_execution}

\noindent\textbf{Decision context and memory.}
Each evaluated model selects an action based on task and role instructions, current observations, robot state, remaining time, memory, and execution feedback. Each structured response specifies an action and its parameters, a brief rationale, a memory update, and an optional communication message. Within each episode, the agent maintains a bounded history of recent decisions and their outcomes, together with a note that preserves useful information no longer included in this history (Appendix~\ref{appendix:agent_context_memory}).

\noindent\textbf{Execution and feedback.}
The execution interface validates and executes the selected action, resolving action targets as described in Section 4.1. The agent relies on environmental feedback to assess its decision's outcome, since command completion alone does not establish success. For example, a completed suppression command may leave the target fire burning. Observations collected during execution, together with execution feedback, update the interaction history and inform subsequent decisions. In real-time runs, the environment continues to evolve during both model inference and action execution (Appendices~\ref{appendix:interaction_timing} and~\ref{appendix:agent_execution_coordination}).

\noindent\textbf{Event-triggered replanning.}
Civilian distress signals and urgent teammate messages trigger replanning before the current action finishes. The new information informs the next decision, and pending commands rendered outdated by the event are discarded. Under decentralized control, these events interrupt the receiving agent’s ongoing action. Under centralized control, distress signals trigger team-level replanning, while ongoing actions are interrupted only when replaced by new assignments (Sectionx~\ref{sec:agent_coordination}).

\subsection{Agent Organization and Coordination}
\label{sec:agent_coordination}
\noindent\textbf{Control and scheduling.}
DMR implements decentralized multi-robot control, with an independent decision-making agent for each robot. CMR implements centralized multi-robot control, with one agent making decisions for the team. Both execute robot actions asynchronously. Under DMR, agents advance independently; Under CMR, any robot completing its action and final observation can trigger a central decision while others continue executing. The center issues a new action or \texttt{continue} for each robot; a new action interrupts and replaces an ongoing one.

\noindent\textbf{Information and communication.}
Under DMR, agents maintain separate memories and broadcast model-selected messages labeled \emph{urgent}, \emph{important}, or \emph{useful}. Urgent messages trigger replanning as described in Section~\ref{sec:decision_memory_execution}, while other messages are incorporated at the next decision. The CMR controller maintains team memory and combines robot states, execution feedback, and the latest complete observation batches permitted by the selected interface. With local inputs, pooling remains limited to information acquired by the team. Further details on scheduling and communication are provided in Appendix~\ref{appendix:agent_execution_coordination}.

\section{Experiments}
\subsection{Experimental Setup}

\noindent\textbf{Tasks and Scenarios.}
We evaluate all five tasks (Section~\ref{sec:task_design}) on six
residential scenarios varying in fire conditions and civilian placement,
using civilian-free variants for suppression-only tasks.
Within each task and comparison condition, all models use the same
scenarios and initial configurations
(Appendix~\ref{appendix:experiment_scenarios}).

\noindent\textbf{Models and Agent Configuration.}
We evaluate GPT-6 Luna (GPT-6), GPT-6.1 Terra (GPT-6.1), Claude Haiku 5.5 (Claude),
DeepSeek V4 Flash S2 (DeepSeek), and GLM-5.3-Flash (GLM) using the \agent{}
framework (Section~\ref{sec:agent_baselines}), with model settings
detailed in Appendix~\ref{appendix:model_configuration}.
Evaluations with symbolic observations use the global or local interface specified by
each task, while visual evaluations compare RGB and RGB+DS
(Appendix~\ref{appendix:interfaces}).
The main T4 and T5 results use four FRs under DMR with communication
enabled; control structure, communication, team size, and role
composition are varied in Section~\ref{sec:coordination_results}.

\noindent\textbf{Evaluation Protocol and Human Comparison.}
Each episode begins after a short preburn period and runs for 300 seconds of environment time. For each experimental condition, we conduct three runs per scenario, compute the mean and variance within each scenario, and average these statistics across the six scenarios. Matched runs without robot intervention provide free-burning references, while single-robot runs with the same observation interface and task objectives provide references for team gain (Appendix G.4). We recruit 15 human participants. 12 form three four-person teams for T5, and the remaining three independently perform the local visual variant of T3 (Appendix G.5).

\subsection{Overall Task Performance}
\label{sec:benchmark_results}

\noindent\textbf{Symbolic results.}
Table~\ref{tab:oracle_results} reports performance across T1--T5
using the symbolic observation interface, with per-scenario
results and dispersion statistics in
Appendix~\ref{appendix:complete_results}.
Adding civilian rescue to single-robot suppression
(T1$\rightarrow$T2) reduces SVR by 8.99--18.81 percentage
points (pp) and increases FCT for all five models.
This reduction must be assessed alongside civilian outcomes,
since T2 prioritizes civilian safety.
Lower property preservation does not necessarily yield better
rescue: Claude shows the largest SVR decrease, whereas GPT-6.1
achieves the highest T2 SVR, HPR, and CRR.
The challenge is therefore to manage suppression and rescue
jointly, rather than simply sacrifice one objective for the other.

\begin{table*}[t]
    \centering
    \footnotesize
    \setlength{\tabcolsep}{1pt}
    \renewcommand{\arraystretch}{1.12}
    \caption{Performance for symbolic observation across T1--T5.
    Each model--task condition includes six scenarios with three trials
    per scenario. Values are scenario-averaged means.
    All metrics are percentages except FCT (s) and TG (pp).
    Bold indicates the best model mean for each task and metric.
    Per-scenario results and dispersion statistics are provided in
    Appendix~\ref{appendix:complete_results}. T4 and T5 use four FRs under DMR with communication.
    TG compares T4 with T1 and T5 with a matched single-robot
    condition retaining local observations and rescue objectives.
    FCT includes the terminal penalty defined in Section~\ref{Chapter:Bench_Metrics}}
    \label{tab:oracle_results}

    \begin{tabular*}{\textwidth}{
        @{\extracolsep{\fill}}l*{18}{c}@{}
    }
        \toprule
        Model
        & \multicolumn{2}{c}{T1}
        & \multicolumn{4}{c}{T2}
        & \multicolumn{3}{c}{T3}
        & \multicolumn{3}{c}{T4}
        & \multicolumn{6}{c}{T5} \\
        \cmidrule(lr){2-3}
        \cmidrule(lr){4-7}
        \cmidrule(lr){8-10}
        \cmidrule(lr){11-13}
        \cmidrule(lr){14-19}
        & SVR $\uparrow$ & FCT $\downarrow$
        & SVR $\uparrow$ & FCT $\downarrow$
        & HPR $\uparrow$ & CRR $\uparrow$
        & SVR $\uparrow$ & FCT $\downarrow$ & EC $\uparrow$
        & SVR $\uparrow$ & FCT $\downarrow$ & TG $\uparrow$
        & SVR $\uparrow$ & FCT $\downarrow$
        & HPR $\uparrow$ & CRR $\uparrow$
        & EC $\uparrow$ & TG $\uparrow$ \\
        \midrule

        GPT-6.1
        & \textbf{37.56} & 1020.3
        & \textbf{22.21} & \textbf{1281.5}
        & \textbf{71.01} & \textbf{94.44}
        & 19.24 & 1785.5 & \textbf{72.39}
        & 57.26 & 376.1 & 19.70
        & 39.97 & 1334.0 & \textbf{51.35}
        & \textbf{86.11} & 80.25 & \textbf{25.36} \\

        GPT-6 
        & 28.27 & 1200.2
        & 19.28 & 1480.9 & 67.66 & 80.56
        & 22.33 & 1188.0 & 38.94
        & \textbf{58.18} & \textbf{302.8} & \textbf{29.92}
        & \textbf{40.98} & \textbf{817.2}
        & 48.31 & 52.78 & 71.86 & 19.58 \\

        Claude
        & 27.04 & 1097.0
        & 8.23 & 1544.4 & 61.03 & 86.11
        & \textbf{23.59} & \textbf{1127.8} & 37.29
        & 48.34 & 575.2 & 21.30
        & 27.46 & 1446.2 & 48.86 & 72.22
        & \textbf{80.62} & 15.34 \\

        DeepSeek
        & 32.51 & \textbf{940.9}
        & 16.34 & 1333.3 & 57.43 & 69.44
        & 17.07 & 1806.0 & 41.25
        & 49.39 & 350.3 & 16.88
        & 26.15 & 2040.9 & 30.74 & 50.00 & 60.34 & 10.04 \\

        GLM
        & 23.02 & 1255.9
        & 12.23 & 1550.0 & 41.27 & 72.22
        & 18.74 & 1394.4 & 40.74
        & 42.76 & 584.4 & 19.73
        & 30.42 & 1476.6 & 44.79 & 55.56 & 66.28 & 15.65 \\

        \bottomrule
    \end{tabular*}

    \vspace{2pt}
\end{table*}

Restricting observations from global to local
(T1$\rightarrow$T3) reduces every model's SVR by
3.45--18.32~pp.
GPT-6.1 leads T1 but experiences the largest decline, while
Claude has the smallest decline and achieves the highest T3 SVR.
Notably, GPT-6.1 attains much higher exploration coverage than
Claude (72.39\% versus 37.29\% EC), yet lower SVR
(19.24\% versus 23.59\%).
This mismatch suggests difficulty in balancing information
acquisition with timely intervention: observing more of the
scene does not ensure that the acquired information translates
into better outcomes.
For an independent DeepSeek T1 comparison of inference-time
effects, see Appendix~\ref{appendix:decision_time}
and~\ref{appendix:fire_sensitivity}.

Expanding from one robot to four under decentralized control
(T1$\rightarrow$T4) improves SVR and lowers FCT for every
model, with team gains of 16.88--29.92~pp.
GPT-6 achieves the largest gain, increasing SVR from
28.27\% to 58.18\%.
Although it trails GPT-6.1 by 9.29~pp in T1, it leads by
0.92~pp in T4.
The best-performing model under single-robot control thus
does not retain its lead when decisions and actions are
distributed across a robot team.

T5 combines local observations, civilian rescue, and
multi-robot operation.
All models achieve positive team gains of 10.04--25.36~pp
over matched single-robot references.
Nevertheless, their SVR falls by 12.34--23.24~pp relative
to T4, and FCT increases for every model, showing that
additional rescue and information-gathering demands
challenge even four robots.
Model rankings vary across objectives: GPT-6 achieves the
highest SVR, while GPT-6.1 leads HPR and CRR.
Their SVR and HPR differ by only 1.01 and 3.04~pp,
respectively, but GPT-6.1's CRR is 33.33~pp higher.
This gap distinguishes protecting civilians from completing
their evacuation: suppression can reduce health loss while
civilians remain outside the safe zone.

\noindent\textbf{Visual results.}
Compared with symbolic observations, both visual configurations
(RGB and RGB+DS) yield lower SVR for every model on T3 and T5
(Tables~\ref{tab:oracle_results} and~\ref{tab:vision_results}).
On T5, HPR and CRR also decrease throughout, with CRR falling
by 19.44--50.00~pp.
Model rankings change across observation interfaces:
Claude leads T3 SVR with symbolic observations but ranks
third with RGB and fifth with RGB+DS.
On T5, GPT-6 achieves the highest symbolic SVR, whereas
GPT-6.1 leads under both visual configurations.
Leading performance with symbolic observations thus does not
consistently translate into leading performance with visual inputs.
Some patterns across objectives nevertheless persist.
GPT-6 achieves higher T5 SVR than Claude under all three
configurations, but consistently lower CRR, indicating stronger
relative performance in property preservation than in
completed evacuation.
GPT-6.1 maintains the highest T5 CRR across all three
configurations, showing a consistent advantage in
evacuation performance.

\begin{table*}[t]
    \centering
    \footnotesize
    \setlength{\tabcolsep}{1pt}
    \renewcommand{\arraystretch}{1.08}
    \caption{Performance for visual observation on T3 and T5 under RGB and RGB+DS.
    For each model--task--interface condition, sampling, aggregation,
    units, and default team settings follow Table~\ref{tab:oracle_results}.
    Bold indicates the best model mean for each task and metric
    across both interfaces. TG uses a matched single-robot visual fire-and-rescue reference
        for each model and interface.
        Human results are descriptive references
        (Appendix~\ref{appendix:human}).
        Human TG is unavailable without a matched single-robot reference.}
    \label{tab:vision_results}

    \begin{tabular*}{\textwidth}{
        @{\extracolsep{\fill}}l*{18}{c}@{}
    }
        \toprule
        Model
        & \multicolumn{6}{c}{T3}
        & \multicolumn{12}{c}{T5} \\
        \cmidrule(lr){2-7}
        \cmidrule(lr){8-19}

        & \multicolumn{3}{c}{RGB}
        & \multicolumn{3}{c}{RGB+DS}
        & \multicolumn{6}{c}{RGB}
        & \multicolumn{6}{c}{RGB+DS} \\
        \cmidrule(lr){2-4}
        \cmidrule(lr){5-7}
        \cmidrule(lr){8-13}
        \cmidrule(lr){14-19}

        & SVR $\uparrow$ & FCT $\downarrow$ & EC $\uparrow$
        & SVR $\uparrow$ & FCT $\downarrow$ & EC $\uparrow$
        & SVR $\uparrow$ & FCT $\downarrow$
        & HPR $\uparrow$ & CRR $\uparrow$
        & EC $\uparrow$ & TG $\uparrow$
        & SVR $\uparrow$ & FCT $\downarrow$
        & HPR $\uparrow$ & CRR $\uparrow$
        & EC $\uparrow$ & TG $\uparrow$ \\
        \midrule

        GPT-6.1
        & \textbf{15.69} & \textbf{1503.2} & \textbf{53.93}
        & 7.74 & 1699.7 & \textbf{49.53}
        & \textbf{31.64} & 1291.3
        & \textbf{38.21} & \textbf{66.67} & 76.64 & \textbf{22.23}
        & \textbf{30.37} & \textbf{1153.4}
        & \textbf{32.86} & \textbf{61.11} & \textbf{71.34} & \textbf{21.76} \\

        GPT-6
        & 5.46 & 1700.0 & 22.70
        & \textbf{22.04} & \textbf{1088.4} & 24.44
        & 27.02 & \textbf{1189.9}
        & 22.87 & 16.67 & 54.85 & 16.71
        & 22.86 & 1272.2
        & 16.48 & 8.33 & 47.50 & 10.70 \\

        Claude
        & 7.77 & 1622.2 & 25.42
        & 7.07 & 1650.0 & 28.96
        & 16.02 & 1616.7
        & 17.65 & 25.00 & 68.68 & 7.56
        & 15.67 & 1344.4
        & 17.84 & 25.00 & 70.51 & 12.32 \\

        DeepSeek
        & 4.07 & 1938.9 & 40.14
        & 9.14 & 1661.1 & 31.12
        & 12.76 & 2294.4
        & 9.18 & 0.00 & \textbf{81.39} & 10.43
        & 24.29 & 1433.3
        & 14.42 & 16.67 & 63.21 & 13.77 \\

        GLM
        & 13.41 & 1600.0 & 28.10
        & 10.41 & 1638.9 & 32.43
        & 24.16 & 1433.3
        & 25.68 & 27.78 & 59.03 & 10.61
        & 10.20 & 1916.7
        & 8.25 & 8.33 & 53.13 & 4.85 \\

        \midrule
        Human
        & 25.74 & 1105.9 & 35.77
        & \multicolumn{3}{c}{\textendash}
        & 42.34 & 627.8
        & 48.88 & 58.33 & 70.81 & \textendash
        & \multicolumn{6}{c}{\textendash} \\
        \bottomrule
    \end{tabular*}

    \vspace{2pt}
        
\end{table*}

\noindent\textbf{RGB versus RGB+DS.}
RGB+DS augments RGB observations with detection and segmentation
annotations, target identifiers, and approximate distances,
allowing identifier-based target selection.
These additions improve SVR for two of five models on T3,
but only DeepSeek on T5.
For GPT-6, RGB+DS raises T3 SVR by 16.58~pp, nearly matching
its symbolic result (22.04\% versus 22.33\%), yet reduces
T5 SVR by 4.16~pp and CRR by 8.34~pp.
Its benefit for single-robot suppression therefore does not
carry over to joint rescue and team operation.
The effect also differs across models within the same task:
on T5, RGB+DS improves DeepSeek's SVR, HPR, and CRR by
11.53, 5.24, and 16.67~pp, respectively, while reducing
all three for GLM.
Additional visual processing and target-selection support
therefore do not consistently improve task outcomes;
their benefits depend on both the model and the task.



\noindent\textbf{Comparison with the human reference.}
Under the RGB configuration, all evaluated models achieve
lower SVR than the human reference on both T3 and T5,
with the best model trailing by 10.05 and 10.70~pp,
respectively (Appendix~\ref{appendix:human}).
On T5, all RGB models also achieve lower HPR and higher FCT,
with the best HPR falling short by 10.67~pp.
Evacuation provides a contrasting result:
GPT-6.1 exceeds the human CRR (66.67\% versus 58.33\%),
whereas the other RGB models achieve at most 27.78\%.
Despite completing more evacuations, GPT-6.1 preserves
less civilian health overall (38.21\% versus 48.88\% HPR).
Its higher evacuation rate therefore coexists with
a remaining gap in health protection.

\subsection{Effects of Team Organization}
\label{sec:coordination_results}

\noindent\textbf{Effects of control organization and communication.}
\begin{table*}[t]
\centering
\footnotesize
\setlength{\tabcolsep}{2.5pt}
\renewcommand{\arraystretch}{1.10}
\caption{Four-FR team performance under different control and
communication settings. DMR uses broadcasts unless marked w/o comm.
TG uses a matched single-robot reference for each task--model--input
condition. Dashes denote inapplicable metrics.
Sampling, aggregation, and metric units follow
Table~\ref{tab:oracle_results}.
FCT and dispersion statistics are reported in
Appendix~\ref{appendix:coordination_details}.}
\label{tab:coordination_results}

\begin{tabular*}{\textwidth}{@{\extracolsep{\fill}}lll*{15}{r}@{}}
\toprule
Task & Model & Obs.
& \multicolumn{5}{c}{CMR}
& \multicolumn{5}{c}{DMR}
& \multicolumn{5}{c}{DMR w/o comm.} \\
\cmidrule(lr){4-8}
\cmidrule(lr){9-13}
\cmidrule(lr){14-18}
& &
& SVR$\uparrow$ & HPR$\uparrow$ & CRR$\uparrow$ & EC$\uparrow$ & TG$\uparrow$
& SVR$\uparrow$ & HPR$\uparrow$ & CRR$\uparrow$ & EC$\uparrow$ & TG$\uparrow$
& SVR$\uparrow$ & HPR$\uparrow$ & CRR$\uparrow$ & EC$\uparrow$ & TG$\uparrow$ \\
\midrule

T4 & DeepSeek & Symbolic
& 28.08 & \textendash & \textendash & \textendash & -4.43
& 49.39 & \textendash & \textendash & \textendash & 16.88
& 38.27 & \textendash & \textendash & \textendash & 5.76 \\
\midrule

T5 & DeepSeek & Symbolic
& 24.29 & 24.58 & 36.11 & 61.58 & 8.18
& 26.15 & 30.74 & 50.00 & 60.34 & 10.04
& 23.47 & 35.96 & 47.22 & 60.85 & 7.37 \\

T5 & GLM & Symbolic
& 35.17 & 40.90 & 50.00 & 72.89 & 20.40
& 30.42 & 44.79 & 55.56 & 66.28 & 15.65
& 33.96 & 52.16 & 55.56 & 63.14 & 19.19 \\

T5 & DeepSeek & RGB
& 12.02 & 10.12 & 0.00 & 85.02 & 9.70
& 12.76 & 9.18 & 0.00 & 81.39 & 10.43
& 11.16 & 8.82 & 0.00 & 83.78 & 8.84 \\

T5 & DeepSeek & RGB+DS
& 9.52 & 8.48 & 0.00 & 65.97 & -1.00
& 24.29 & 14.42 & 16.67 & 63.21 & 13.77
& 15.28 & 18.13 & 25.00 & 67.97 & 4.77 \\

\bottomrule
\end{tabular*}

\par\vspace{2pt}
\end{table*}
Table~\ref{tab:coordination_results} compares centralized control
and decentralized control with and without communication,
using four firefighting robots throughout.
We primarily evaluate DeepSeek and include GLM on T5 with
symbolic observations as an additional model comparison.
Decentralized control with communication achieves the highest
mean SVR in four of the five evaluated conditions.
However, its advantage is model-dependent.
On T5 with symbolic observations, DeepSeek achieves higher SVR
under decentralized than centralized control
(26.15\% versus 24.29\%), whereas GLM shows the opposite
pattern (30.42\% versus 35.17\%).

Communication likewise produces mixed effects.
Enabling broadcasts increases DeepSeek's SVR in every
evaluated condition, but decreases GLM's T5 SVR from
33.96\% to 30.42\%.
Moreover, broadcasts reduce HPR in all four T5 conditions,
while CRR improves in only one, remains unchanged in two,
and declines in one.
For example, DeepSeek with RGB+DS gains 9.01~pp SVR with
communication, but loses 3.71~pp HPR and 8.33~pp CRR.
Information sharing therefore does not uniformly improve
joint suppression and rescue: its benefits depend on the
model and the task outcome considered.

\begin{figure}[t]
    \centering
    \includegraphics[width=\columnwidth]{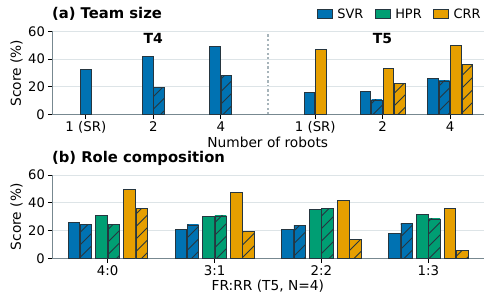}
    \caption{
    Effects of team size and role composition with DeepSeek
    and symbolic observations.
    (a) Team size on T4 (SVR) and T5 (SVR, CRR);
    $N=1$ denotes matched single-robot (SR) references.
    (b) FR:RR role composition on T5 with $N=4$.
    For multi-robot conditions, solid and hatched bars denote
    DMR and CMR, respectively; DMR uses communication.
    Bars show scenario-averaged means over six scenarios
    with three trials each.
    FCT, T5 team-size HPR, and dispersion statistics are reported
    in Appendix~\ref{appendix:coordination_details}.
    }
    \label{fig:mas_team_roles}
\end{figure}

\noindent\textbf{Effects of team size.}
Figure~\ref{fig:mas_team_roles}(a) compares multi-robot
configurations with SR using DeepSeek.
On T4, both two- and four-robot teams improve SVR under
decentralized control, whereas centralized teams remain below SR.
On T5, four-robot teams achieve higher SVR than SR under both
control organizations, but evacuation gains remain limited:
decentralized CRR only slightly exceeds SR, while centralized
CRR remains lower.
These results show that the benefits of adding robots depend
on the team's control organization and differ between
suppression and evacuation.

\noindent\textbf{Effects of role composition.}
Figure~\ref{fig:mas_team_roles}(b) varies FR/RR composition
in four-robot teams using DeepSeek.
HPR peaks at 2FR+2RR under both control organizations,
while CRR decreases as the RR share grows.
The SVR ranking reverses from decentralized control leading
in homogeneous FR teams to centralized control leading
in heterogeneous teams, whereas decentralized control
retains higher CRR throughout.
These results show that the relative performance of control
organizations depends on both role composition and task objective.
Per-scenario results and dispersion statistics appear in
Appendix~\ref{appendix:coordination_details}.
\section{Conclusion and Limitations}
We introduce \bench{} to evaluate embodied multi-agent decision-making in residential fires, where fire spread continues during deliberation and action and responds to agent interventions. Five tasks separately and jointly assess information gathering, rescue, and teamwork using outcome and temporal metrics. \agent{} enables unified evaluation across foundation models, observation interfaces, and control structures. Experiments show that similar property and health protection can conceal substantial differences in evacuation success. Benefits from visual assistance and team organization depend on the model, task, and control structure, with improvements in property preservation not necessarily extending to civilian rescue.
Our evaluation covers residential scenarios with object-level fire dynamics, abstracted low-level control, and reliable communication. Generalization to larger incidents and deployment on physical robots remain untested.

\bibliographystyle{ieeenat_fullname}
\bibliography{references}
\clearpage
\appendix
\section{Environment Assets and Scene Generation}
\label{appendix:assets}

\begin{figure*}[t]
  \centering
  \includegraphics[width=\textwidth]{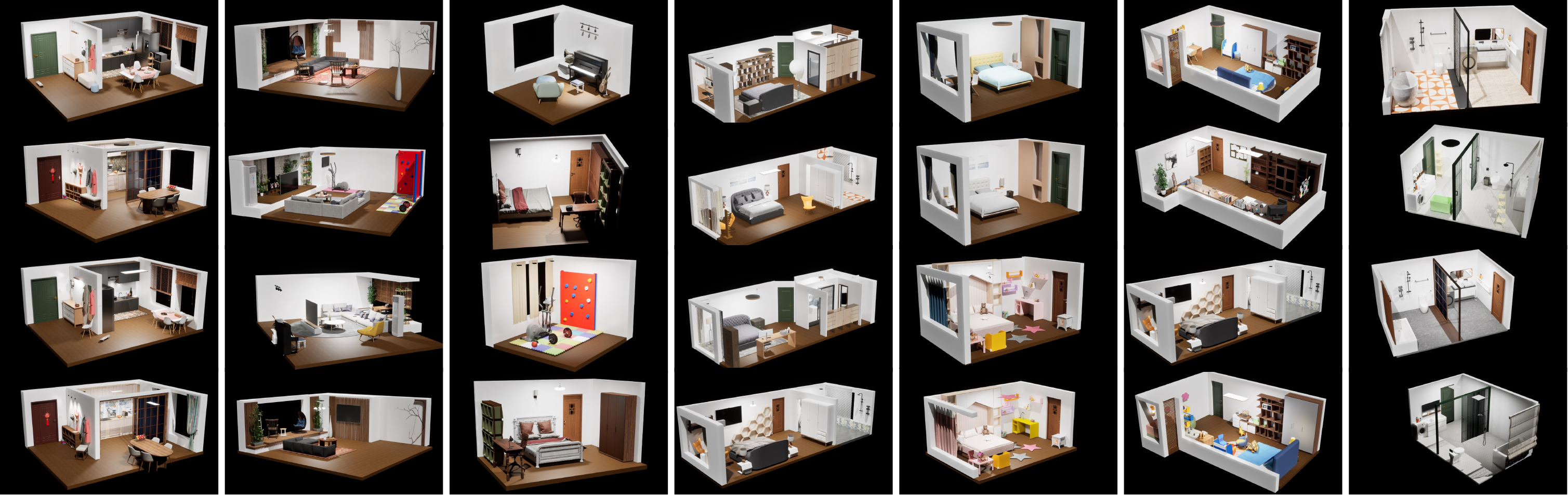}
  \caption{\textbf{Residential room library.}
  Four furnishing and layout variants for each of the seven rooms yield
  28 base templates used to compose the residential environments.}
  \label{fig:appendix_room_templates}
\end{figure*}

\bench{} constructs residential environments from a seven-room apartment
composition: a living room, a kitchen, a bathroom with bathing facilities,
a study, two secondary bedrooms, and a primary bedroom with an en-suite
bathroom. Each room has four furnishing and layout variants, giving
28 base room templates (Fig.~\ref{fig:appendix_room_templates}).
The left and right apartments use corresponding layouts, with mirrored
placements and local detail differences.

\begin{figure}[t]
  \centering
  \includegraphics[width=\columnwidth]{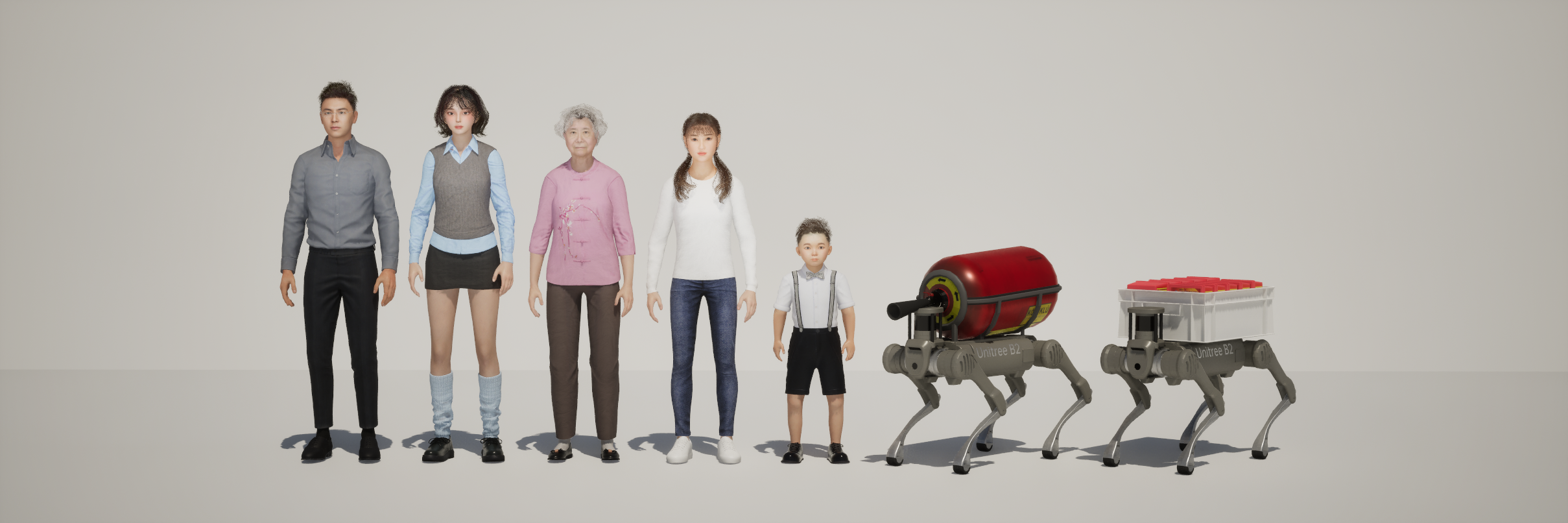}
  \caption{\textbf{Civilian and robot assets.}
  Five civilian appearances are shown on the left; the firefighting and
  rescue robots are shown on the right.}
  \label{fig:appendix_entity_assets}
\end{figure}

The library includes furniture and combustible objects, five civilian
appearances, and firefighting and rescue robot assets
(Fig.~\ref{fig:appendix_entity_assets}). All assets were created by the
authors and correspond to the version included in the platform release.
Fire dynamics, civilian protection, and robot action interfaces are
described in Appendices~\ref{appendix:fire_model},
\ref{appendix:victim_model}, and~\ref{appendix:action_space}.

\subsection{Scene Configuration and Episode Initialization}
\label{appendix:scene_generation}

A scene is specified by room variants, ignition and civilian placement,
fire-dynamics parameters, and a safe area
(Table~\ref{tab:appendix_scene_fields}). The three room-indexed arrays use
a shared order over the 14 spatial room slots in the two apartments.
Users can specify room variants explicitly; otherwise, the generator
samples them. The random seed controls room variants, the initial fire
configuration, and the NPC configuration. Baseline lighting parameters
remain fixed, while rendered illumination and visibility change with
the evolving fire and smoke.

\begin{table}[h]
  \centering
  \small
  \setlength{\tabcolsep}{3pt}
  \renewcommand{\arraystretch}{1.12}
  \begin{tabular}{@{}p{0.40\columnwidth}p{0.56\columnwidth}@{}}
    \hline
    \textbf{Field} & \textbf{Meaning and units} \\
    \hline
    \texttt{RoomGenerate} &
      Variant codes indexed by spatial room slot. \\
    \texttt{FireRoom} &
      Room-indexed 0/1 flags enabling initial ignition. \\
    \texttt{NPCRoom} &
      Room-indexed 0/1 flags enabling civilians. \\
    \texttt{SafeArea} &
      Minimum and maximum world-coordinate bounds of the safe area,
      in UE units (100 units per meter). \\
    \texttt{FireMultiply1} &
      Initial (preburn) burning-rate multiplier. \\
    \texttt{FireMultiply2} &
      Subsequent burning-rate multiplier. \\
    \texttt{FireRateDelayTime} &
      Fire-rate phase-transition delay, in seconds;
      aligned with the preburn wait in the reported setup. \\
    \texttt{FireSpreadScale} &
      Fire-spread scaling factor. \\
    \texttt{OutfireMultiply} &
      Suppression multiplier. \\
    \hline
  \end{tabular}
  \caption{\textbf{Scene-configuration fields.}
  The three room arrays share the same spatial-slot order.
  Multipliers and scaling factors are dimensionless.
  Array flags specify enabled rooms; the post-preburn world state is
  measured after initialization.}
  \label{tab:appendix_scene_fields}
\end{table}

Each episode reconstructs the configured scene and, where applicable,
prepares and deploys the robots before ignition. A pre-ignition state is
recorded, the fire is started, and the configured preburn period elapses
before the evaluation clock begins. The initial evaluation observation
therefore describes the post-preburn world and may contain more burning
objects than the configured ignition sources. Clock definitions are
specified in Appendix~\ref{appendix:interaction_timing}; the adopted
evaluation settings are documented in
Appendix~\ref{appendix:experiment_scenarios}. Complete configuration
examples accompany the supplementary material.

\subsection{Dataset Construction and Release}
\label{appendix:dataset_release}

We select ignition sources from free-burning observations, excluding
fires that burn out before robots can act. Among eligible sources, type~A
fires stay on one object and type~B fires spread within the same room.
Types~C and~D reach a neighboring room late and early, respectively.
The single-source observations use a provisional 30-s boundary between them.

We combine one to four of the 42 eligible ignition sources at a time,
keeping each source in the layout in which we observed it. Across four
layouts, this yields 2,077 candidate combinations. We select 692 by
source-profile composition, covering different ignition counts and both
single-wing and cross-wing fires. We retain every single-source case
and all combinations made up solely of the rare type~A sources.
For each selected combination, we hold the layout, ignition rooms, and
fire parameters fixed across three civilian settings. The setting
without civilians provides a firefighting reference. The other two
place two civilians in separate rooms, either in the same wing or one
in each wing. At a fixed civilian count, this pairing lets researchers
compare how teams divide their efforts between suppression and evacuation
when rescue targets are concentrated or dispersed. The 692 matched
triplets yield 2,076 complete configurations.

In \texttt{manifest.csv}, each triplet shares an identifier. The field
\texttt{measurement\_status} marks 27 configurations as
\texttt{direct\_measured}, each with a matching free-burning run (5-s
preburn, 300-s evaluation). The other 2,049 carry the label
\texttt{source\_evidence\_only} for evidence from source observations or
related runs. The archive includes the 12 original configurations for the
six evaluation scenarios (Appendix~\ref{appendix:experiment_scenarios}).
We plan to release raw trajectories separately on Hugging Face.

\section{Hazard Dynamics and Civilian Protection}

\subsection{Object-Level Fire Dynamics}
\label{appendix:fire_model}
\subsubsection{State Variables and Initialization}
In the benchmark model, $Q_{j,t}$ denotes the remaining fuel fraction
of combustible object $j$ at time $t$, and $I_{j,t}$ denotes its
normalized fire intensity. We report both states on a $0$--$1$ scale.
A full reserve has $Q_{j,t}=1$, and an exhausted one has $Q_{j,t}=0$.
Evaluation starts from the state reached at the end of preburn.

\subsubsection{Combustion and Suppression Updates}
During combustion, the fuel reserve and fire intensity evolve over
an interval of $\Delta t$ seconds as
\begin{align}
  Q_{j,t+\Delta t}
  &= \max\!\left(0,\, Q_{j,t}-\beta_j\Delta t\right),
  \label{eq:appendix_fuel_update}\\
  I_{j,t+\Delta t}
  &= \max\!\left(0,\, I_{j,t}+g_j\Delta t
     -\eta_j N_{j,t}^{\mathrm{sup}}\Delta t\right).
  \label{eq:appendix_intensity_update}
\end{align}
The intensity update enforces only the lower bound of zero.
The fuel-consumption rate $\beta_j$ and natural fire-growth rate $g_j$
are fixed for each object. Each of the $N_{j,t}^{\mathrm{sup}}$
firefighting robots suppressing object $j$ contributes a suppression
rate of $\eta_j$, giving a total rate of $\eta_j N_{j,t}^{\mathrm{sup}}$.
Fuel loss is permanent. Table~\ref{tab:appendix_scene_fields} gives
the burning-rate multipliers for preburn and evaluation.

\subsubsection{Propagation and Extinction}
Object $j$ triggers one neighborhood propagation event the first time
$I_{j,t}>\theta_j^{\mathrm{prop}}$. The propagation threshold is
$0.40$ in the reported setting. The query is centered at the world-space
center of the source object's bounding box. Each half-extent is
multiplied by \texttt{FireSpreadScale}. The event ignites each unburned
combustible object found by this box query. Contact can also ignite an
unburned combustible object when its mesh begins to overlap an actor
tagged \texttt{burning}, independently of the propagation threshold.

After a combustion update, fuel exhaustion ($Q_{j,t+\Delta t}=0$) or
zero intensity ($I_{j,t+\Delta t}=0$) makes the object permanently
inactive. The simulator removes an extinguished object if
$Q_{j,t+\Delta t}\leq 0.10$ and its \texttt{destructible} flag is 1.

\begin{figure*}[t]
  \centering
  \includegraphics[width=\textwidth]{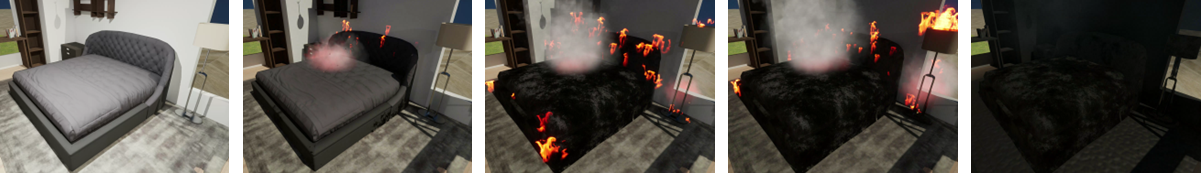}
  \caption{\textbf{Fire evolution in a bedroom.} From left to right,
  the bed before ignition, the first visible flames, growth of the
  fire on the bed, spread to nearby furnishings, and the darkened room
  after burnout.}
  \label{fig:appendix_fire_sequence}
\end{figure*}

\subsubsection{Effects on Visual Observations}
Fire dynamics alter visual observations through evolving flame and smoke particles, persistent changes in object appearance and geometry, light deactivation, and the accumulation of smoke stains on surrounding surfaces.
Flames and airborne smoke are rendered using a Niagara particle system with multiple emitters. Some particles are spawned from the object's mesh surface to align the effect with its geometry. Fire intensity controls inputs for flame strength, particle size, spawn count, and drag, while velocity and acceleration govern particle motion. Lifetime-dependent curves modulate size, color, and opacity, and SubUV animation adds temporal variation to particle textures. Collision queries further adapt particle motion and, for selected emitters, size and color to the surrounding geometry.

As remaining fuel decreases, the overlay material advances the burn pattern across the object's surface. World-space height and animated noise produce an irregular burning boundary, while dark surface textures and orange-red emissive highlights depict charring and glowing burn edges. Objects without the \texttt{Unscale} tag also shrink uniformly along all axes as fuel is consumed. Upon extinguishment, the overlay switches to a non-burning appearance while retaining the damage pattern and scale associated with the final remaining fuel. 

At ignition, the simulator switches off the nearest light returned by its light-selection function. Lights detected by the propagation query are also switched off. 

Smoke stains are simulated using dynamic decals. When an object ignites, the system checks for an existing smoke-stain decal in the corresponding region and creates one if none is detected, with its projection extent determined by the object's bounding box. The spread duration depends on the base burning rate and the burning speed multiplier, such that faster burning leads to more rapid stain expansion and darkening. After creation, the decal's radius and opacity parameters increase linearly over time until the radius exceeds a predefined threshold, after which the final stain is retained. This process proceeds independently of subsequent fire intensity, allowing stains to continue developing after extinguishment without automatically fading.

\subsection{Civilian Exposure and Protection}
\label{appendix:victim_model}
\subsubsection{Exposure and Health Updates}
Each civilian starts with health $H=100$ on a $0$--$100$ scale.
The benchmark evaluates exposure every $h=0.2$~s. A civilian is
exposed when it lies within radius $R_{\mathrm{exp}}$ of a burning
object. On each exposed check, the exposure counter 
increases by one, adding $h$ to the accumulated exposure time
$\tau^{\mathrm{exp}}$ before health is updated.

Let $d_t^{\min}$ be the distance in meters from the civilian to the
center of the nearest burning object. With $\tau^{\mathrm{exp}}$ in
seconds, the health update during exposure is
\begin{equation}
  H_{t+h}=\max\!\left(0,\,
  H_t-\lambda_H\mu_t^{\mathrm{mask}}
  \frac{\tau_{t+h}^{\mathrm{exp}}}{d_t^{\min}}\right).
  \label{eq:appendix_health_update}
\end{equation}
Here $\lambda_H$ is the per-update damage coefficient for the fixed
health-update interval. If no burning object lies within
$R_{\mathrm{exp}}$, health and accumulated exposure retain their
values. Accumulation resumes from the stored total on re-entry. Both
quantities carry over from preburn to the start of evaluation.

\subsubsection{Protection, Following, and Evacuation}
Every 5~s, the simulator checks a $2\,\mathrm{m}\times2\,\mathrm{m}$
box around each unmasked civilian for a rescue robot (RR) with masks
in stock. If one is present, the civilian automatically receives a
mask. Each civilian receives at most one. Equipping it changes
$\mu_t^{\mathrm{mask}}$ from 1 to 0.2, reducing subsequent damage to
one fifth for the same exposure and distance.

A mobile, unbound civilian can follow a firefighting robot (FR) or an RR.
While bound, it follows its escort's route and ignores other robots'
recruitment requests. On release, it waits in place and can be recruited
again if mobile. Appendix~\ref{appendix:action_space} gives the commands
and recruitment range.

A civilian with $0<H<10$ is incapacitated and stops following.
A civilian with $H=0$ is dead. A mobile civilian who enters the safe
zone continues to follow until explicitly released by the escort.
The same exposure rule governs health inside the zone. The civilian rescue rate
counts civilians who finish the episode with positive health in the
safe zone, including its boundary (Eq.~\ref{eq:appendix_crr}).

\subsubsection{Environmental Distress Signals}
Every 11~s, a mobile, unbound civilian outside the safe zone broadcasts
an SOS within a $10\,\mathrm{m}\times10\,\mathrm{m}$ neighborhood.
Broadcasts stop when the civilian starts following an escort, enters
the safe zone, becomes incapacitated, or dies. FRs and RRs receive
these signals under the same conditions. Each SOS packet identifies
the receiving robot and provides approximate bearings and ranges for
nearby calls.
Appendix~\ref{appendix:agent_execution_coordination} describes how
FireAgents handles repeated signals and replans after reception.

\section{Action Interfaces}

\subsection{High-Level Actions and Robot Roles}
\label{appendix:action_space}
FRs and RRs share the actions in
Table~\ref{tab:appendix_action_interfaces}, except that suppression is
restricted to FRs. RR mask provision is automatic
(Appendix~\ref{appendix:victim_model}).
Target representations follow the selected observation interface
(Appendix~\ref{appendix:observation_interfaces}).
Command completion and observed effects are distinguished in
Appendix~\ref{appendix:agent_execution_coordination}.

\begin{table*}[t]
  \centering
  \small
  \setlength{\tabcolsep}{3pt}
  \renewcommand{\arraystretch}{1.14}
  \begin{tabular}{@{}p{0.23\textwidth}p{0.36\textwidth}p{0.35\textwidth}@{}}
    \hline
    \textbf{Action / subtype} & \textbf{Inputs} & \textbf{Operation} \\
    \hline
    \texttt{navigate / entity} &
      \texttt{target}: available entity reference or room name &
      Navigate to the resolved destination. \\
    \texttt{navigate / relative} &
      \texttt{distance}: positive finite meters;
      \texttt{direction}: relative direction &
      Navigate to a destination relative to the heading at execution start. \\
    \texttt{navigate / image} &
      \texttt{target}: \texttt{frame\_ref} and \texttt{pixel\_xy} &
      Navigate to a standable point selected in an actionable visual frame. \\
    \texttt{suppress\_fire} &
      \texttt{target}: entity reference or supported image target &
      Aim and spray from the current position. \\
    \texttt{instruct\_civilian / follow} &
      No parameters &
      Ask nearby mobile, unbound civilians to follow. \\
    \texttt{instruct\_civilian / stop\_following} &
      No parameters &
      Release the robot's current followers. \\
    \texttt{wait} &
      No parameters &
      Remain in place for 10~s while the world continues to evolve. \\
    \texttt{explore} &
      \texttt{room}: known room name with supported geometry &
      Follow a planned route through the selected room. \\
    \hline
  \end{tabular}
  \caption{\textbf{High-level action interfaces.}
  Relative directions are \texttt{forward}, \texttt{backward},
  \texttt{left}, \texttt{right}, \texttt{front\_left}, and
  \texttt{front\_right}; the two diagonals are 45 degrees from forward.}
  \label{tab:appendix_action_interfaces}
\end{table*}

\paragraph{Room exploration.}
The executor generates an inset perimeter route from the room geometry
and visits its waypoints, adjusting or skipping unreachable points.
A full traversal returns to its starting waypoint; a partial traversal
may end at the last useful position.
Route completion does not guarantee exhaustive visual coverage.
Appendix~\ref{appendix:observation_interfaces} explains how images taken
along the route inform planning and which frames can supply targets.
After completion or interruption, a final observation precedes the next
decision (Appendix~\ref{appendix:agent_execution_coordination}).

\paragraph{Fire suppression.}
Water follows a parabolic trajectory with a square cross section of
side length 0.5~m and a maximum range of 5~m; walls block the stream.
One activation sprays for 15~s by default, and pausing the decision
process does not pause an already triggered spray.
The action does not automatically approach its target.
The platform also supports a configurable limit of $n$ spray activations
followed by a cooldown of duration $x$.
Whether this optional constraint is enabled and its parameter values
are episode settings.

\paragraph{Civilian instructions.}
One \texttt{follow} instruction can recruit multiple mobile, unbound
civilians ($H\geq 10$) in the 2~m recruitment range. The escort's
\texttt{stop\_following} instruction releases its current followers.
Appendix~\ref{appendix:victim_model} describes following behavior
and civilian state transitions.

\subsection{Low-Level Movement Controls}
\label{appendix:navigation_and_movement}
\paragraph{Navigation controls.}
World positions use UE units (100~UU = 1~m), speeds use UU/s, and
navigation is evaluated in the horizontal $XY$ plane, with $Z$ denoting
height.
The navigation primitive accepts a world-space destination, speed,
arrival radius, partial-path flag, and timeout.
Entity and image targets are resolved to destinations; relative
navigation computes its destination from the pose at execution start;
exploration invokes navigation for successive waypoints.
The controller handles NavMesh path planning and movement subject to
scene collisions, ending on arrival, failure, timeout, or cancellation.
A partial-path endpoint may differ from the original destination.

\paragraph{Other movement controls.}
Direct movement accepts a destination, speed, and positional tolerance,
with options to preserve heading, face the movement direction, or apply
a supplied forward vector.
The platform also supports manual turning and continuous-velocity
control.
These interfaces support alternative controllers; the reported experiments
use the high-level actions and do not evaluate manual-turning or
continuous-velocity policies.

\section{Observation and Interaction Protocols}

\subsection{Observation Interfaces and Target Grounding}
\label{appendix:observation_interfaces}

\paragraph{Observation scope and visibility.}
Global Symbolic reports task objects and civilians throughout the scene.
Local Symbolic reports the same attributes for entities hit by rays cast
from the robot's camera across its RGB field of view. Each ray travels
up to 20~m and stops at the first surface with collision geometry.
A wall or a piece of furniture can therefore hide an entity behind it.
Smoke and flames can obscure the RGB image while leaving these rays unobstructed.
Local Visual provides first-person views from a $640\times640$ camera
with a $90^\circ$ horizontal field of view. When the robot looks in four
directions, it captures the views in sequence, each with its own timestamp.

\paragraph{What the model observes.}
Symbolic records give each entity's name and world $XY$ position in meters
when available. For objects, they report whether the object is burning
and the percentage of the object that remains unburned. For civilians, they report health,
fire exposure, and whether the civilian is inside the safe area.
With RGB, the model must identify fire from the image. RGB+DS also
provides detector predictions. Neither visual input exposes simulator
attributes such as civilian health (Table~\ref{tab:appendix_observation_inputs}).
The executor keeps depth maps and resolved 3D target coordinates internal.
Across configurations, the model also receives its robot's state,
known room anchors, and any SOS signals or teammate reports available
to its controller (Appendix~\ref{appendix:agent_context_memory}).

\begin{table*}[t]
  \centering
  \small
  \setlength{\tabcolsep}{4pt}
  \renewcommand{\arraystretch}{1.12}
  \begin{tabular}{@{}p{0.19\textwidth}p{0.45\textwidth}p{0.30\textwidth}@{}}
    \hline
    \textbf{Configuration} & \textbf{Observation} & \textbf{How the model specifies a target} \\
    \hline
    Global Symbolic & Named entity records for the entire scene & Entity name \\
    Local Symbolic & Named entity records for camera-ray hits & Entity name \\
    Local Visual / RGB & Original RGB views with capture times and frame references &
      Frame reference and pixel in the original image \\
    Local Visual / RGB+DS & RGB inputs plus a detection list, estimated distances
      for grounded detections, and annotated views when available &
      Grounded detection reference or original-image pixel with its frame reference \\
    \hline
  \end{tabular}
  \caption{\textbf{Observation inputs and target formats.}
  Known room names can also serve as navigation targets in all four configurations.}
  \label{tab:appendix_observation_inputs}
\end{table*}

\begin{figure*}[t]
  \centering
  \begin{minipage}[t]{0.485\textwidth}
    \centering
    \includegraphics[width=0.8\linewidth]{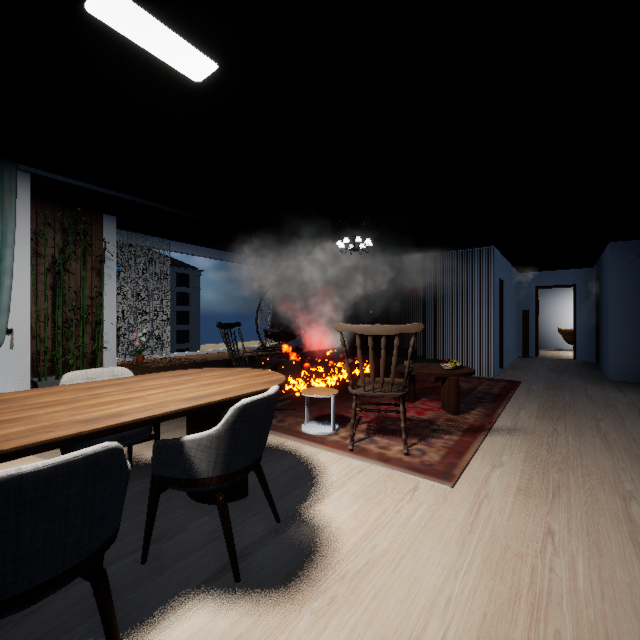}
    \\[-1pt]\small (a) Original RGB
  \end{minipage}\hfill
  \begin{minipage}[t]{0.485\textwidth}
    \centering
    \includegraphics[width=0.8\linewidth]{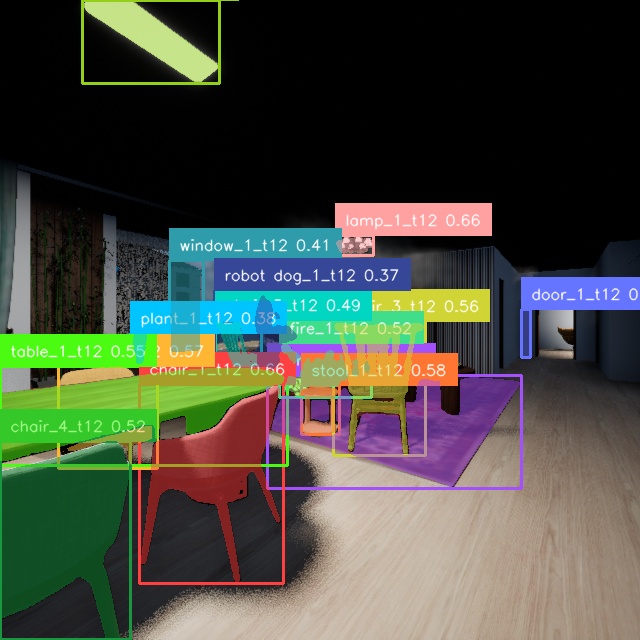}
    \\[-1pt]\small (b) RGB+DS annotation
  \end{minipage}
  \caption{\textbf{Original and annotated views of the same scene.}
  In (a), flames and smoke are visible among the chairs. Panel (b) overlays
  predicted boxes, masks, class labels, and confidence scores on the same frame.}
  \label{fig:appendix_visual_observation_example}
\end{figure*}

\paragraph{Grounding action targets.}
For a symbolic action, the executor looks up the named entity.
For a pixel target, the model selects standable ground for navigation
or a visible solid surface near the base of a fire for suppression.
It specifies the pixel in the original RGB frame, measured from the top left.
The executor estimates depth from the median of valid samples around that
pixel and uses the camera calibration and pose at capture time to recover
a 3D point. RGB+DS adds Grounding DINO detections and SAM~2 masks
(Fig.~\ref{fig:appendix_visual_observation_example}).
Depth samples within an object or person mask give a representative
position. A fire detection inherits a position only if it can be linked
unambiguously to a grounded object or person in the same frame.
The reported distance is the estimated horizontal straight-line distance
from the observing robot to that point at capture time.
If a detection is missing or unreliable, the original RGB view remains
available for pixel selection.

\paragraph{Target validity and feedback.}
At dispatch, the executor resolves visual references against the observations
bound to the current decision. A pixel must come from the commanded
robot's own views marked actionable at initialization or after an action
finishes or is interrupted. Under centralized control, a grounded detection
from a teammate's current batch can also supply a target. Images captured
during an action and older views retained in memory serve only as evidence
for planning.
If a reference is unavailable, a pixel falls outside the image, or its depth
is missing or unstable, the executor returns failure feedback. The model
can then choose another target or move to obtain a clearer view.

\subsection{Interaction Timing and Episode Lifecycle}
\label{appendix:interaction_timing}

\paragraph{Shared episode clock.}
The evaluation clock starts after preburn, before the robots take their
first observations (Appendix~\ref{appendix:scene_generation}).
All robots share one time budget and act asynchronously. In the default
real-time protocol, the world continues to evolve while a robot observes,
the model deliberates, and an action executes. A target can therefore
move, or a fire can spread, between observation and execution.

\paragraph{Pauses.}
In pause-enabled experiments, active time is elapsed wall time since the
episode began minus the total time the simulator was confirmed paused.
A pending pause request still consumes budget, and overlapping pauses
count once. A global pause stops the world for every robot.
Interrupting one robot's action leaves the others running.
Appendices~\ref{appendix:visual_trial_protocols} and~\ref{appendix:decision_time}
describe when each experimental condition pauses and resumes the world.

\paragraph{Action feedback and episode end.}
When an action finishes or is interrupted, the executor reports its
outcome to the controller. That feedback and the next observation let
the model check the effect of its action and decide what to do next
(Appendix~\ref{appendix:agent_execution_coordination}).
Evaluation continues to the shared deadline, including when all fires
have already gone out. At the deadline the controller stops dispatching
actions, so a late model reply cannot start a new action.

\section{Task Configurations and Evaluation}

\subsection{Task Configurations}
\label{appendix:task_configurations}

In T1, a firefighting robot (FR) chooses which fire to reach first
while other fires continue to spread. T2, T3, and T4 introduce
civilian rescue, first-person observation, and a team, respectively,
and T5 combines all three changes (Table~\ref{tab:appendix_task_configurations}).

\begin{table*}[t]
  \centering
  \small
  \setlength{\tabcolsep}{3pt}
  \renewcommand{\arraystretch}{1.12}
  \begin{tabular*}{\textwidth}{@{\extracolsep{\fill}}lllll@{}}
    \toprule
    \textbf{Task} & \textbf{Objective} & \textbf{Observation interface}
      & \textbf{Robot team} & \textbf{Applicable metrics} \\
    \midrule
    T1 & Fire suppression & Global Symbolic & One FR
      & SVR, FCT \\
    T2 & Suppression and rescue & Global Symbolic & One FR
      & SVR, FCT, HPR, CRR \\
    T3 & Fire suppression & Local Symbolic or Local Visual & One FR
      & SVR, FCT, EC \\
    T4 & Fire suppression & Global Symbolic & Multiple FRs
      & SVR, FCT, TG \\
    T5 & Suppression and rescue & Local Symbolic or Local Visual & Multiple robots
      & SVR, FCT, HPR, CRR, EC, TG \\
    \bottomrule
  \end{tabular*}
  \caption{\textbf{Task configurations and applicable metrics.}
  RGB and RGB+DS are Local Visual configurations (Appendix~\ref{appendix:observation_interfaces}).
  Table~\ref{tab:appendix_h2_four_robot_visual_suppression} reports EC for visual T4.}
  \label{tab:appendix_task_configurations}
\end{table*}

In T2 and T5, the task instructions give civilian safety priority over
property preservation. The robots decide whether to suppress a nearby
fire before escorting a civilian, and we score the two outcomes
separately. An FR can both suppress fires and escort civilians, so
one FR can perform T2. The main T4 and T5 evaluations use four FRs.
The T5 role-composition experiments also include rescue robots (RRs),
which can equip nearby civilians with protective masks.

\subsection{Reference Episodes and Evaluation Protocol}
\label{appendix:evaluation_protocol}

\paragraph{Evaluation window.}
We record initial property value and civilian health before ignition.
After preburn, the formal clock starts at $t=0$ and runs to the
evaluation horizon $T$. Fire clearance and exploration use samples
from $[0,T]$. We use the final snapshot for property value, civilian
health, and evacuation status even if all fires go out earlier.
Runs with pauses use the active clock in
Appendix~\ref{appendix:interaction_timing}.

\paragraph{Free-burning reference.}
To measure losses without intervention, we match a run without robots
to the evaluated scenario by room layout, initial fires, natural fire
parameters, preburn duration, horizon, and sampling protocol.
For each of the six default scenarios, we compare seven free-burning
runs by the property value lost between the pre-ignition and final
snapshots. The run with the greatest loss serves as the fixed reference
across methods. The civilian-free variant shares the corresponding property
reference. For rescue, we also match civilian placement and use the
same reference run for both property and health. A change in
suppression strength alone leaves this reference applicable because
no suppression takes place.

\paragraph{Single-robot reference.}
For team gain, we match each team trial to a single-FR trial by model,
scenario, observation interface, fire settings, timing protocol, and
repetition index. Global Symbolic T4 uses T1 as its reference. Visual T4
uses T3 with the same RGB or RGB+DS input. For T5, the single FR performs
both suppression and rescue with the team's local observation configuration
(Appendix~\ref{appendix:experiment_references}).

\subsection{Metric Definitions and Computation}
\label{appendix:metrics}

\paragraph{Fire Suppression (T1--T5).}
The \emph{Saved Value Ratio (SVR)} measures the fraction of the
free-burning property loss that intervention prevents. We measure
property value by summing the remaining durability of scene objects.
Let $V_{\mathrm{init}}$ be the total before ignition in the evaluated
run, and let $V_{\mathrm{agent}}$ and $V_{\mathrm{free}}$ be the final
totals in the intervention and free-burning runs, respectively. Then
\begin{equation}
  \mathrm{SVR} =
  \frac{V_{\mathrm{agent}}-V_{\mathrm{free}}}
       {V_{\mathrm{init}}-V_{\mathrm{free}}}.
  \label{eq:appendix_svr}
\end{equation}

The \emph{Fire Clearance Time (FCT)} uses the first valid sample
with zero active fires anywhere in the scene. Let $\mathcal{T}_0$
contain all such sample times in $[0,T]$. If none is observed, we
use the last valid fire count $n_T$ at or before $T$, provided a
sample falls at $T$ or valid samples bracket it. We compute
\begin{equation}
  \mathrm{FCT} =
  \begin{cases}
    \min \mathcal{T}_0, & \mathcal{T}_0 \ne \varnothing,\\
    T+\lambda_{\mathrm{FCT}} n_T, & \mathcal{T}_0 = \varnothing.
  \end{cases}
  \label{eq:appendix_fct}
\end{equation}
In our experiments, $\lambda_{\mathrm{FCT}}=100$ seconds per active
fire. An episode that runs for 300 seconds without clearing and still
has two active fires scores $300+100\times 2=500$ seconds. Natural
fuel exhaustion also clears fires, so we interpret FCT alongside SVR.

\paragraph{Civilian Rescue (T2, T5).}
The \emph{Health Protection Ratio (HPR)} measures the fraction of
civilian health loss prevented. We sum the recorded health of the
initial civilian roster in each run. At the same time points and with
the same free-burning reference as in Eq.~\eqref{eq:appendix_svr},
the corresponding health totals give
\begin{equation}
  \mathrm{HPR} =
  \frac{H_{\mathrm{agent}}-H_{\mathrm{free}}}
       {H_{\mathrm{init}}-H_{\mathrm{free}}}.
  \label{eq:appendix_hpr}
\end{equation}
The \emph{Civilian Rescue Rate (CRR)} is the fraction of civilians
evacuated alive.
Let $\mathcal C$ be the initial civilian roster, and let $h_c(T)$ and
$p_c(T)$ denote civilian $c$'s final health and horizontal position.
For the designated safe zone $\mathcal S$, including its boundary,
\begin{equation}
  \mathrm{CRR} =
  \frac{\left|\{c\in\mathcal C \mid h_c(T)>0,\ p_c(T)\in\mathcal S\}\right|}
       {|\mathcal C|}.
  \label{eq:appendix_crr}
\end{equation}
Suppressing a fire near a civilian may preserve their health even
when they remain outside the safe zone at the end of the episode.

\paragraph{Exploration (T3, visual T4, and T5).}
\emph{Exploration Coverage (EC)} measures the fraction of eligible
object instances seen at least once. The initial scene inventory
$\mathcal E_{\mathrm{all}}$ contains instances marked as readable in
the benchmark's object catalog and excludes civilians. Let
$\mathcal E_{\mathrm{obs}}$ be the union of visible object IDs over
all robots and all valid observation samples in $[0,T]$. Then
\begin{equation}
  \mathrm{EC} =
  \frac{|\mathcal E_{\mathrm{obs}}\cap\mathcal E_{\mathrm{all}}|}
       {|\mathcal E_{\mathrm{all}}|}.
  \label{eq:appendix_ec}
\end{equation}
In visual trials the evaluator records visibility independently of
the detector, so a visible object can count even when the detector
misses it.

\paragraph{Team Performance (T4, T5).}
\emph{Team Gain (TG)} measures the change in property preservation
when an $N$-robot team performs the task. For each matched team and
single-robot pair in Appendix~\ref{appendix:evaluation_protocol},
\begin{equation}
  \mathrm{TG} = \mathrm{SVR}_{N}-\mathrm{SVR}_{1}.
  \label{eq:appendix_tg}
\end{equation}
This difference includes the effects of parallel execution, role
composition, and coordination on property preservation.

We report SVR, HPR, CRR, and EC as percentages, FCT in seconds, and
TG in percentage points. Ratios require a positive denominator.
We mark a metric as unavailable if its denominator fails this condition
or a required reference, state, or sample is missing.
Appendix~\ref{appendix:complete_results} gives the aggregation across
repetitions and scenarios.

\section{FireAgents Implementation Details}

\subsection{Decision Context, Prompts, and Memory}
\label{appendix:agent_context_memory}

Each episode's system message specifies the mission, room anchors, robot
roles (Appendix~\ref{appendix:action_space}), observation interface, and
controller. At each decision, \agent{} adds the current observation, robot
state, signals, teammate messages, time remaining, memory, and previous
execution feedback to the user message. Visual inputs follow
Appendix~\ref{appendix:observation_interfaces}.

\paragraph{Decision instructions.}

\begin{table}[htb]
\centering
\small
\setlength{\tabcolsep}{0pt}
\renewcommand{\arraystretch}{1.13}
\caption{\textbf{Mode-specific instructions, paraphrased.}
Each episode uses one observation interface and either single-robot (SR),
decentralized multi-robot (DMR), or centralized multi-robot (CMR) control.}
\label{tab:fireagents_mode_instructions}
\begin{tabular}{@{}>{\raggedright\arraybackslash}p{0.23\linewidth}@{\hspace{0.035\linewidth}}>{\raggedright\arraybackslash}p{0.735\linewidth}@{}}
\toprule
\textbf{Mode} & \textbf{Instruction given to the model} \\
\midrule
Global Symbolic & Choose named entity targets from the latest complete scene snapshot. \\
Local Symbolic & Check when a target was last seen. Treat the contents of unseen rooms as unknown. \\
RGB & Specify the frame and pixel from a current, actionable original RGB view. Use earlier views as context. \\
RGB+DS & Check detections against RGB. Use a reliable, currently grounded \texttt{target\_ref}. Otherwise select a pixel as in RGB, or change viewpoint. \\
\midrule
SR & Choose one action. Set \texttt{communication} to \texttt{null}. \\
DMR & Choose one action for your robot. Submit any message you want to broadcast. Use \texttt{null} when communication is disabled or you have nothing to send. \\
CMR & Assign each robot a new action or \texttt{continue} in \texttt{robot\_actions}. \\
\bottomrule
\end{tabular}
\end{table}

We paraphrase the shared guidance below. Table~\ref{tab:fireagents_mode_instructions}
gives the additional instructions for each observation interface and controller.
\begin{quote}
\small\itshape
Protect civilian lives first and limit property damage. Weigh urgency,
travel time, and the age of each observation. Before a close-range action,
approach if the target is farther than about 2\,m or its proximity is
unclear. Check the effect in subsequent observations or feedback. At the
safe zone, confirm that escorted civilians have arrived before releasing
them, then verify that they have stopped following.
\end{quote}

Local CMR pools each robot's latest complete four-view observation.
The center checks proximity for the commanded robot.
See Appendix~\ref{appendix:observation_interfaces} for visual target formats
and validity rules.

\paragraph{Worked example.}
Figure~\ref{fig:fireagents_worked_decision} follows FR-1 from observing a
civilian to announcing its next action and recording the outcome.

\par\smallskip\noindent
\begin{minipage}{\columnwidth}
\captionsetup{type=figure}
\centering
\small
\setlength{\fboxsep}{4pt}
\setlength{\fboxrule}{0.35pt}
\newcommand{\fgmessage}[3]{%
  \noindent\fcolorbox{black!40}{#1}{%
    \parbox{\dimexpr\linewidth-2\fboxsep-2\fboxrule\relax}{%
      \raggedright\textsf{\textbf{#2}}\par\smallskip #3}}%
  \par}
\fgmessage{black!3}{System instruction}{%
  For a close-range interaction, approach first if the target is farther
  than about 2\,m or its proximity is uncertain.}
\smallskip
\fgmessage{black!3}{Runtime input (decision 2)}{%
  FR-1 reached Room1 at $(17,-0.7)$\,m. Its front view recorded civilian
  \texttt{weiguo} at $(21,-0.7)$\,m. FR-1 has no followers.\par
  \emph{Current note.} “Inspect Room1 for civilians.”}
\smallskip
\fgmessage{cvprblue!6}{Model response (FR-1)}{%
  \emph{Basis.} The civilian was observed 4\,m away. Approach before asking
  them to follow.\par
  \emph{Action.} \texttt{navigate} / \texttt{entity} $\to$ \texttt{weiguo}\par
  \emph{Message (useful).} “I plan to approach weiguo in Room1.”\par
  \emph{New note.} “weiguo seen in Room1. Approach planned. Evacuation
  unconfirmed.”}
\smallskip
\parbox{\linewidth}{\raggedright
  \textsf{\textbf{Execution feedback.}} The controller stores the note on
  acceptance. When navigation ends at $(20,-0.7)$\,m, it adds the executor's
  \texttt{target\_reached} result to decision~2.}
\caption{\textbf{One FireAgents decision.}
Condensed, illustrative Local Symbolic exchange under DMR with communication
enabled. The message announces an intended approach. Subsequent navigation
feedback confirms arrival at the target.}
\label{fig:fireagents_worked_decision}
\end{minipage}\par

\paragraph{Responses and episode memory.}
Every JSON reply includes a one-sentence \texttt{decision\_basis} and a
\texttt{memory\_update}. SR and DMR add \texttt{action} and
\texttt{communication}. CMR adds \texttt{robot\_actions}.

Each SR or DMR robot keeps its last $K$ accepted decisions and an episode
note of up to $L$ characters (Appendix~\ref{appendix:model_configuration}).
CMR keeps one team note and one history, with outcomes indexed by robot.
In an accepted reply, a \texttt{null} note update keeps the previous text,
an empty string clears it, and a new string replaces it. Execution
feedback is added to the decision while it remains in the window.
Appendix~\ref{appendix:agent_execution_coordination} describes how the
controllers handle events that arrive during inference.

In Local Symbolic SR and DMR, entity records update independently of reply
acceptance. By default, they capture each entity's first sighting per action
and changes in object burning, civilian fire exposure, or safety. Stored positions
and health can therefore predate the latest observation. Individual visual
agents track earlier sightings through images, notes, and decision history.

\subsection{Execution and Event-Triggered Coordination}
\label{appendix:agent_execution_coordination}

\begin{figure*}[ht]
  \centering
  \includegraphics[width=\textwidth]{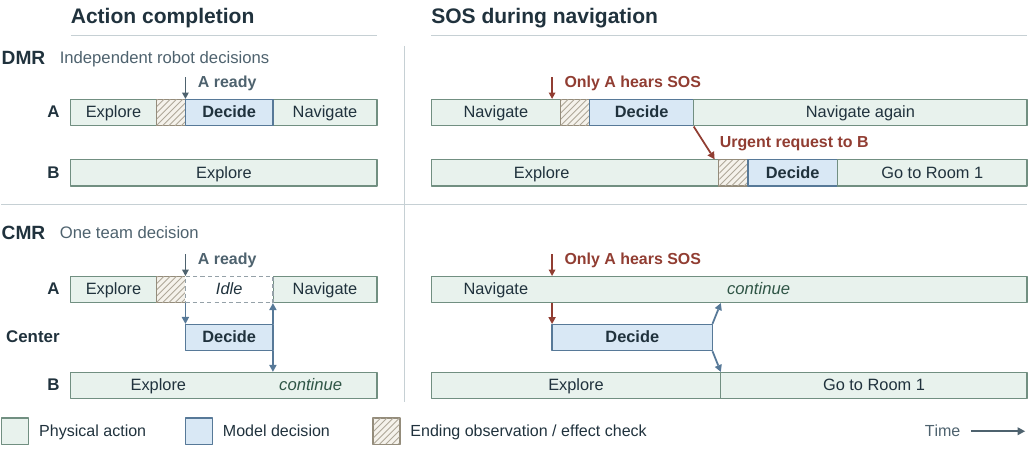}
  \caption{\textbf{Asynchronous scheduling and SOS-triggered replanning.}
  The columns show independent examples. At left, A completes its
  post-action checks while B keeps exploring. At right, only A hears an SOS while navigating in
  Room~1. In both modes B goes to check the room. In DMR, B decides
  after receiving A's urgent request. In CMR, the center redirects B
  and leaves A's navigation running. Time runs left to right under the
  default real-time protocol (Appendix~\ref{appendix:interaction_timing}).
  Bar lengths illustrate event order and overlap.}
  \label{fig:agent_event_timeline}
\end{figure*}

\paragraph{Action completion and scheduling.}
After a command returns, the robot takes a final observation, refreshes
its state, and checks the action's effect when possible.
For example, a visual agent can receive \texttt{completed} for a spray
command while its effect remains \texttt{unknown}. The agent uses
subsequent RGB views to judge whether the target is still burning.
In SR and DMR, the robot then chooses its next action. In CMR, any
robot finishing these checks can trigger a team decision while the
others keep working. The center gives each robot a new action or
\texttt{continue}. With \texttt{continue}, a running action proceeds
and an idle robot stays idle.

\paragraph{Replanning during execution.}
When an SR or DMR robot hears a new civilian SOS
(Appendix~\ref{appendix:victim_model}), its controller requests an
interruption, takes an ending observation, and replans. The receiver
filters repeat SOS samples it can identify. With DMR communication
enabled, a robot can send an \emph{urgent} message to request help.
This can interrupt the recipient's action and trigger its own decision.
The recipient reads \emph{important} and \emph{useful} messages at its
next decision. In CMR, the SOS reaches the center through the receiving
robot. Ongoing actions continue during central inference, and the center
can replace any robot's action with a new assignment.
Figure~\ref{fig:agent_event_timeline} illustrates both responses to the
same call.

\paragraph{Events during inference.}
When a new SOS arrives during inference in SR or DMR, the controller
discards the pending reply when it returns and uses the SOS in its next
decision. An urgent teammate message in DMR triggers the same response.
The controller uses only accepted replies to update its note or send messages.
In CMR, the center replaces the proposed command for each robot that
received a new SOS with \texttt{continue}, dispatches the other valid
commands, and retains the previous team note. It then replans using the
new signal. A robot that finishes an action during central inference
can still receive a valid command from that reply.

\section{Experimental Setup and Reproducibility}

\subsection{Evaluation Scenarios and Task Assignments}
\label{appendix:experiment_scenarios}

All six scenarios share a central safe area and two residential wings
with seven rooms each (Fig.~\ref{fig:appendix_scenario_floor_plan}).
The layout ID in Table~\ref{tab:appendix_evaluation_scenarios} identifies
which of four furnishing variants is used across these rooms.
Before formal model comparisons, we selected the scenarios to cover three
patterns of fire spread observed in development runs without robots.
In S1 and S2, furniture fires stayed within the ignition rooms.
S2 had more objects burning at its peak and greater property loss than S1.
In S3 and S4, fire reached other rooms later in the episode.
In S5 and S6, it reached other rooms earlier and spread more widely.

T1, T3, and T4 use the civilian-free version (F) of each scenario.
T2 and T5 use the rescue version (R), which adds two civilians while
keeping the layout and ignition rooms fixed. For example, S5R and S6R
both use layout~3, with four ignition rooms and one civilian in each wing.
In S5R, the ignition rooms span both wings and include R8, where the
left-wing civilian starts. In S6R, all four ignition rooms are in the
right wing, and both civilians start in other rooms (R5 and R12).
Each episode begins with a 5-second preburn followed by 300 seconds of
evaluation. All 12 configurations are available in
\texttt{benchmark\_core6/} (Appendix~\ref{appendix:dataset_release}).

\begingroup
\ifdefined\nolinenumbers\nolinenumbers\fi
\noindent\begin{minipage}{\columnwidth}
  \centering
  \begingroup
  \captionsetup{type=figure}
  \includegraphics[width=\columnwidth]{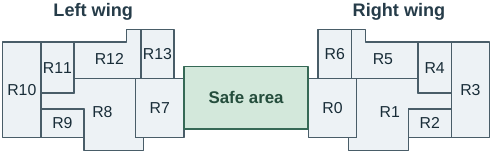}
  \caption{\textbf{Floor plan shared by S1--S6.}
  Room outlines follow the simulator coordinates. The safe area is
  enlarged in this schematic to meet R7 and R0. A single firefighting
  robot (FR) starts at the center. Four FRs start at four fixed
  positions around the center.}
  \label{fig:appendix_scenario_floor_plan}
  \endgroup
  \par\medskip
  \captionsetup{type=table}
  \small
  \setlength{\tabcolsep}{2.5pt}
  \renewcommand{\arraystretch}{1.04}
  \begin{tabular*}{\columnwidth}{@{\extracolsep{\fill}}lcll@{}}
    \toprule
    \textbf{Scenario} & \textbf{Layout ID} & \textbf{Ignition rooms}
      & \textbf{Civilian starts (R)} \\
    \midrule
    \multicolumn{4}{@{}l}{\emph{Fire confined to the ignition rooms}} \\
    S1 & 3 & R2, R3           & R2, R3 \\
    S2 & 1 & R0, R2           & R0, R2 \\
    \addlinespace[3pt]
    \multicolumn{4}{@{}l}{\emph{Later spread to other rooms}} \\
    S3 & 4 & R1, R3, R5       & R2, R10 \\
    S4 & 2 & R0, R5, R8, R10  & R0, R5 \\
    \addlinespace[3pt]
    \multicolumn{4}{@{}l}{\emph{Earlier, more extensive spread to other rooms}} \\
    S5 & 3 & R1, R5, R8, R10  & R0, R8 \\
    S6 & 3 & R0, R1, R2, R4   & R5, R12 \\
    \bottomrule
  \end{tabular*}
  \caption{\textbf{Initial conditions for the six evaluation scenarios.}
  Groups describe furniture fire spread in free-burning development
  runs. The last column gives one starting room per civilian in the
  rescue version.}
  \label{tab:appendix_evaluation_scenarios}
\end{minipage}\par
\endgroup

\begin{samepage}
\subsection{Model and Runtime Configurations}
\label{appendix:model_configuration}

In the main experiments, we requested the settings below
(\texttt{reasoning\_effort} for GPT, \texttt{thinking.type} for Claude).
\emph{Default} leaves the reasoning setting to the provider.
\par\smallskip
{\small
\noindent\begin{tabular*}{\columnwidth}{@{\extracolsep{\fill}}lll@{}}
  \toprule
  \textbf{Model} & \textbf{Symbolic} & \textbf{Visual} \\
  \midrule
  GPT-6 Luna           & \texttt{none}     & \texttt{none} \\
  GPT-6.1 Terra        & \texttt{low}      & \texttt{none} \\
  Claude Haiku 5.5     & \texttt{disabled} & \texttt{disabled} \\
  DeepSeek V4 Flash S2 & Default           & Default \\
  GLM-5.3-Flash        & Default           & Default \\
  \bottomrule
\end{tabular*}\par}
\smallskip
\noindent \agent{} allowed 8,192 output tokens per model call. Memory held
the last three accepted decisions and a note of up to 2,000 characters
(Appendix~\ref{appendix:agent_context_memory}). The main T4 and T5 teams
used four FRs under DMR with communication enabled.
\par
\end{samepage}

\subsection{Timing in Visual Trials}
\label{appendix:interfaces}
\label{appendix:visual_trial_protocols}

In the main RGB and RGB+DS trials, a decision still pending after
20 seconds of active environment time triggered a simulator pause
request (Appendix~\ref{appendix:interaction_timing}).

\subsection{Experiment Records and Matched References}
\label{appendix:experiment_references}

Each main model condition comprised three runs per scenario, yielding
18 runs across S1--S6. Following Appendix~\ref{appendix:evaluation_protocol},
we reused the symbolic T1 and visual T3 trials in
Appendix~\ref{appendix:complete_results} for the corresponding T4 teams.
The single-FR T5 trials in Appendix~\ref{appendix:visual_results} provide
the local symbolic and visual references for joint suppression and rescue.

The main paper reuses trials from four-FR DMR teams with broadcasts
enabled. Tables~2 and~4 share the 18 symbolic trials for each of three
conditions (DeepSeek on T4 and T5, and GLM on T5). Tables~3 and~4 share
all 18 RGB trials and all 18 RGB+DS trials for DeepSeek on T5.

\subsection{Human Evaluation Procedure}
\label{appendix:human}

\paragraph{Participants and tasks.}
All 15 participants gave informed consent after reading a questionnaire that explained the purpose of the study, the tasks they would perform, and how their trial results would be used. Participants selected robot actions in simulated firefighting and rescue scenarios that contained no sensitive content. We report results in aggregate without identifying individual participants.
Three participants performed T3 individually, each controlling one
firefighting robot (FR) to suppress fires using local RGB observations.
The remaining 12 formed three teams of four for T5. Each team member
controlled one FR, and teams were instructed to prioritize civilian
safety while suppressing fires. Each T3 participant and each T5 team
completed one trial in each of the six scenarios (S1--S6).
This yielded 18 individual trials for T3 and 18 team trials for T5.
We shuffled the scenario order in both tasks to reduce the influence of
practice on comparisons across scenarios.
Table~\ref{tab:appendix_h1_human_results} reports the human results for each scenario.

\begin{figure}[t]
  \centering
  \includegraphics[width=\columnwidth]{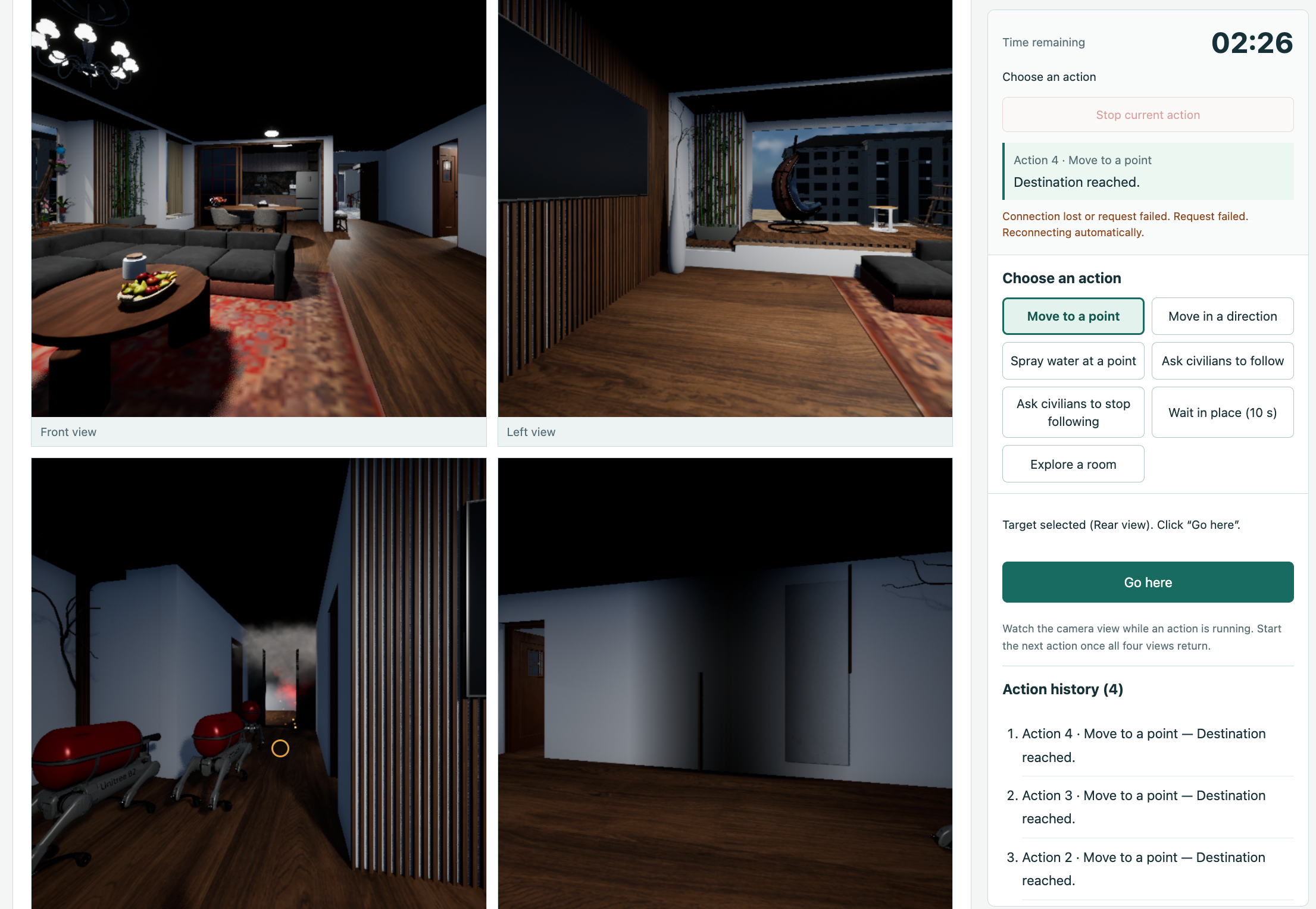}
  \caption{\textbf{Human evaluation interface.}
  The left panel shows the human subject's front and left views in the top
  row and its rear and right views in the bottom row. The right
  panel shows the remaining time, available actions, and execution
  feedback. Here, point navigation is selected, and the participant
  can click a destination in any of the four views.}
  \label{fig:appendix_human_ui}
\end{figure}

\paragraph{Practice and interface.}
Before the scored trials, participants had access to separate practice
scenarios for individual fire suppression and team-based suppression
and rescue. At each decision point, the interface displayed four RGB
views from the participant's robot (Fig.~\ref{fig:appendix_human_ui}).
Participants selected an action and could click directly in any view
to specify a navigation destination or aim the water jet. For room
exploration, they could select a room on a static floor plan. During
action execution, the interface displayed a single updating camera
view. All four views returned when the action ended.

\paragraph{Team communication.}
During scored T5 trials, teammates sat together, and each participant
was permitted to view only their own screen. They could report the
locations of fires and civilians aloud and agree on who would suppress
a fire or escort a civilian. The simulation continued during these
discussions. We omit Team Gain (TG) for human T5 because the individual
trials covered suppression only and therefore provided no matched
single-robot reference for the joint suppression and rescue task.

\onecolumn
\section{Supplementary Experimental Results}

\subsection{Complete Results and Dispersion Statistics}
\label{appendix:complete_results}
\begingroup
\setlength{\parindent}{0pt}
\setlength{\parskip}{0pt}
\setlength{\tabcolsep}{2pt}
\setlength{\aboverulesep}{1.1pt}
\setlength{\belowrulesep}{1.1pt}
\hfuzz=0.1pt
\vfuzz=0.1pt
\captionsetup{hypcap=false,font=footnotesize,labelfont=bf,position=top,justification=raggedright,singlelinecheck=false,skip=2.2pt}
\newcommand{\HOneScenarioScore}[2]{%
  \makebox[22pt][r]{#1}\kern0.8pt\textpm\kern0.8pt%
  \makebox[22pt][l]{\textcolor{black!76}{#2}}%
}
{\fontsize{8.3}{9.5}\selectfont
Each scenario gives mean \textpm{} sample SD (3 trials). Overall gives the unweighted
mean of the six means \textpm{} the root mean square of the six SDs.\par}
\vspace{4pt}
\fontsize{8.2}{8.5}\selectfont
\captionof{table}{Per-scenario symbolic results (main-paper Table~2).
Bold marks each column's best mean per task and metric, including ties (FCT $\downarrow$, others $\uparrow$).}
\label{tab:appendix_h1_symbolic_results}
\noindent\begin{tabular*}{\dimexpr\linewidth-10pt\relax}{@{}ll@{\extracolsep{\fill}}*{7}{r}@{}}
\toprule
\textbf{Model} & \textbf{Metric} & \multicolumn{6}{c}{\textbf{Scenario}} & {} \\
\midrule
\multicolumn{2}{@{}l}{\textbf{T1. Global symbolic}} & \makebox[51.4pt][c]{\textbf{S1F}} & \makebox[51.4pt][c]{\textbf{S2F}} & \makebox[51.4pt][c]{\textbf{S3F}} & \makebox[51.4pt][c]{\textbf{S4F}} & \makebox[51.4pt][c]{\textbf{S5F}} & \makebox[51.4pt][c]{\textbf{S6F}} & \makebox[51.4pt][c]{\textbf{Overall}} \\
GPT-6.1 & SVR (\%) & \HOneScenarioScore{\textbf{37.29}}{7.68} & \HOneScenarioScore{\textbf{58.69}}{0.38} & \HOneScenarioScore{\textbf{33.80}}{20.15} & \HOneScenarioScore{52.38}{32.75} & \HOneScenarioScore{\textbf{39.40}}{8.67} & \HOneScenarioScore{3.80}{2.24} & \HOneScenarioScore{\textbf{37.56}}{16.42} \\
 & FCT (s) & \HOneScenarioScore{\textbf{146.2}}{32.2} & \HOneScenarioScore{\textbf{242.2}}{0.0} & \HOneScenarioScore{1633.3}{351.2} & \HOneScenarioScore{1766.7}{1457.2} & \HOneScenarioScore{666.7}{461.9} & \HOneScenarioScore{1666.7}{57.7} & \HOneScenarioScore{1020.3}{640.9} \\
\addlinespace[0.6pt]
GPT-6 & SVR (\%) & \HOneScenarioScore{25.49}{27.93} & \HOneScenarioScore{39.34}{28.37} & \HOneScenarioScore{18.32}{5.59} & \HOneScenarioScore{\textbf{58.36}}{29.34} & \HOneScenarioScore{20.55}{15.67} & \HOneScenarioScore{7.54}{11.06} & \HOneScenarioScore{28.27}{21.77} \\
 & FCT (s) & \HOneScenarioScore{440.0}{277.1} & \HOneScenarioScore{661.4}{556.1} & \HOneScenarioScore{2266.7}{288.7} & \HOneScenarioScore{1233.3}{1270.2} & \HOneScenarioScore{1266.7}{750.6} & \HOneScenarioScore{1333.3}{416.3} & \HOneScenarioScore{1200.2}{685.5} \\
\addlinespace[0.6pt]
Claude & SVR (\%) & \HOneScenarioScore{26.84}{6.08} & \HOneScenarioScore{56.18}{3.43} & \HOneScenarioScore{7.81}{2.83} & \HOneScenarioScore{44.43}{0.50} & \HOneScenarioScore{15.93}{24.39} & \HOneScenarioScore{\textbf{11.06}}{6.93} & \HOneScenarioScore{27.04}{10.80} \\
 & FCT (s) & \HOneScenarioScore{303.4}{257.1} & \HOneScenarioScore{312.1}{80.1} & \HOneScenarioScore{2533.3}{416.3} & \HOneScenarioScore{833.3}{115.5} & \HOneScenarioScore{1666.7}{1097.0} & \HOneScenarioScore{\textbf{933.3}}{251.7} & \HOneScenarioScore{1097.0}{504.3} \\
\addlinespace[0.6pt]
DeepSeek & SVR (\%) & \HOneScenarioScore{30.59}{5.57} & \HOneScenarioScore{50.17}{26.05} & \HOneScenarioScore{30.19}{4.91} & \HOneScenarioScore{37.77}{27.46} & \HOneScenarioScore{39.32}{4.48} & \HOneScenarioScore{7.02}{8.15} & \HOneScenarioScore{32.51}{16.20} \\
 & FCT (s) & \HOneScenarioScore{182.2}{0.0} & \HOneScenarioScore{1100.0}{1126.9} & \HOneScenarioScore{\textbf{1100.0}}{0.0} & \HOneScenarioScore{1466.7}{1159.0} & \HOneScenarioScore{\textbf{363.1}}{64.0} & \HOneScenarioScore{1433.3}{378.6} & \HOneScenarioScore{\textbf{940.9}}{678.3} \\
\addlinespace[0.6pt]
GLM & SVR (\%) & \HOneScenarioScore{26.62}{4.65} & \HOneScenarioScore{54.13}{0.22} & \HOneScenarioScore{0.73}{1.21} & \HOneScenarioScore{43.72}{0.78} & \HOneScenarioScore{9.36}{2.22} & \HOneScenarioScore{3.56}{3.17} & \HOneScenarioScore{23.02}{2.54} \\
 & FCT (s) & \HOneScenarioScore{168.9}{23.0} & \HOneScenarioScore{400.0}{0.0} & \HOneScenarioScore{2533.3}{57.7} & \HOneScenarioScore{\textbf{766.7}}{57.7} & \HOneScenarioScore{2200.0}{100.0} & \HOneScenarioScore{1466.7}{351.2} & \HOneScenarioScore{1255.9}{153.0} \\
\midrule
\multicolumn{2}{@{}l}{\textbf{T2. Global symbolic}} & \makebox[51.4pt][c]{\textbf{S1R}} & \makebox[51.4pt][c]{\textbf{S2R}} & \makebox[51.4pt][c]{\textbf{S3R}} & \makebox[51.4pt][c]{\textbf{S4R}} & \makebox[51.4pt][c]{\textbf{S5R}} & \makebox[51.4pt][c]{\textbf{S6R}} & \makebox[51.4pt][c]{\textbf{Overall}} \\
GPT-6.1 & SVR (\%) & \HOneScenarioScore{\textbf{34.04}}{11.87} & \HOneScenarioScore{\textbf{53.80}}{1.19} & \HOneScenarioScore{4.30}{2.45} & \HOneScenarioScore{31.44}{21.52} & \HOneScenarioScore{3.19}{2.86} & \HOneScenarioScore{\textbf{6.47}}{1.56} & \HOneScenarioScore{\textbf{22.21}}{10.18} \\
 & FCT (s) & \HOneScenarioScore{\textbf{155.9}}{22.9} & \HOneScenarioScore{\textbf{466.7}}{57.7} & \HOneScenarioScore{2400.0}{556.8} & \HOneScenarioScore{1233.3}{577.4} & \HOneScenarioScore{2233.3}{115.5} & \HOneScenarioScore{\textbf{1200.0}}{100.0} & \HOneScenarioScore{\textbf{1281.5}}{334.3} \\
 & HPR (\%) & \HOneScenarioScore{23.13}{29.57} & \HOneScenarioScore{\textbf{59.57}}{7.10} & \HOneScenarioScore{91.35}{0.19} & \HOneScenarioScore{\textbf{74.41}}{5.13} & \HOneScenarioScore{\textbf{88.95}}{0.11} & \HOneScenarioScore{\textbf{88.64}}{0.05} & \HOneScenarioScore{\textbf{71.01}}{12.59} \\
 & CRR (\%) & \HOneScenarioScore{66.67}{28.87} & \HOneScenarioScore{\textbf{100.00}}{0.00} & \HOneScenarioScore{\textbf{100.00}}{0.00} & \HOneScenarioScore{\textbf{100.00}}{0.00} & \HOneScenarioScore{\textbf{100.00}}{0.00} & \HOneScenarioScore{\textbf{100.00}}{0.00} & \HOneScenarioScore{\textbf{94.44}}{11.79} \\
\addlinespace[0.6pt]
GPT-6 & SVR (\%) & \HOneScenarioScore{27.15}{26.85} & \HOneScenarioScore{35.50}{30.40} & \HOneScenarioScore{\textbf{12.97}}{19.25} & \HOneScenarioScore{30.02}{5.53} & \HOneScenarioScore{6.14}{3.26} & \HOneScenarioScore{3.90}{1.66} & \HOneScenarioScore{19.28}{18.53} \\
 & FCT (s) & \HOneScenarioScore{385.4}{222.3} & \HOneScenarioScore{900.0}{692.8} & \HOneScenarioScore{2300.0}{346.4} & \HOneScenarioScore{1433.3}{1011.6} & \HOneScenarioScore{2166.7}{57.7} & \HOneScenarioScore{1700.0}{173.2} & \HOneScenarioScore{1480.9}{533.2} \\
 & HPR (\%) & \HOneScenarioScore{\textbf{37.21}}{5.71} & \HOneScenarioScore{44.61}{3.82} & \HOneScenarioScore{\textbf{91.42}}{0.09} & \HOneScenarioScore{59.76}{17.85} & \HOneScenarioScore{84.66}{2.46} & \HOneScenarioScore{88.28}{0.61} & \HOneScenarioScore{67.66}{7.88} \\
 & CRR (\%) & \HOneScenarioScore{\textbf{100.00}}{0.00} & \HOneScenarioScore{66.67}{28.87} & \HOneScenarioScore{83.33}{28.87} & \HOneScenarioScore{83.33}{28.87} & \HOneScenarioScore{\textbf{100.00}}{0.00} & \HOneScenarioScore{50.00}{0.00} & \HOneScenarioScore{80.56}{20.41} \\
\addlinespace[0.6pt]
Claude & SVR (\%) & \HOneScenarioScore{-0.00}{0.03} & \HOneScenarioScore{19.87}{28.68} & \HOneScenarioScore{3.49}{0.75} & \HOneScenarioScore{25.41}{7.01} & \HOneScenarioScore{2.35}{1.12} & \HOneScenarioScore{-1.72}{4.20} & \HOneScenarioScore{8.23}{12.19} \\
 & FCT (s) & \HOneScenarioScore{700.0}{0.0} & \HOneScenarioScore{1066.7}{550.8} & \HOneScenarioScore{\textbf{2166.7}}{208.2} & \HOneScenarioScore{1533.3}{230.9} & \HOneScenarioScore{2300.0}{0.0} & \HOneScenarioScore{1500.0}{264.6} & \HOneScenarioScore{1544.4}{279.9} \\
 & HPR (\%) & \HOneScenarioScore{29.67}{1.56} & \HOneScenarioScore{41.69}{0.13} & \HOneScenarioScore{89.26}{1.05} & \HOneScenarioScore{39.74}{7.13} & \HOneScenarioScore{84.69}{2.13} & \HOneScenarioScore{81.10}{5.89} & \HOneScenarioScore{61.03}{3.95} \\
 & CRR (\%) & \HOneScenarioScore{\textbf{100.00}}{0.00} & \HOneScenarioScore{50.00}{0.00} & \HOneScenarioScore{\textbf{100.00}}{0.00} & \HOneScenarioScore{66.67}{28.87} & \HOneScenarioScore{\textbf{100.00}}{0.00} & \HOneScenarioScore{\textbf{100.00}}{0.00} & \HOneScenarioScore{86.11}{11.79} \\
\addlinespace[0.6pt]
DeepSeek & SVR (\%) & \HOneScenarioScore{4.50}{5.32} & \HOneScenarioScore{23.08}{25.94} & \HOneScenarioScore{9.02}{0.76} & \HOneScenarioScore{\textbf{34.34}}{13.25} & \HOneScenarioScore{\textbf{24.79}}{16.97} & \HOneScenarioScore{2.33}{1.96} & \HOneScenarioScore{16.34}{13.96} \\
 & FCT (s) & \HOneScenarioScore{633.3}{57.7} & \HOneScenarioScore{1100.0}{600.0} & \HOneScenarioScore{2700.0}{360.6} & \HOneScenarioScore{\textbf{1166.7}}{378.6} & \HOneScenarioScore{\textbf{833.3}}{577.4} & \HOneScenarioScore{1566.7}{230.9} & \HOneScenarioScore{1333.3}{413.0} \\
 & HPR (\%) & \HOneScenarioScore{20.07}{9.99} & \HOneScenarioScore{41.65}{0.07} & \HOneScenarioScore{89.10}{0.18} & \HOneScenarioScore{33.78}{4.15} & \HOneScenarioScore{76.83}{7.37} & \HOneScenarioScore{83.14}{1.02} & \HOneScenarioScore{57.43}{5.36} \\
 & CRR (\%) & \HOneScenarioScore{\textbf{100.00}}{0.00} & \HOneScenarioScore{50.00}{0.00} & \HOneScenarioScore{66.67}{28.87} & \HOneScenarioScore{50.00}{0.00} & \HOneScenarioScore{66.67}{28.87} & \HOneScenarioScore{83.33}{28.87} & \HOneScenarioScore{69.44}{20.41} \\
\addlinespace[0.6pt]
GLM & SVR (\%) & \HOneScenarioScore{5.23}{1.45} & \HOneScenarioScore{26.38}{23.71} & \HOneScenarioScore{0.65}{1.14} & \HOneScenarioScore{34.07}{14.82} & \HOneScenarioScore{4.04}{5.28} & \HOneScenarioScore{3.00}{2.59} & \HOneScenarioScore{12.23}{11.69} \\
 & FCT (s) & \HOneScenarioScore{700.0}{0.0} & \HOneScenarioScore{1366.7}{1001.7} & \HOneScenarioScore{2400.0}{173.2} & \HOneScenarioScore{1200.0}{608.3} & \HOneScenarioScore{2066.7}{230.9} & \HOneScenarioScore{1566.7}{57.7} & \HOneScenarioScore{1550.0}{493.3} \\
 & HPR (\%) & \HOneScenarioScore{-1.43}{0.39} & \HOneScenarioScore{30.10}{9.96} & \HOneScenarioScore{79.16}{15.09} & \HOneScenarioScore{45.09}{24.12} & \HOneScenarioScore{52.08}{42.22} & \HOneScenarioScore{42.65}{40.69} & \HOneScenarioScore{41.27}{26.92} \\
 & CRR (\%) & \HOneScenarioScore{50.00}{0.00} & \HOneScenarioScore{66.67}{28.87} & \HOneScenarioScore{\textbf{100.00}}{0.00} & \HOneScenarioScore{66.67}{28.87} & \HOneScenarioScore{83.33}{28.87} & \HOneScenarioScore{66.67}{57.74} & \HOneScenarioScore{72.22}{31.18} \\
\midrule
\multicolumn{2}{@{}l}{\textbf{T3. Local symbolic}} & \makebox[51.4pt][c]{\textbf{S1F}} & \makebox[51.4pt][c]{\textbf{S2F}} & \makebox[51.4pt][c]{\textbf{S3F}} & \makebox[51.4pt][c]{\textbf{S4F}} & \makebox[51.4pt][c]{\textbf{S5F}} & \makebox[51.4pt][c]{\textbf{S6F}} & \makebox[51.4pt][c]{\textbf{Overall}} \\
GPT-6.1 & SVR (\%) & \HOneScenarioScore{\textbf{21.69}}{18.16} & \HOneScenarioScore{44.94}{21.14} & \HOneScenarioScore{9.64}{8.98} & \HOneScenarioScore{26.36}{9.22} & \HOneScenarioScore{9.46}{10.53} & \HOneScenarioScore{3.36}{0.35} & \HOneScenarioScore{19.24}{13.25} \\
 & FCT (s) & \HOneScenarioScore{\textbf{348.8}}{304.4} & \HOneScenarioScore{864.1}{1070.3} & \HOneScenarioScore{2733.3}{288.7} & \HOneScenarioScore{2600.0}{556.8} & \HOneScenarioScore{2400.0}{100.0} & \HOneScenarioScore{1766.7}{57.7} & \HOneScenarioScore{1785.5}{523.6} \\
 & EC (\%) & \HOneScenarioScore{\textbf{80.06}}{4.97} & \HOneScenarioScore{\textbf{69.08}}{5.59} & \HOneScenarioScore{\textbf{79.70}}{0.91} & \HOneScenarioScore{\textbf{69.89}}{1.97} & \HOneScenarioScore{\textbf{60.55}}{5.31} & \HOneScenarioScore{\textbf{75.06}}{0.80} & \HOneScenarioScore{\textbf{72.39}}{3.86} \\
\addlinespace[0.6pt]
GPT-6 & SVR (\%) & \HOneScenarioScore{2.74}{1.43} & \HOneScenarioScore{54.46}{0.40} & \HOneScenarioScore{6.18}{7.73} & \HOneScenarioScore{24.13}{18.54} & \HOneScenarioScore{\textbf{35.98}}{1.10} & \HOneScenarioScore{\textbf{10.47}}{5.79} & \HOneScenarioScore{22.33}{8.57} \\
 & FCT (s) & \HOneScenarioScore{600.0}{0.0} & \HOneScenarioScore{361.4}{66.9} & \HOneScenarioScore{2633.3}{57.7} & \HOneScenarioScore{1900.0}{1053.6} & \HOneScenarioScore{\textbf{433.3}}{57.7} & \HOneScenarioScore{\textbf{1200.0}}{264.6} & \HOneScenarioScore{1188.0}{445.6} \\
 & EC (\%) & \HOneScenarioScore{48.49}{0.56} & \HOneScenarioScore{38.88}{4.29} & \HOneScenarioScore{44.34}{10.85} & \HOneScenarioScore{37.69}{4.64} & \HOneScenarioScore{37.61}{8.46} & \HOneScenarioScore{26.65}{0.26} & \HOneScenarioScore{38.94}{6.19} \\
\addlinespace[0.6pt]
Claude & SVR (\%) & \HOneScenarioScore{0.54}{0.33} & \HOneScenarioScore{53.90}{0.04} & \HOneScenarioScore{\textbf{10.34}}{15.68} & \HOneScenarioScore{\textbf{44.69}}{1.60} & \HOneScenarioScore{25.72}{15.80} & \HOneScenarioScore{6.35}{6.92} & \HOneScenarioScore{\textbf{23.59}}{9.54} \\
 & FCT (s) & \HOneScenarioScore{700.0}{0.0} & \HOneScenarioScore{400.0}{0.0} & \HOneScenarioScore{\textbf{2033.3}}{814.5} & \HOneScenarioScore{\textbf{1100.0}}{360.6} & \HOneScenarioScore{1100.0}{1039.2} & \HOneScenarioScore{1433.3}{461.9} & \HOneScenarioScore{\textbf{1127.8}}{589.7} \\
 & EC (\%) & \HOneScenarioScore{50.85}{19.99} & \HOneScenarioScore{45.22}{11.71} & \HOneScenarioScore{38.50}{11.27} & \HOneScenarioScore{35.29}{0.96} & \HOneScenarioScore{25.07}{2.81} & \HOneScenarioScore{28.80}{14.43} & \HOneScenarioScore{37.29}{12.11} \\
\addlinespace[0.6pt]
DeepSeek & SVR (\%) & \HOneScenarioScore{11.81}{5.01} & \HOneScenarioScore{\textbf{61.06}}{5.60} & \HOneScenarioScore{4.90}{0.81} & \HOneScenarioScore{9.21}{6.54} & \HOneScenarioScore{8.30}{2.69} & \HOneScenarioScore{7.12}{7.92} & \HOneScenarioScore{17.07}{5.32} \\
 & FCT (s) & \HOneScenarioScore{471.4}{222.7} & \HOneScenarioScore{\textbf{297.8}}{88.5} & \HOneScenarioScore{2900.0}{0.0} & \HOneScenarioScore{3266.7}{288.7} & \HOneScenarioScore{2433.3}{288.7} & \HOneScenarioScore{1466.7}{493.3} & \HOneScenarioScore{1806.0}{279.1} \\
 & EC (\%) & \HOneScenarioScore{39.66}{16.96} & \HOneScenarioScore{37.35}{10.57} & \HOneScenarioScore{60.52}{0.65} & \HOneScenarioScore{45.90}{16.35} & \HOneScenarioScore{24.04}{7.62} & \HOneScenarioScore{40.01}{14.23} & \HOneScenarioScore{41.25}{12.43} \\
\addlinespace[0.6pt]
GLM & SVR (\%) & \HOneScenarioScore{2.09}{1.24} & \HOneScenarioScore{54.06}{0.08} & \HOneScenarioScore{3.08}{0.39} & \HOneScenarioScore{33.83}{14.28} & \HOneScenarioScore{15.99}{17.93} & \HOneScenarioScore{3.39}{1.96} & \HOneScenarioScore{18.74}{9.41} \\
 & FCT (s) & \HOneScenarioScore{700.0}{0.0} & \HOneScenarioScore{400.0}{0.0} & \HOneScenarioScore{2733.3}{115.5} & \HOneScenarioScore{1366.7}{461.9} & \HOneScenarioScore{1666.7}{1101.5} & \HOneScenarioScore{1500.0}{264.6} & \HOneScenarioScore{1394.4}{501.7} \\
 & EC (\%) & \HOneScenarioScore{39.44}{8.10} & \HOneScenarioScore{43.53}{13.64} & \HOneScenarioScore{50.26}{20.13} & \HOneScenarioScore{55.13}{7.21} & \HOneScenarioScore{32.52}{9.32} & \HOneScenarioScore{23.53}{5.47} & \HOneScenarioScore{40.74}{11.73} \\
\midrule
\multicolumn{2}{@{}l}{\textbf{T4. Global symbolic}} & \makebox[51.4pt][c]{\textbf{S1F}} & \makebox[51.4pt][c]{\textbf{S2F}} & \makebox[51.4pt][c]{\textbf{S3F}} & \makebox[51.4pt][c]{\textbf{S4F}} & \makebox[51.4pt][c]{\textbf{S5F}} & \makebox[51.4pt][c]{\textbf{S6F}} & \makebox[51.4pt][c]{\textbf{Overall}} \\
GPT-6.1 & SVR (\%) & \HOneScenarioScore{\textbf{50.95}}{4.97} & \HOneScenarioScore{64.85}{2.55} & \HOneScenarioScore{66.14}{8.80} & \HOneScenarioScore{\textbf{92.50}}{1.68} & \HOneScenarioScore{\textbf{50.99}}{16.45} & \HOneScenarioScore{18.10}{9.13} & \HOneScenarioScore{57.26}{8.81} \\
 & FCT (s) & \HOneScenarioScore{\textbf{106.2}}{0.1} & \HOneScenarioScore{242.6}{0.6} & \HOneScenarioScore{435.8}{204.1} & \HOneScenarioScore{\textbf{110.2}}{27.0} & \HOneScenarioScore{528.5}{495.1} & \HOneScenarioScore{833.3}{585.9} & \HOneScenarioScore{376.1}{324.3} \\
 & TG (pp) & \HOneScenarioScore{13.66}{9.87} & \HOneScenarioScore{6.16}{2.28} & \HOneScenarioScore{32.34}{12.73} & \HOneScenarioScore{40.12}{31.81} & \HOneScenarioScore{11.60}{24.01} & \HOneScenarioScore{14.30}{11.12} & \HOneScenarioScore{19.70}{18.15} \\
\addlinespace[0.6pt]
GPT-6 & SVR (\%) & \HOneScenarioScore{41.64}{15.93} & \HOneScenarioScore{64.78}{6.79} & \HOneScenarioScore{\textbf{76.72}}{7.32} & \HOneScenarioScore{92.06}{0.57} & \HOneScenarioScore{49.66}{11.65} & \HOneScenarioScore{\textbf{24.24}}{4.51} & \HOneScenarioScore{\textbf{58.18}}{9.22} \\
 & FCT (s) & \HOneScenarioScore{237.4}{227.4} & \HOneScenarioScore{\textbf{205.1}}{41.9} & \HOneScenarioScore{\textbf{146.6}}{30.7} & \HOneScenarioScore{182.1}{0.0} & \HOneScenarioScore{593.8}{525.1} & \HOneScenarioScore{\textbf{451.7}}{226.9} & \HOneScenarioScore{\textbf{302.8}}{252.2} \\
 & TG (pp) & \HOneScenarioScore{\textbf{16.15}}{23.89} & \HOneScenarioScore{\textbf{25.44}}{26.08} & \HOneScenarioScore{\textbf{58.40}}{12.90} & \HOneScenarioScore{33.70}{29.57} & \HOneScenarioScore{29.11}{26.90} & \HOneScenarioScore{\textbf{16.71}}{7.01} & \HOneScenarioScore{\textbf{29.92}}{22.60} \\
\addlinespace[0.6pt]
Claude & SVR (\%) & \HOneScenarioScore{39.36}{12.26} & \HOneScenarioScore{62.18}{4.47} & \HOneScenarioScore{26.63}{4.65} & \HOneScenarioScore{89.74}{0.83} & \HOneScenarioScore{50.96}{2.75} & \HOneScenarioScore{21.18}{1.41} & \HOneScenarioScore{48.34}{5.81} \\
 & FCT (s) & \HOneScenarioScore{275.4}{281.2} & \HOneScenarioScore{226.8}{14.2} & \HOneScenarioScore{1933.3}{152.8} & \HOneScenarioScore{187.1}{80.3} & \HOneScenarioScore{295.1}{91.7} & \HOneScenarioScore{533.3}{230.9} & \HOneScenarioScore{575.2}{168.7} \\
 & TG (pp) & \HOneScenarioScore{12.52}{14.90} & \HOneScenarioScore{6.00}{6.96} & \HOneScenarioScore{18.82}{6.54} & \HOneScenarioScore{\textbf{45.31}}{1.25} & \HOneScenarioScore{\textbf{35.03}}{25.95} & \HOneScenarioScore{10.12}{8.29} & \HOneScenarioScore{21.30}{13.27} \\
\addlinespace[0.6pt]
DeepSeek & SVR (\%) & \HOneScenarioScore{38.78}{1.86} & \HOneScenarioScore{\textbf{66.16}}{2.63} & \HOneScenarioScore{68.28}{1.48} & \HOneScenarioScore{63.02}{14.07} & \HOneScenarioScore{45.29}{0.86} & \HOneScenarioScore{14.83}{9.28} & \HOneScenarioScore{49.39}{7.04} \\
 & FCT (s) & \HOneScenarioScore{130.2}{30.3} & \HOneScenarioScore{205.5}{15.0} & \HOneScenarioScore{356.7}{74.9} & \HOneScenarioScore{500.0}{173.2} & \HOneScenarioScore{\textbf{275.9}}{14.4} & \HOneScenarioScore{633.3}{321.5} & \HOneScenarioScore{350.3}{152.9} \\
 & TG (pp) & \HOneScenarioScore{8.19}{6.75} & \HOneScenarioScore{15.99}{28.47} & \HOneScenarioScore{38.09}{4.04} & \HOneScenarioScore{25.25}{32.41} & \HOneScenarioScore{5.97}{4.94} & \HOneScenarioScore{7.82}{6.50} & \HOneScenarioScore{16.88}{18.21} \\
\addlinespace[0.6pt]
GLM & SVR (\%) & \HOneScenarioScore{33.18}{3.15} & \HOneScenarioScore{58.08}{3.23} & \HOneScenarioScore{41.43}{18.76} & \HOneScenarioScore{68.91}{23.07} & \HOneScenarioScore{38.50}{20.61} & \HOneScenarioScore{16.42}{1.58} & \HOneScenarioScore{42.76}{14.90} \\
 & FCT (s) & \HOneScenarioScore{149.3}{12.3} & \HOneScenarioScore{242.3}{0.1} & \HOneScenarioScore{1200.0}{608.3} & \HOneScenarioScore{444.7}{189.1} & \HOneScenarioScore{870.1}{1065.1} & \HOneScenarioScore{600.0}{264.6} & \HOneScenarioScore{584.4}{518.1} \\
 & TG (pp) & \HOneScenarioScore{6.55}{6.10} & \HOneScenarioScore{3.95}{3.30} & \HOneScenarioScore{40.70}{18.23} & \HOneScenarioScore{25.19}{23.76} & \HOneScenarioScore{29.14}{18.54} & \HOneScenarioScore{12.86}{3.45} & \HOneScenarioScore{19.73}{14.72} \\
\bottomrule
\end{tabular*}\par
\newpage
{\fontsize{11}{12}\selectfont\bfseries H.1. Complete Results and Dispersion Statistics (continued)\par}
\vspace{5pt}
\fontsize{8.2}{8.5}\selectfont
\captionof*{table}{\textbf{Table~\ref{tab:appendix_h1_symbolic_results} (continued).} Per-scenario symbolic results.}
\noindent\begin{tabular*}{\dimexpr\linewidth-10pt\relax}{@{}ll@{\extracolsep{\fill}}*{7}{r}@{}}
\toprule
\textbf{Model} & \textbf{Metric} & \multicolumn{6}{c}{\textbf{Scenario}} & {} \\
\midrule
\multicolumn{2}{@{}l}{\textbf{T5. Local symbolic}} & \makebox[51.4pt][c]{\textbf{S1R}} & \makebox[51.4pt][c]{\textbf{S2R}} & \makebox[51.4pt][c]{\textbf{S3R}} & \makebox[51.4pt][c]{\textbf{S4R}} & \makebox[51.4pt][c]{\textbf{S5R}} & \makebox[51.4pt][c]{\textbf{S6R}} & \makebox[51.4pt][c]{\textbf{Overall}} \\
GPT-6.1 & SVR (\%) & \HOneScenarioScore{\textbf{20.76}}{2.11} & \HOneScenarioScore{\textbf{56.01}}{1.39} & \HOneScenarioScore{52.40}{2.14} & \HOneScenarioScore{54.28}{8.82} & \HOneScenarioScore{41.34}{3.86} & \HOneScenarioScore{15.01}{4.93} & \HOneScenarioScore{39.97}{4.62} \\
 & FCT (s) & \HOneScenarioScore{\textbf{304.1}}{83.0} & \HOneScenarioScore{600.0}{100.0} & \HOneScenarioScore{1666.7}{230.9} & \HOneScenarioScore{1733.3}{850.5} & \HOneScenarioScore{2100.0}{100.0} & \HOneScenarioScore{1600.0}{173.2} & \HOneScenarioScore{1334.0}{372.7} \\
 & HPR (\%) & \HOneScenarioScore{\textbf{12.84}}{6.01} & \HOneScenarioScore{38.61}{12.31} & \HOneScenarioScore{\textbf{87.10}}{2.12} & \HOneScenarioScore{35.76}{6.11} & \HOneScenarioScore{85.69}{4.60} & \HOneScenarioScore{48.10}{10.72} & \HOneScenarioScore{\textbf{51.35}}{7.80} \\
 & CRR (\%) & \HOneScenarioScore{\textbf{83.33}}{28.87} & \HOneScenarioScore{\textbf{66.67}}{28.87} & \HOneScenarioScore{\textbf{100.00}}{0.00} & \HOneScenarioScore{\textbf{66.67}}{28.87} & \HOneScenarioScore{\textbf{100.00}}{0.00} & \HOneScenarioScore{\textbf{100.00}}{0.00} & \HOneScenarioScore{\textbf{86.11}}{20.41} \\
 & EC (\%) & \HOneScenarioScore{88.96}{1.01} & \HOneScenarioScore{84.66}{4.24} & \HOneScenarioScore{82.92}{0.98} & \HOneScenarioScore{\textbf{82.85}}{1.35} & \HOneScenarioScore{\textbf{71.76}}{2.95} & \HOneScenarioScore{70.38}{6.62} & \HOneScenarioScore{80.25}{3.52} \\
 & TG (pp) & \HOneScenarioScore{\textbf{17.63}}{2.14} & \HOneScenarioScore{3.01}{1.39} & \HOneScenarioScore{42.63}{2.82} & \HOneScenarioScore{\textbf{43.74}}{8.72} & \HOneScenarioScore{\textbf{36.70}}{5.03} & \HOneScenarioScore{8.44}{9.98} & \HOneScenarioScore{\textbf{25.36}}{5.99} \\
\addlinespace[0.6pt]
GPT-6 & SVR (\%) & \HOneScenarioScore{11.81}{7.11} & \HOneScenarioScore{54.39}{0.80} & \HOneScenarioScore{\textbf{58.31}}{9.33} & \HOneScenarioScore{\textbf{54.64}}{9.19} & \HOneScenarioScore{\textbf{41.55}}{0.92} & \HOneScenarioScore{\textbf{25.16}}{4.41} & \HOneScenarioScore{\textbf{40.98}}{6.36} \\
 & FCT (s) & \HOneScenarioScore{469.7}{225.7} & \HOneScenarioScore{400.0}{0.0} & \HOneScenarioScore{\textbf{900.0}}{435.9} & \HOneScenarioScore{\textbf{1433.3}}{642.9} & \HOneScenarioScore{\textbf{833.3}}{57.7} & \HOneScenarioScore{\textbf{866.7}}{288.7} & \HOneScenarioScore{\textbf{817.2}}{351.4} \\
 & HPR (\%) & \HOneScenarioScore{4.93}{0.54} & \HOneScenarioScore{\textbf{40.98}}{0.19} & \HOneScenarioScore{62.01}{28.49} & \HOneScenarioScore{36.85}{0.45} & \HOneScenarioScore{72.04}{3.88} & \HOneScenarioScore{\textbf{73.08}}{11.62} & \HOneScenarioScore{48.31}{12.67} \\
 & CRR (\%) & \HOneScenarioScore{33.33}{28.87} & \HOneScenarioScore{50.00}{0.00} & \HOneScenarioScore{33.33}{28.87} & \HOneScenarioScore{50.00}{0.00} & \HOneScenarioScore{66.67}{28.87} & \HOneScenarioScore{83.33}{28.87} & \HOneScenarioScore{52.78}{23.57} \\
 & EC (\%) & \HOneScenarioScore{87.05}{4.65} & \HOneScenarioScore{72.69}{4.88} & \HOneScenarioScore{68.16}{11.75} & \HOneScenarioScore{80.45}{2.69} & \HOneScenarioScore{57.74}{2.39} & \HOneScenarioScore{65.03}{5.98} & \HOneScenarioScore{71.86}{6.22} \\
 & TG (pp) & \HOneScenarioScore{3.35}{1.14} & \HOneScenarioScore{0.56}{1.07} & \HOneScenarioScore{\textbf{52.74}}{10.30} & \HOneScenarioScore{32.96}{24.11} & \HOneScenarioScore{6.43}{1.11} & \HOneScenarioScore{\textbf{21.44}}{5.51} & \HOneScenarioScore{19.58}{10.97} \\
\addlinespace[0.6pt]
Claude & SVR (\%) & \HOneScenarioScore{6.80}{2.53} & \HOneScenarioScore{54.17}{1.11} & \HOneScenarioScore{37.82}{9.07} & \HOneScenarioScore{36.78}{7.92} & \HOneScenarioScore{24.46}{8.09} & \HOneScenarioScore{4.73}{2.81} & \HOneScenarioScore{27.46}{6.14} \\
 & FCT (s) & \HOneScenarioScore{500.0}{0.0} & \HOneScenarioScore{\textbf{343.7}}{136.6} & \HOneScenarioScore{1733.3}{152.8} & \HOneScenarioScore{2666.7}{850.5} & \HOneScenarioScore{1866.7}{850.5} & \HOneScenarioScore{1566.7}{321.5} & \HOneScenarioScore{1446.2}{515.1} \\
 & HPR (\%) & \HOneScenarioScore{4.03}{1.03} & \HOneScenarioScore{39.39}{0.89} & \HOneScenarioScore{58.86}{37.01} & \HOneScenarioScore{35.48}{2.71} & \HOneScenarioScore{\textbf{86.91}}{0.38} & \HOneScenarioScore{68.47}{12.13} & \HOneScenarioScore{48.86}{15.95} \\
 & CRR (\%) & \HOneScenarioScore{16.67}{28.87} & \HOneScenarioScore{50.00}{0.00} & \HOneScenarioScore{\textbf{100.00}}{0.00} & \HOneScenarioScore{\textbf{66.67}}{28.87} & \HOneScenarioScore{\textbf{100.00}}{0.00} & \HOneScenarioScore{\textbf{100.00}}{0.00} & \HOneScenarioScore{72.22}{16.67} \\
 & EC (\%) & \HOneScenarioScore{\textbf{90.14}}{4.08} & \HOneScenarioScore{\textbf{92.85}}{1.69} & \HOneScenarioScore{\textbf{85.17}}{3.90} & \HOneScenarioScore{81.00}{0.84} & \HOneScenarioScore{62.02}{9.39} & \HOneScenarioScore{\textbf{72.53}}{1.73} & \HOneScenarioScore{\textbf{80.62}}{4.59} \\
 & TG (pp) & \HOneScenarioScore{0.37}{8.70} & \HOneScenarioScore{\textbf{13.60}}{21.02} & \HOneScenarioScore{36.10}{9.67} & \HOneScenarioScore{17.93}{10.06} & \HOneScenarioScore{20.38}{9.15} & \HOneScenarioScore{3.64}{3.55} & \HOneScenarioScore{15.34}{11.61} \\
\addlinespace[0.6pt]
DeepSeek & SVR (\%) & \HOneScenarioScore{13.28}{4.41} & \HOneScenarioScore{30.97}{19.44} & \HOneScenarioScore{36.92}{10.07} & \HOneScenarioScore{45.07}{7.80} & \HOneScenarioScore{17.28}{2.61} & \HOneScenarioScore{13.34}{3.82} & \HOneScenarioScore{26.15}{9.84} \\
 & FCT (s) & \HOneScenarioScore{478.7}{232.6} & \HOneScenarioScore{2766.7}{680.7} & \HOneScenarioScore{2866.7}{152.8} & \HOneScenarioScore{2500.0}{866.0} & \HOneScenarioScore{2466.7}{493.3} & \HOneScenarioScore{1166.7}{57.7} & \HOneScenarioScore{2040.9}{506.2} \\
 & HPR (\%) & \HOneScenarioScore{9.98}{5.50} & \HOneScenarioScore{26.28}{12.28} & \HOneScenarioScore{47.59}{5.95} & \HOneScenarioScore{\textbf{40.05}}{7.80} & \HOneScenarioScore{14.38}{2.14} & \HOneScenarioScore{46.19}{8.49} & \HOneScenarioScore{30.74}{7.68} \\
 & CRR (\%) & \HOneScenarioScore{16.67}{28.87} & \HOneScenarioScore{16.67}{28.87} & \HOneScenarioScore{50.00}{0.00} & \HOneScenarioScore{\textbf{66.67}}{28.87} & \HOneScenarioScore{50.00}{0.00} & \HOneScenarioScore{\textbf{100.00}}{0.00} & \HOneScenarioScore{50.00}{20.41} \\
 & EC (\%) & \HOneScenarioScore{68.73}{1.69} & \HOneScenarioScore{51.24}{20.10} & \HOneScenarioScore{69.89}{0.67} & \HOneScenarioScore{76.62}{3.49} & \HOneScenarioScore{33.19}{1.60} & \HOneScenarioScore{62.36}{4.68} & \HOneScenarioScore{60.34}{8.60} \\
 & TG (pp) & \HOneScenarioScore{5.25}{10.51} & \HOneScenarioScore{-22.59}{19.25} & \HOneScenarioScore{23.28}{22.73} & \HOneScenarioScore{34.61}{7.14} & \HOneScenarioScore{12.09}{3.02} & \HOneScenarioScore{7.59}{4.49} & \HOneScenarioScore{10.04}{13.40} \\
\addlinespace[0.6pt]
GLM & SVR (\%) & \HOneScenarioScore{10.72}{5.50} & \HOneScenarioScore{53.95}{0.55} & \HOneScenarioScore{33.59}{9.89} & \HOneScenarioScore{40.63}{8.55} & \HOneScenarioScore{30.18}{6.54} & \HOneScenarioScore{13.43}{4.87} & \HOneScenarioScore{30.42}{6.68} \\
 & FCT (s) & \HOneScenarioScore{403.5}{167.2} & \HOneScenarioScore{355.9}{124.9} & \HOneScenarioScore{2566.7}{208.2} & \HOneScenarioScore{2600.0}{529.2} & \HOneScenarioScore{1833.3}{1078.6} & \HOneScenarioScore{1100.0}{300.0} & \HOneScenarioScore{1476.6}{519.6} \\
 & HPR (\%) & \HOneScenarioScore{6.23}{0.19} & \HOneScenarioScore{32.03}{10.73} & \HOneScenarioScore{73.96}{6.53} & \HOneScenarioScore{33.20}{5.77} & \HOneScenarioScore{69.60}{26.66} & \HOneScenarioScore{53.72}{21.58} & \HOneScenarioScore{44.79}{15.10} \\
 & CRR (\%) & \HOneScenarioScore{0.00}{0.00} & \HOneScenarioScore{50.00}{0.00} & \HOneScenarioScore{66.67}{28.87} & \HOneScenarioScore{\textbf{66.67}}{28.87} & \HOneScenarioScore{83.33}{28.87} & \HOneScenarioScore{66.67}{28.87} & \HOneScenarioScore{55.56}{23.57} \\
 & EC (\%) & \HOneScenarioScore{81.75}{4.80} & \HOneScenarioScore{75.66}{13.41} & \HOneScenarioScore{61.50}{9.43} & \HOneScenarioScore{67.42}{2.77} & \HOneScenarioScore{59.73}{8.05} & \HOneScenarioScore{51.60}{19.75} & \HOneScenarioScore{66.28}{11.21} \\
 & TG (pp) & \HOneScenarioScore{8.31}{4.95} & \HOneScenarioScore{0.35}{1.09} & \HOneScenarioScore{28.82}{9.12} & \HOneScenarioScore{19.06}{19.28} & \HOneScenarioScore{26.45}{5.18} & \HOneScenarioScore{10.91}{4.47} & \HOneScenarioScore{15.65}{9.38} \\
\bottomrule
\end{tabular*}\par
\vspace{7pt}
\captionof{table}{Per-scenario visual results (Table~3 of the main paper).
Symbolic model trials used fourfold suppression in real time. Visual model trials used
eightfold suppression with the timing protocol in Appendix~\ref{appendix:visual_trial_protocols}.
For each task and metric, bold marks each column's best mean separately for RGB and RGB+DS,
including ties (FCT $\downarrow$, others $\uparrow$).}
\label{tab:appendix_h1_visual_results}
\noindent\begin{tabular*}{\dimexpr\linewidth-10pt\relax}{@{}ll@{\extracolsep{\fill}}*{7}{r}@{}}
\toprule
\textbf{Model} & \textbf{Metric} & \multicolumn{6}{c}{\textbf{Scenario}} & {} \\
\midrule
\multicolumn{2}{@{}l}{\textbf{T3. RGB}} & \makebox[51.4pt][c]{\textbf{S1F}} & \makebox[51.4pt][c]{\textbf{S2F}} & \makebox[51.4pt][c]{\textbf{S3F}} & \makebox[51.4pt][c]{\textbf{S4F}} & \makebox[51.4pt][c]{\textbf{S5F}} & \makebox[51.4pt][c]{\textbf{S6F}} & \makebox[51.4pt][c]{\textbf{Overall}} \\
GPT-6.1 & SVR (\%) & \HOneScenarioScore{\textbf{4.79}}{1.92} & \HOneScenarioScore{\textbf{53.39}}{0.12} & \HOneScenarioScore{\textbf{10.23}}{3.42} & \HOneScenarioScore{11.24}{4.11} & \HOneScenarioScore{6.52}{2.39} & \HOneScenarioScore{7.95}{5.97} & \HOneScenarioScore{\textbf{15.69}}{3.50} \\
 & FCT (s) & \HOneScenarioScore{\textbf{466.1}}{208.9} & \HOneScenarioScore{\textbf{319.9}}{69.4} & \HOneScenarioScore{2200.0}{0.0} & \HOneScenarioScore{2400.0}{100.0} & \HOneScenarioScore{2233.3}{288.7} & \HOneScenarioScore{\textbf{1400.0}}{264.6} & \HOneScenarioScore{\textbf{1503.2}}{187.9} \\
 & EC (\%) & \HOneScenarioScore{\textbf{54.08}}{10.71} & \HOneScenarioScore{\textbf{63.45}}{2.82} & \HOneScenarioScore{\textbf{69.29}}{0.34} & \HOneScenarioScore{\textbf{60.08}}{2.72} & \HOneScenarioScore{36.36}{2.59} & \HOneScenarioScore{40.31}{11.19} & \HOneScenarioScore{\textbf{53.93}}{6.61} \\
\addlinespace[0.6pt]
GPT-6 & SVR (\%) & \HOneScenarioScore{0.66}{0.50} & \HOneScenarioScore{1.27}{1.21} & \HOneScenarioScore{9.08}{0.40} & \HOneScenarioScore{4.76}{4.11} & \HOneScenarioScore{15.33}{16.75} & \HOneScenarioScore{1.69}{1.17} & \HOneScenarioScore{5.46}{7.08} \\
 & FCT (s) & \HOneScenarioScore{700.0}{0.0} & \HOneScenarioScore{1666.7}{57.7} & \HOneScenarioScore{\textbf{1766.7}}{251.7} & \HOneScenarioScore{2800.0}{0.0} & \HOneScenarioScore{1666.7}{850.5} & \HOneScenarioScore{1600.0}{0.0} & \HOneScenarioScore{1700.0}{362.9} \\
 & EC (\%) & \HOneScenarioScore{22.52}{1.01} & \HOneScenarioScore{24.58}{0.42} & \HOneScenarioScore{21.05}{4.95} & \HOneScenarioScore{23.82}{4.40} & \HOneScenarioScore{26.62}{12.77} & \HOneScenarioScore{17.59}{1.61} & \HOneScenarioScore{22.70}{5.93} \\
\addlinespace[0.6pt]
Claude & SVR (\%) & \HOneScenarioScore{2.19}{2.02} & \HOneScenarioScore{18.60}{30.43} & \HOneScenarioScore{8.07}{1.52} & \HOneScenarioScore{1.46}{2.19} & \HOneScenarioScore{13.89}{18.28} & \HOneScenarioScore{2.41}{3.55} & \HOneScenarioScore{7.77}{14.63} \\
 & FCT (s) & \HOneScenarioScore{600.0}{173.2} & \HOneScenarioScore{1333.3}{814.5} & \HOneScenarioScore{1900.0}{100.0} & \HOneScenarioScore{2766.7}{57.7} & \HOneScenarioScore{\textbf{1633.3}}{1069.3} & \HOneScenarioScore{1500.0}{346.4} & \HOneScenarioScore{1622.2}{573.0} \\
 & EC (\%) & \HOneScenarioScore{32.60}{6.63} & \HOneScenarioScore{29.08}{8.33} & \HOneScenarioScore{21.87}{0.69} & \HOneScenarioScore{25.32}{8.35} & \HOneScenarioScore{19.25}{3.46} & \HOneScenarioScore{24.42}{10.10} & \HOneScenarioScore{25.42}{7.04} \\
\addlinespace[0.6pt]
DeepSeek & SVR (\%) & \HOneScenarioScore{0.59}{0.51} & \HOneScenarioScore{1.50}{0.89} & \HOneScenarioScore{2.39}{0.42} & \HOneScenarioScore{\textbf{13.87}}{11.77} & \HOneScenarioScore{3.82}{0.68} & \HOneScenarioScore{2.25}{0.73} & \HOneScenarioScore{4.07}{4.84} \\
 & FCT (s) & \HOneScenarioScore{700.0}{0.0} & \HOneScenarioScore{1833.3}{115.5} & \HOneScenarioScore{2733.3}{115.5} & \HOneScenarioScore{\textbf{2333.3}}{1069.3} & \HOneScenarioScore{2333.3}{57.7} & \HOneScenarioScore{1700.0}{0.0} & \HOneScenarioScore{1938.9}{442.2} \\
 & EC (\%) & \HOneScenarioScore{39.37}{6.89} & \HOneScenarioScore{42.01}{2.94} & \HOneScenarioScore{39.70}{2.92} & \HOneScenarioScore{40.58}{4.15} & \HOneScenarioScore{\textbf{38.64}}{2.27} & \HOneScenarioScore{\textbf{40.53}}{5.30} & \HOneScenarioScore{40.14}{4.38} \\
\addlinespace[0.6pt]
GLM & SVR (\%) & \HOneScenarioScore{2.51}{1.30} & \HOneScenarioScore{36.28}{30.63} & \HOneScenarioScore{8.73}{1.87} & \HOneScenarioScore{8.08}{2.38} & \HOneScenarioScore{\textbf{15.55}}{15.42} & \HOneScenarioScore{\textbf{9.33}}{8.27} & \HOneScenarioScore{13.41}{14.46} \\
 & FCT (s) & \HOneScenarioScore{666.7}{57.7} & \HOneScenarioScore{833.3}{750.6} & \HOneScenarioScore{2000.0}{0.0} & \HOneScenarioScore{2966.7}{57.7} & \HOneScenarioScore{1700.0}{888.8} & \HOneScenarioScore{1433.3}{321.5} & \HOneScenarioScore{1600.0}{493.9} \\
 & EC (\%) & \HOneScenarioScore{36.28}{9.76} & \HOneScenarioScore{21.29}{3.27} & \HOneScenarioScore{26.37}{6.82} & \HOneScenarioScore{25.25}{4.64} & \HOneScenarioScore{28.83}{2.89} & \HOneScenarioScore{30.59}{16.42} & \HOneScenarioScore{28.10}{8.68} \\
\midrule
\multicolumn{2}{@{}l}{\textbf{T3. RGB+DS}} & \makebox[51.4pt][c]{\textbf{S1F}} & \makebox[51.4pt][c]{\textbf{S2F}} & \makebox[51.4pt][c]{\textbf{S3F}} & \makebox[51.4pt][c]{\textbf{S4F}} & \makebox[51.4pt][c]{\textbf{S5F}} & \makebox[51.4pt][c]{\textbf{S6F}} & \makebox[51.4pt][c]{\textbf{Overall}} \\
GPT-6.1 & SVR (\%) & \HOneScenarioScore{4.28}{2.13} & \HOneScenarioScore{15.58}{3.12} & \HOneScenarioScore{4.44}{2.01} & \HOneScenarioScore{0.07}{7.49} & \HOneScenarioScore{9.43}{1.20} & \HOneScenarioScore{\textbf{12.63}}{8.55} & \HOneScenarioScore{7.74}{4.98} \\
 & FCT (s) & \HOneScenarioScore{431.4}{232.6} & \HOneScenarioScore{1066.7}{57.7} & \HOneScenarioScore{2600.0}{346.4} & \HOneScenarioScore{2833.3}{351.2} & \HOneScenarioScore{2366.7}{57.7} & \HOneScenarioScore{\textbf{900.0}}{608.3} & \HOneScenarioScore{1699.7}{335.2} \\
 & EC (\%) & \HOneScenarioScore{\textbf{58.94}}{0.80} & \HOneScenarioScore{\textbf{57.91}}{0.14} & \HOneScenarioScore{\textbf{52.21}}{10.79} & \HOneScenarioScore{\textbf{54.97}}{9.54} & \HOneScenarioScore{\textbf{40.56}}{1.22} & \HOneScenarioScore{32.59}{18.84} & \HOneScenarioScore{\textbf{49.53}}{9.70} \\
\addlinespace[0.6pt]
GPT-6 & SVR (\%) & \HOneScenarioScore{\textbf{5.57}}{0.29} & \HOneScenarioScore{\textbf{53.33}}{0.26} & \HOneScenarioScore{17.91}{6.46} & \HOneScenarioScore{\textbf{24.29}}{15.16} & \HOneScenarioScore{\textbf{23.70}}{19.59} & \HOneScenarioScore{7.43}{5.92} & \HOneScenarioScore{\textbf{22.04}}{10.73} \\
 & FCT (s) & \HOneScenarioScore{\textbf{363.5}}{118.3} & \HOneScenarioScore{\textbf{400.0}}{0.0} & \HOneScenarioScore{\textbf{1700.0}}{556.8} & \HOneScenarioScore{\textbf{1800.0}}{608.3} & \HOneScenarioScore{\textbf{933.3}}{923.8} & \HOneScenarioScore{1333.3}{152.8} & \HOneScenarioScore{\textbf{1088.4}}{511.6} \\
 & EC (\%) & \HOneScenarioScore{23.99}{0.64} & \HOneScenarioScore{26.91}{3.11} & \HOneScenarioScore{25.02}{0.34} & \HOneScenarioScore{28.50}{1.63} & \HOneScenarioScore{20.58}{0.96} & \HOneScenarioScore{21.68}{1.34} & \HOneScenarioScore{24.44}{1.61} \\
\addlinespace[0.6pt]
Claude & SVR (\%) & \HOneScenarioScore{0.43}{0.28} & \HOneScenarioScore{0.51}{0.46} & \HOneScenarioScore{1.03}{0.44} & \HOneScenarioScore{14.94}{24.19} & \HOneScenarioScore{15.06}{15.63} & \HOneScenarioScore{10.48}{6.50} & \HOneScenarioScore{7.07}{12.06} \\
 & FCT (s) & \HOneScenarioScore{666.7}{57.7} & \HOneScenarioScore{1700.0}{0.0} & \HOneScenarioScore{2600.0}{0.0} & \HOneScenarioScore{2133.3}{1069.3} & \HOneScenarioScore{1633.3}{907.4} & \HOneScenarioScore{1166.7}{305.5} & \HOneScenarioScore{1650.0}{586.4} \\
 & EC (\%) & \HOneScenarioScore{31.05}{1.13} & \HOneScenarioScore{40.16}{2.53} & \HOneScenarioScore{29.21}{3.90} & \HOneScenarioScore{29.46}{6.95} & \HOneScenarioScore{23.16}{4.66} & \HOneScenarioScore{20.71}{0.22} & \HOneScenarioScore{28.96}{3.94} \\
\addlinespace[0.6pt]
DeepSeek & SVR (\%) & \HOneScenarioScore{3.48}{2.89} & \HOneScenarioScore{1.06}{0.72} & \HOneScenarioScore{\textbf{18.78}}{8.63} & \HOneScenarioScore{16.99}{25.68} & \HOneScenarioScore{13.00}{18.83} & \HOneScenarioScore{1.56}{0.31} & \HOneScenarioScore{9.14}{13.52} \\
 & FCT (s) & \HOneScenarioScore{666.7}{57.7} & \HOneScenarioScore{1766.7}{115.5} & \HOneScenarioScore{2033.3}{896.3} & \HOneScenarioScore{2200.0}{1039.2} & \HOneScenarioScore{1666.7}{1097.0} & \HOneScenarioScore{1633.3}{115.5} & \HOneScenarioScore{1661.1}{720.7} \\
 & EC (\%) & \HOneScenarioScore{27.45}{9.09} & \HOneScenarioScore{36.63}{4.46} & \HOneScenarioScore{29.89}{5.95} & \HOneScenarioScore{40.39}{8.07} & \HOneScenarioScore{22.64}{6.36} & \HOneScenarioScore{29.70}{6.46} & \HOneScenarioScore{31.12}{6.90} \\
\addlinespace[0.6pt]
GLM & SVR (\%) & \HOneScenarioScore{2.35}{1.56} & \HOneScenarioScore{35.99}{30.34} & \HOneScenarioScore{5.27}{4.66} & \HOneScenarioScore{14.61}{26.37} & \HOneScenarioScore{3.00}{1.28} & \HOneScenarioScore{1.25}{0.78} & \HOneScenarioScore{10.41}{16.54} \\
 & FCT (s) & \HOneScenarioScore{666.7}{57.7} & \HOneScenarioScore{833.3}{750.6} & \HOneScenarioScore{2233.3}{251.7} & \HOneScenarioScore{2166.7}{1193.0} & \HOneScenarioScore{2266.7}{57.7} & \HOneScenarioScore{1666.7}{57.7} & \HOneScenarioScore{1638.9}{585.9} \\
 & EC (\%) & \HOneScenarioScore{34.22}{9.51} & \HOneScenarioScore{29.72}{7.86} & \HOneScenarioScore{32.43}{14.62} & \HOneScenarioScore{29.51}{2.82} & \HOneScenarioScore{27.95}{6.62} & \HOneScenarioScore{\textbf{40.76}}{4.25} & \HOneScenarioScore{32.43}{8.52} \\
\bottomrule
\end{tabular*}\par
\newpage
{\fontsize{11}{12}\selectfont\bfseries H.1. Complete Results and Dispersion Statistics (continued)\par}
\vspace{5pt}
\fontsize{8.2}{8.5}\selectfont
\captionof*{table}{\textbf{Table~\ref{tab:appendix_h1_visual_results} (continued).} Per-scenario visual results.}
{\fontsize{8.3}{9.5}\selectfont
Across all six T5 scenarios, adding DS to RGB raised mean SVR for DeepSeek
(by 31.09 percentage points in S4R) and lowered it for GLM.\par}
\vspace{3pt}
\noindent\begin{tabular*}{\dimexpr\linewidth-10pt\relax}{@{}ll@{\extracolsep{\fill}}*{7}{r}@{}}
\toprule
\textbf{Model} & \textbf{Metric} & \multicolumn{6}{c}{\textbf{Scenario}} & {} \\
\midrule
\multicolumn{2}{@{}l}{\textbf{T5. RGB}} & \makebox[51.4pt][c]{\textbf{S1R}} & \makebox[51.4pt][c]{\textbf{S2R}} & \makebox[51.4pt][c]{\textbf{S3R}} & \makebox[51.4pt][c]{\textbf{S4R}} & \makebox[51.4pt][c]{\textbf{S5R}} & \makebox[51.4pt][c]{\textbf{S6R}} & \makebox[51.4pt][c]{\textbf{Overall}} \\
GPT-6.1 & SVR (\%) & \HOneScenarioScore{\textbf{12.34}}{1.10} & \HOneScenarioScore{\textbf{55.07}}{0.56} & \HOneScenarioScore{\textbf{34.75}}{2.72} & \HOneScenarioScore{37.98}{11.90} & \HOneScenarioScore{32.75}{5.41} & \HOneScenarioScore{\textbf{16.96}}{3.64} & \HOneScenarioScore{\textbf{31.64}}{5.67} \\
 & FCT (s) & \HOneScenarioScore{\textbf{466.7}}{57.7} & \HOneScenarioScore{\textbf{281.3}}{16.8} & \HOneScenarioScore{\textbf{1733.3}}{115.5} & \HOneScenarioScore{2200.0}{1126.9} & \HOneScenarioScore{1966.7}{838.6} & \HOneScenarioScore{\textbf{1100.0}}{500.0} & \HOneScenarioScore{1291.3}{611.0} \\
 & HPR (\%) & \HOneScenarioScore{\textbf{5.89}}{0.72} & \HOneScenarioScore{\textbf{39.12}}{0.26} & \HOneScenarioScore{\textbf{60.33}}{6.95} & \HOneScenarioScore{\textbf{24.42}}{7.86} & \HOneScenarioScore{53.30}{38.81} & \HOneScenarioScore{\textbf{46.22}}{19.55} & \HOneScenarioScore{\textbf{38.21}}{18.26} \\
 & CRR (\%) & \HOneScenarioScore{\textbf{33.33}}{28.87} & \HOneScenarioScore{\textbf{50.00}}{0.00} & \HOneScenarioScore{\textbf{100.00}}{0.00} & \HOneScenarioScore{\textbf{50.00}}{0.00} & \HOneScenarioScore{\textbf{83.33}}{28.87} & \HOneScenarioScore{\textbf{83.33}}{28.87} & \HOneScenarioScore{\textbf{66.67}}{20.41} \\
 & EC (\%) & \HOneScenarioScore{80.13}{8.18} & \HOneScenarioScore{\textbf{79.52}}{10.88} & \HOneScenarioScore{\textbf{79.93}}{5.07} & \HOneScenarioScore{78.87}{6.83} & \HOneScenarioScore{67.40}{15.62} & \HOneScenarioScore{74.02}{8.42} & \HOneScenarioScore{76.64}{9.77} \\
 & TG (pp) & \HOneScenarioScore{\textbf{7.96}}{2.36} & \HOneScenarioScore{\textbf{47.53}}{1.55} & \HOneScenarioScore{\textbf{25.25}}{2.70} & \HOneScenarioScore{25.58}{11.38} & \HOneScenarioScore{18.11}{20.92} & \HOneScenarioScore{8.97}{2.03} & \HOneScenarioScore{\textbf{22.23}}{9.89} \\
\addlinespace[0.6pt]
GPT-6 & SVR (\%) & \HOneScenarioScore{5.98}{1.09} & \HOneScenarioScore{24.09}{26.05} & \HOneScenarioScore{27.31}{10.90} & \HOneScenarioScore{\textbf{57.77}}{31.28} & \HOneScenarioScore{\textbf{34.86}}{1.50} & \HOneScenarioScore{12.08}{4.01} & \HOneScenarioScore{27.02}{17.30} \\
 & FCT (s) & \HOneScenarioScore{533.3}{57.7} & \HOneScenarioScore{1266.7}{757.2} & \HOneScenarioScore{1900.0}{953.9} & \HOneScenarioScore{\textbf{1572.5}}{1224.6} & \HOneScenarioScore{\textbf{733.3}}{57.7} & \HOneScenarioScore{1133.3}{57.7} & \HOneScenarioScore{\textbf{1189.9}}{706.3} \\
 & HPR (\%) & \HOneScenarioScore{4.88}{0.20} & \HOneScenarioScore{18.19}{10.09} & \HOneScenarioScore{31.63}{2.54} & \HOneScenarioScore{17.65}{18.65} & \HOneScenarioScore{\textbf{60.07}}{11.20} & \HOneScenarioScore{4.78}{8.27} & \HOneScenarioScore{22.87}{10.41} \\
 & CRR (\%) & \HOneScenarioScore{0.00}{0.00} & \HOneScenarioScore{0.00}{0.00} & \HOneScenarioScore{0.00}{0.00} & \HOneScenarioScore{33.33}{28.87} & \HOneScenarioScore{33.33}{28.87} & \HOneScenarioScore{33.33}{28.87} & \HOneScenarioScore{16.67}{20.41} \\
 & EC (\%) & \HOneScenarioScore{58.87}{4.47} & \HOneScenarioScore{56.39}{12.00} & \HOneScenarioScore{52.28}{4.18} & \HOneScenarioScore{53.68}{15.52} & \HOneScenarioScore{58.04}{8.00} & \HOneScenarioScore{49.81}{3.22} & \HOneScenarioScore{54.85}{9.10} \\
 & TG (pp) & \HOneScenarioScore{5.33}{1.10} & \HOneScenarioScore{5.37}{51.76} & \HOneScenarioScore{23.86}{7.90} & \HOneScenarioScore{\textbf{35.38}}{35.13} & \HOneScenarioScore{20.17}{14.61} & \HOneScenarioScore{10.16}{5.11} & \HOneScenarioScore{16.71}{26.51} \\
\addlinespace[0.6pt]
Claude & SVR (\%) & \HOneScenarioScore{6.03}{2.40} & \HOneScenarioScore{21.29}{27.54} & \HOneScenarioScore{10.68}{2.32} & \HOneScenarioScore{25.39}{17.72} & \HOneScenarioScore{26.32}{16.23} & \HOneScenarioScore{6.38}{4.35} & \HOneScenarioScore{16.02}{15.09} \\
 & FCT (s) & \HOneScenarioScore{533.3}{57.7} & \HOneScenarioScore{1333.3}{737.1} & \HOneScenarioScore{2766.7}{404.1} & \HOneScenarioScore{2133.3}{1115.0} & \HOneScenarioScore{1300.0}{1135.8} & \HOneScenarioScore{1633.3}{305.5} & \HOneScenarioScore{1616.7}{745.7} \\
 & HPR (\%) & \HOneScenarioScore{4.02}{0.34} & \HOneScenarioScore{26.65}{10.38} & \HOneScenarioScore{22.27}{1.35} & \HOneScenarioScore{6.08}{6.34} & \HOneScenarioScore{46.91}{28.66} & \HOneScenarioScore{0.00}{0.00} & \HOneScenarioScore{17.65}{12.72} \\
 & CRR (\%) & \HOneScenarioScore{0.00}{0.00} & \HOneScenarioScore{33.33}{28.87} & \HOneScenarioScore{16.67}{28.87} & \HOneScenarioScore{16.67}{28.87} & \HOneScenarioScore{50.00}{0.00} & \HOneScenarioScore{33.33}{28.87} & \HOneScenarioScore{25.00}{23.57} \\
 & EC (\%) & \HOneScenarioScore{82.05}{9.12} & \HOneScenarioScore{69.00}{0.37} & \HOneScenarioScore{76.03}{6.42} & \HOneScenarioScore{71.08}{1.07} & \HOneScenarioScore{55.83}{7.62} & \HOneScenarioScore{58.13}{4.24} & \HOneScenarioScore{68.68}{5.80} \\
 & TG (pp) & \HOneScenarioScore{4.59}{3.34} & \HOneScenarioScore{3.92}{3.40} & \HOneScenarioScore{5.33}{4.12} & \HOneScenarioScore{6.63}{7.87} & \HOneScenarioScore{\textbf{24.55}}{17.58} & \HOneScenarioScore{0.33}{9.17} & \HOneScenarioScore{7.56}{9.08} \\
\addlinespace[0.6pt]
DeepSeek & SVR (\%) & \HOneScenarioScore{5.06}{0.66} & \HOneScenarioScore{9.94}{2.67} & \HOneScenarioScore{17.75}{1.82} & \HOneScenarioScore{19.65}{1.21} & \HOneScenarioScore{8.65}{2.59} & \HOneScenarioScore{15.51}{20.03} & \HOneScenarioScore{12.76}{8.37} \\
 & FCT (s) & \HOneScenarioScore{600.0}{0.0} & \HOneScenarioScore{2000.0}{0.0} & \HOneScenarioScore{2833.3}{115.5} & \HOneScenarioScore{3666.7}{115.5} & \HOneScenarioScore{2666.7}{57.7} & \HOneScenarioScore{2000.0}{173.2} & \HOneScenarioScore{2294.4}{100.0} \\
 & HPR (\%) & \HOneScenarioScore{5.56}{0.91} & \HOneScenarioScore{14.36}{2.12} & \HOneScenarioScore{32.49}{2.59} & \HOneScenarioScore{0.00}{0.00} & \HOneScenarioScore{2.69}{0.19} & \HOneScenarioScore{0.00}{0.00} & \HOneScenarioScore{9.18}{1.42} \\
 & CRR (\%) & \HOneScenarioScore{0.00}{0.00} & \HOneScenarioScore{0.00}{0.00} & \HOneScenarioScore{0.00}{0.00} & \HOneScenarioScore{0.00}{0.00} & \HOneScenarioScore{0.00}{0.00} & \HOneScenarioScore{0.00}{0.00} & \HOneScenarioScore{0.00}{0.00} \\
 & EC (\%) & \HOneScenarioScore{\textbf{82.93}}{3.89} & \HOneScenarioScore{76.47}{4.98} & \HOneScenarioScore{78.80}{9.82} & \HOneScenarioScore{\textbf{92.56}}{2.19} & \HOneScenarioScore{\textbf{74.63}}{3.22} & \HOneScenarioScore{\textbf{82.93}}{3.57} & \HOneScenarioScore{\textbf{81.39}}{5.23} \\
 & TG (pp) & \HOneScenarioScore{4.12}{0.70} & \HOneScenarioScore{8.02}{3.71} & \HOneScenarioScore{15.65}{1.98} & \HOneScenarioScore{15.65}{3.81} & \HOneScenarioScore{4.69}{2.59} & \HOneScenarioScore{\textbf{14.46}}{20.66} & \HOneScenarioScore{10.43}{8.81} \\
\addlinespace[0.6pt]
GLM & SVR (\%) & \HOneScenarioScore{8.17}{2.65} & \HOneScenarioScore{53.30}{0.06} & \HOneScenarioScore{25.76}{6.71} & \HOneScenarioScore{24.45}{17.98} & \HOneScenarioScore{23.08}{10.14} & \HOneScenarioScore{10.20}{7.16} & \HOneScenarioScore{24.16}{9.39} \\
 & FCT (s) & \HOneScenarioScore{533.3}{57.7} & \HOneScenarioScore{500.0}{0.0} & \HOneScenarioScore{2000.0}{346.4} & \HOneScenarioScore{2633.3}{1193.0} & \HOneScenarioScore{1600.0}{700.0} & \HOneScenarioScore{1333.3}{416.3} & \HOneScenarioScore{1433.3}{606.9} \\
 & HPR (\%) & \HOneScenarioScore{5.07}{0.15} & \HOneScenarioScore{31.22}{7.40} & \HOneScenarioScore{41.12}{10.33} & \HOneScenarioScore{7.84}{11.71} & \HOneScenarioScore{58.64}{10.91} & \HOneScenarioScore{10.17}{17.62} & \HOneScenarioScore{25.68}{11.01} \\
 & CRR (\%) & \HOneScenarioScore{0.00}{0.00} & \HOneScenarioScore{\textbf{50.00}}{0.00} & \HOneScenarioScore{16.67}{28.87} & \HOneScenarioScore{33.33}{28.87} & \HOneScenarioScore{50.00}{0.00} & \HOneScenarioScore{16.67}{28.87} & \HOneScenarioScore{27.78}{20.41} \\
 & EC (\%) & \HOneScenarioScore{79.40}{10.08} & \HOneScenarioScore{60.72}{6.90} & \HOneScenarioScore{64.79}{8.80} & \HOneScenarioScore{42.97}{14.41} & \HOneScenarioScore{50.88}{1.17} & \HOneScenarioScore{55.38}{1.85} & \HOneScenarioScore{59.03}{8.55} \\
 & TG (pp) & \HOneScenarioScore{-1.35}{10.15} & \HOneScenarioScore{17.48}{30.06} & \HOneScenarioScore{17.41}{2.18} & \HOneScenarioScore{16.45}{18.19} & \HOneScenarioScore{13.36}{8.12} & \HOneScenarioScore{0.28}{11.50} & \HOneScenarioScore{10.61}{16.02} \\
\midrule
\multicolumn{2}{@{}l}{\textbf{T5. RGB+DS}} & \makebox[51.4pt][c]{\textbf{S1R}} & \makebox[51.4pt][c]{\textbf{S2R}} & \makebox[51.4pt][c]{\textbf{S3R}} & \makebox[51.4pt][c]{\textbf{S4R}} & \makebox[51.4pt][c]{\textbf{S5R}} & \makebox[51.4pt][c]{\textbf{S6R}} & \makebox[51.4pt][c]{\textbf{Overall}} \\
GPT-6.1 & SVR (\%) & \HOneScenarioScore{\textbf{22.39}}{10.34} & \HOneScenarioScore{\textbf{54.12}}{0.39} & \HOneScenarioScore{\textbf{34.54}}{8.52} & \HOneScenarioScore{33.48}{28.97} & \HOneScenarioScore{26.97}{7.19} & \HOneScenarioScore{10.72}{4.87} & \HOneScenarioScore{\textbf{30.37}}{13.50} \\
 & FCT (s) & \HOneScenarioScore{\textbf{271.2}}{198.6} & \HOneScenarioScore{\textbf{315.6}}{73.1} & \HOneScenarioScore{\textbf{1200.0}}{300.0} & \HOneScenarioScore{2166.7}{1569.5} & \HOneScenarioScore{1666.7}{945.2} & \HOneScenarioScore{1300.0}{200.0} & \HOneScenarioScore{\textbf{1153.4}}{767.2} \\
 & HPR (\%) & \HOneScenarioScore{4.99}{0.96} & \HOneScenarioScore{\textbf{37.00}}{2.50} & \HOneScenarioScore{\textbf{48.68}}{14.15} & \HOneScenarioScore{10.56}{18.29} & \HOneScenarioScore{\textbf{64.22}}{21.84} & \HOneScenarioScore{\textbf{31.74}}{23.21} & \HOneScenarioScore{\textbf{32.86}}{16.11} \\
 & CRR (\%) & \HOneScenarioScore{\textbf{50.00}}{0.00} & \HOneScenarioScore{\textbf{50.00}}{0.00} & \HOneScenarioScore{\textbf{66.67}}{28.87} & \HOneScenarioScore{16.67}{28.87} & \HOneScenarioScore{\textbf{100.00}}{0.00} & \HOneScenarioScore{\textbf{83.33}}{28.87} & \HOneScenarioScore{\textbf{61.11}}{20.41} \\
 & EC (\%) & \HOneScenarioScore{\textbf{85.36}}{3.75} & \HOneScenarioScore{\textbf{81.04}}{4.45} & \HOneScenarioScore{73.48}{3.84} & \HOneScenarioScore{65.09}{12.84} & \HOneScenarioScore{60.77}{1.05} & \HOneScenarioScore{\textbf{62.29}}{7.40} & \HOneScenarioScore{\textbf{71.34}}{6.70} \\
 & TG (pp) & \HOneScenarioScore{\textbf{18.33}}{9.96} & \HOneScenarioScore{\textbf{50.89}}{0.32} & \HOneScenarioScore{\textbf{23.05}}{10.54} & \HOneScenarioScore{22.17}{31.21} & \HOneScenarioScore{21.78}{5.98} & \HOneScenarioScore{-5.66}{4.20} & \HOneScenarioScore{\textbf{21.76}}{14.37} \\
\addlinespace[0.6pt]
GPT-6 & SVR (\%) & \HOneScenarioScore{6.58}{2.48} & \HOneScenarioScore{37.61}{28.44} & \HOneScenarioScore{20.21}{1.51} & \HOneScenarioScore{33.35}{19.85} & \HOneScenarioScore{25.27}{12.72} & \HOneScenarioScore{14.14}{9.37} & \HOneScenarioScore{22.86}{15.60} \\
 & FCT (s) & \HOneScenarioScore{500.0}{100.0} & \HOneScenarioScore{866.7}{808.3} & \HOneScenarioScore{2033.3}{513.2} & \HOneScenarioScore{1833.3}{1443.4} & \HOneScenarioScore{1333.3}{1101.5} & \HOneScenarioScore{1066.7}{635.1} & \HOneScenarioScore{1272.2}{878.1} \\
 & HPR (\%) & \HOneScenarioScore{\textbf{5.44}}{1.38} & \HOneScenarioScore{26.00}{19.24} & \HOneScenarioScore{27.16}{1.52} & \HOneScenarioScore{5.08}{8.80} & \HOneScenarioScore{35.21}{27.16} & \HOneScenarioScore{0.00}{0.00} & \HOneScenarioScore{16.48}{14.08} \\
 & CRR (\%) & \HOneScenarioScore{16.67}{28.87} & \HOneScenarioScore{33.33}{28.87} & \HOneScenarioScore{0.00}{0.00} & \HOneScenarioScore{0.00}{0.00} & \HOneScenarioScore{0.00}{0.00} & \HOneScenarioScore{0.00}{0.00} & \HOneScenarioScore{8.33}{16.67} \\
 & EC (\%) & \HOneScenarioScore{62.62}{12.72} & \HOneScenarioScore{42.73}{11.54} & \HOneScenarioScore{47.79}{5.71} & \HOneScenarioScore{46.31}{11.08} & \HOneScenarioScore{49.63}{1.05} & \HOneScenarioScore{35.93}{12.93} & \HOneScenarioScore{47.50}{10.15} \\
 & TG (pp) & \HOneScenarioScore{-8.22}{6.15} & \HOneScenarioScore{36.39}{28.30} & \HOneScenarioScore{10.50}{7.22} & \HOneScenarioScore{16.87}{19.34} & \HOneScenarioScore{3.35}{4.72} & \HOneScenarioScore{5.30}{14.39} & \HOneScenarioScore{10.70}{15.78} \\
\addlinespace[0.6pt]
Claude & SVR (\%) & \HOneScenarioScore{4.07}{1.21} & \HOneScenarioScore{6.13}{5.71} & \HOneScenarioScore{11.10}{1.31} & \HOneScenarioScore{32.85}{18.72} & \HOneScenarioScore{\textbf{28.03}}{12.95} & \HOneScenarioScore{11.82}{4.64} & \HOneScenarioScore{15.67}{9.79} \\
 & FCT (s) & \HOneScenarioScore{533.3}{115.5} & \HOneScenarioScore{1700.0}{346.4} & \HOneScenarioScore{1966.7}{152.8} & \HOneScenarioScore{1833.3}{1101.5} & \HOneScenarioScore{\textbf{1133.3}}{757.2} & \HOneScenarioScore{\textbf{900.0}}{100.0} & \HOneScenarioScore{1344.4}{570.6} \\
 & HPR (\%) & \HOneScenarioScore{4.16}{0.31} & \HOneScenarioScore{18.37}{22.27} & \HOneScenarioScore{21.93}{2.62} & \HOneScenarioScore{\textbf{10.62}}{15.56} & \HOneScenarioScore{51.93}{10.22} & \HOneScenarioScore{0.00}{0.00} & \HOneScenarioScore{17.84}{11.90} \\
 & CRR (\%) & \HOneScenarioScore{0.00}{0.00} & \HOneScenarioScore{33.33}{28.87} & \HOneScenarioScore{0.00}{0.00} & \HOneScenarioScore{\textbf{33.33}}{28.87} & \HOneScenarioScore{50.00}{50.00} & \HOneScenarioScore{33.33}{28.87} & \HOneScenarioScore{25.00}{28.87} \\
 & EC (\%) & \HOneScenarioScore{81.68}{8.43} & \HOneScenarioScore{74.22}{10.07} & \HOneScenarioScore{\textbf{74.83}}{6.62} & \HOneScenarioScore{\textbf{69.76}}{7.13} & \HOneScenarioScore{\textbf{62.02}}{4.60} & \HOneScenarioScore{60.58}{5.35} & \HOneScenarioScore{70.51}{7.27} \\
 & TG (pp) & \HOneScenarioScore{2.67}{2.44} & \HOneScenarioScore{6.12}{5.35} & \HOneScenarioScore{7.72}{6.12} & \HOneScenarioScore{29.11}{22.82} & \HOneScenarioScore{\textbf{21.96}}{14.98} & \HOneScenarioScore{6.35}{5.84} & \HOneScenarioScore{12.32}{11.91} \\
\addlinespace[0.6pt]
DeepSeek & SVR (\%) & \HOneScenarioScore{5.22}{1.36} & \HOneScenarioScore{22.39}{27.26} & \HOneScenarioScore{24.97}{18.26} & \HOneScenarioScore{\textbf{50.74}}{9.13} & \HOneScenarioScore{26.72}{9.07} & \HOneScenarioScore{\textbf{15.69}}{1.19} & \HOneScenarioScore{24.29}{14.41} \\
 & FCT (s) & \HOneScenarioScore{533.3}{57.7} & \HOneScenarioScore{1400.0}{866.0} & \HOneScenarioScore{2666.7}{702.4} & \HOneScenarioScore{\textbf{1100.0}}{360.6} & \HOneScenarioScore{1866.7}{1101.5} & \HOneScenarioScore{1033.3}{57.7} & \HOneScenarioScore{1433.3}{657.4} \\
 & HPR (\%) & \HOneScenarioScore{4.33}{0.38} & \HOneScenarioScore{31.12}{13.80} & \HOneScenarioScore{22.15}{7.14} & \HOneScenarioScore{8.55}{8.09} & \HOneScenarioScore{20.36}{21.51} & \HOneScenarioScore{0.00}{0.00} & \HOneScenarioScore{14.42}{11.33} \\
 & CRR (\%) & \HOneScenarioScore{0.00}{0.00} & \HOneScenarioScore{33.33}{28.87} & \HOneScenarioScore{16.67}{28.87} & \HOneScenarioScore{16.67}{28.87} & \HOneScenarioScore{16.67}{28.87} & \HOneScenarioScore{16.67}{28.87} & \HOneScenarioScore{16.67}{26.35} \\
 & EC (\%) & \HOneScenarioScore{71.52}{3.07} & \HOneScenarioScore{65.94}{7.88} & \HOneScenarioScore{63.22}{4.73} & \HOneScenarioScore{69.45}{8.84} & \HOneScenarioScore{55.24}{4.85} & \HOneScenarioScore{53.90}{4.25} & \HOneScenarioScore{63.21}{5.97} \\
 & TG (pp) & \HOneScenarioScore{3.14}{3.36} & \HOneScenarioScore{4.17}{4.39} & \HOneScenarioScore{21.85}{18.60} & \HOneScenarioScore{\textbf{30.08}}{9.31} & \HOneScenarioScore{9.47}{18.42} & \HOneScenarioScore{\textbf{13.92}}{1.51} & \HOneScenarioScore{13.77}{11.58} \\
\addlinespace[0.6pt]
GLM & SVR (\%) & \HOneScenarioScore{4.28}{1.04} & \HOneScenarioScore{6.33}{1.51} & \HOneScenarioScore{17.47}{5.99} & \HOneScenarioScore{13.96}{8.88} & \HOneScenarioScore{9.79}{2.27} & \HOneScenarioScore{9.35}{5.53} & \HOneScenarioScore{10.20}{5.06} \\
 & FCT (s) & \HOneScenarioScore{566.7}{57.7} & \HOneScenarioScore{1900.0}{100.0} & \HOneScenarioScore{2233.3}{378.6} & \HOneScenarioScore{3000.0}{173.2} & \HOneScenarioScore{2433.3}{288.7} & \HOneScenarioScore{1366.7}{472.6} & \HOneScenarioScore{1916.7}{286.7} \\
 & HPR (\%) & \HOneScenarioScore{2.79}{2.15} & \HOneScenarioScore{19.21}{12.07} & \HOneScenarioScore{8.61}{28.13} & \HOneScenarioScore{0.00}{0.00} & \HOneScenarioScore{18.92}{29.32} & \HOneScenarioScore{0.00}{0.00} & \HOneScenarioScore{8.25}{17.33} \\
 & CRR (\%) & \HOneScenarioScore{0.00}{0.00} & \HOneScenarioScore{16.67}{28.87} & \HOneScenarioScore{0.00}{0.00} & \HOneScenarioScore{0.00}{0.00} & \HOneScenarioScore{16.67}{28.87} & \HOneScenarioScore{16.67}{28.87} & \HOneScenarioScore{8.33}{20.41} \\
 & EC (\%) & \HOneScenarioScore{65.19}{8.89} & \HOneScenarioScore{57.51}{6.96} & \HOneScenarioScore{59.70}{2.70} & \HOneScenarioScore{42.24}{6.31} & \HOneScenarioScore{43.88}{16.54} & \HOneScenarioScore{50.26}{1.70} & \HOneScenarioScore{53.13}{8.67} \\
 & TG (pp) & \HOneScenarioScore{-0.55}{4.31} & \HOneScenarioScore{1.10}{4.34} & \HOneScenarioScore{5.45}{1.97} & \HOneScenarioScore{13.68}{11.80} & \HOneScenarioScore{3.37}{3.12} & \HOneScenarioScore{6.07}{5.12} & \HOneScenarioScore{4.85}{6.01} \\
\bottomrule
\end{tabular*}\par

\newpage
{\fontsize{11}{12}\selectfont\bfseries H.1. Complete Results and Dispersion Statistics (continued)\par}
\vspace{3pt}
\fontsize{8.2}{8.5}\selectfont
\captionof{table}{Descriptive human reference results for T3 and T5 with RGB observations
(Appendix~\ref{appendix:human}). All 18 T3 trials used eightfold suppression.
T5 pools six fourfold and twelve eightfold trials, one and two per scenario, respectively.}
\label{tab:appendix_h1_human_results}
\noindent\begin{tabular*}{\dimexpr\linewidth-10pt\relax}{@{}ll@{\extracolsep{\fill}}*{7}{r}@{}}
\toprule
\textbf{Task} & \textbf{Metric} & \multicolumn{6}{c}{\textbf{Scenario}} & {} \\
\midrule
\multicolumn{2}{@{}l}{\textbf{T3. RGB}} & \makebox[51.4pt][c]{\textbf{S1F}} & \makebox[51.4pt][c]{\textbf{S2F}} & \makebox[51.4pt][c]{\textbf{S3F}} & \makebox[51.4pt][c]{\textbf{S4F}} & \makebox[51.4pt][c]{\textbf{S5F}} & \makebox[51.4pt][c]{\textbf{S6F}} & \makebox[51.4pt][c]{\textbf{Overall}} \\
T3 & SVR (\%) & \HOneScenarioScore{20.39}{15.49} & \HOneScenarioScore{39.72}{28.19} & \HOneScenarioScore{28.64}{14.41} & \HOneScenarioScore{20.49}{19.49} & \HOneScenarioScore{28.88}{12.91} & \HOneScenarioScore{16.34}{1.74} & \HOneScenarioScore{25.74}{17.28} \\
 & FCT (s) & \HOneScenarioScore{236.2}{99.6} & \HOneScenarioScore{565.9}{549.2} & \HOneScenarioScore{1900.0}{953.9} & \HOneScenarioScore{2300.0}{1300.0} & \HOneScenarioScore{900.0}{866.0} & \HOneScenarioScore{733.3}{115.5} & \HOneScenarioScore{1105.9}{782.6} \\
 & EC (\%) & \HOneScenarioScore{41.72}{15.65} & \HOneScenarioScore{45.06}{9.41} & \HOneScenarioScore{27.57}{3.71} & \HOneScenarioScore{37.89}{8.41} & \HOneScenarioScore{39.16}{12.13} & \HOneScenarioScore{23.24}{2.20} & \HOneScenarioScore{35.77}{9.74} \\
\midrule
\multicolumn{2}{@{}l}{\textbf{T5. RGB}} & \makebox[51.4pt][c]{\textbf{S1R}} & \makebox[51.4pt][c]{\textbf{S2R}} & \makebox[51.4pt][c]{\textbf{S3R}} & \makebox[51.4pt][c]{\textbf{S4R}} & \makebox[51.4pt][c]{\textbf{S5R}} & \makebox[51.4pt][c]{\textbf{S6R}} & \makebox[51.4pt][c]{\textbf{Overall}} \\
T5 & SVR (\%) & \HOneScenarioScore{20.39}{13.90} & \HOneScenarioScore{58.03}{0.70} & \HOneScenarioScore{61.61}{3.28} & \HOneScenarioScore{45.08}{13.88} & \HOneScenarioScore{46.54}{4.70} & \HOneScenarioScore{22.41}{5.53} & \HOneScenarioScore{42.34}{8.66} \\
 & FCT (s) & \HOneScenarioScore{302.5}{258.5} & \HOneScenarioScore{237.5}{5.5} & \HOneScenarioScore{500.0}{173.2} & \HOneScenarioScore{1700.0}{1300.0} & \HOneScenarioScore{326.6}{63.6} & \HOneScenarioScore{700.0}{264.6} & \HOneScenarioScore{627.8}{556.9} \\
 & HPR (\%) & \HOneScenarioScore{13.16}{4.27} & \HOneScenarioScore{40.58}{9.55} & \HOneScenarioScore{79.38}{12.68} & \HOneScenarioScore{35.55}{6.81} & \HOneScenarioScore{83.43}{9.25} & \HOneScenarioScore{41.14}{35.63} & \HOneScenarioScore{48.88}{16.69} \\
 & CRR (\%) & \HOneScenarioScore{33.33}{28.87} & \HOneScenarioScore{66.67}{28.87} & \HOneScenarioScore{66.67}{28.87} & \HOneScenarioScore{50.00}{0.00} & \HOneScenarioScore{83.33}{28.87} & \HOneScenarioScore{50.00}{0.00} & \HOneScenarioScore{58.33}{23.57} \\
 & EC (\%) & \HOneScenarioScore{83.89}{7.46} & \HOneScenarioScore{59.84}{10.86} & \HOneScenarioScore{69.21}{11.92} & \HOneScenarioScore{74.98}{8.48} & \HOneScenarioScore{69.99}{4.55} & \HOneScenarioScore{66.96}{9.75} & \HOneScenarioScore{70.81}{9.16} \\
\bottomrule
\end{tabular*}\par
\endgroup

\subsection{Additional Visual Conditions and Single-Robot References}
\label{appendix:visual_results}
\begingroup
\setlength{\parindent}{0pt}
\setlength{\parskip}{0pt}
\setlength{\tabcolsep}{2pt}
\setlength{\aboverulesep}{1.1pt}
\setlength{\belowrulesep}{1.1pt}
\hfuzz=0.1pt
\vfuzz=0.1pt
\captionsetup{hypcap=false,font=footnotesize,labelfont=bf,position=top,justification=raggedright,singlelinecheck=false,skip=2.2pt}
\newcommand{\HtwoScenarioScore}[2]{%
  \makebox[22pt][r]{#1}\kern0.8pt\textpm\kern0.8pt%
  \makebox[22pt][l]{\textcolor{black!76}{#2}}%
}
{\fontsize{8.3}{9.5}\selectfont
In the symbolic T5 reference, one FR uses the team's local observation
interface to suppress fires and rescue civilians. We use the reference
SVRs in Table~\ref{tab:appendix_h2_single_robot_symbolic_references} to
compute TG in main-paper Table~2. DeepSeek's 18 runs also provide the
T5 $N=1$ reference in Table~\ref{tab:appendix_h3_team_size}.\par}
\vspace{4pt}
\fontsize{8.2}{8.4}\selectfont
\captionof{table}{Single-robot T5 references for joint fire suppression and rescue with local symbolic observations.}
\label{tab:appendix_h2_single_robot_symbolic_references}
\noindent\begin{tabular*}{\dimexpr\linewidth-10pt\relax}{@{}ll@{\extracolsep{\fill}}*{7}{r}@{}}
\toprule
\textbf{Model} & \textbf{Metric} & \multicolumn{6}{c}{\textbf{Scenario}} & {} \\
\midrule
\multicolumn{2}{@{}l}{\textbf{Local symbolic}} & \makebox[51.4pt][c]{\textbf{S1R}} & \makebox[51.4pt][c]{\textbf{S2R}} & \makebox[51.4pt][c]{\textbf{S3R}} & \makebox[51.4pt][c]{\textbf{S4R}} & \makebox[51.4pt][c]{\textbf{S5R}} & \makebox[51.4pt][c]{\textbf{S6R}} & \makebox[51.4pt][c]{\textbf{Overall}} \\
GPT-6.1 & SVR (\%) & \HtwoScenarioScore{3.13}{0.04} & \HtwoScenarioScore{53.00}{0.01} & \HtwoScenarioScore{9.77}{2.38} & \HtwoScenarioScore{10.54}{1.23} & \HtwoScenarioScore{4.64}{1.40} & \HtwoScenarioScore{6.58}{5.98} & \HtwoScenarioScore{14.61}{2.74} \\
 & FCT (s) & \HtwoScenarioScore{566.7}{57.7} & \HtwoScenarioScore{500.0}{0.0} & \HtwoScenarioScore{2666.7}{321.5} & \HtwoScenarioScore{3033.3}{57.7} & \HtwoScenarioScore{2433.3}{57.7} & \HtwoScenarioScore{1566.7}{493.3} & \HtwoScenarioScore{1794.4}{243.8} \\
 & HPR (\%) & \HtwoScenarioScore{1.33}{0.13} & \HtwoScenarioScore{40.95}{0.69} & \HtwoScenarioScore{61.38}{2.35} & \HtwoScenarioScore{7.07}{4.58} & \HtwoScenarioScore{66.31}{30.05} & \HtwoScenarioScore{0.00}{0.00} & \HtwoScenarioScore{29.51}{12.45} \\
 & CRR (\%) & \HtwoScenarioScore{50.00}{0.00} & \HtwoScenarioScore{50.00}{0.00} & \HtwoScenarioScore{100.00}{0.00} & \HtwoScenarioScore{50.00}{0.00} & \HtwoScenarioScore{100.00}{0.00} & \HtwoScenarioScore{50.00}{0.00} & \HtwoScenarioScore{66.67}{0.00} \\
 & EC (\%) & \HtwoScenarioScore{82.86}{2.63} & \HtwoScenarioScore{58.96}{8.24} & \HtwoScenarioScore{65.09}{1.82} & \HtwoScenarioScore{60.13}{4.01} & \HtwoScenarioScore{54.72}{12.20} & \HtwoScenarioScore{52.71}{5.72} & \HtwoScenarioScore{62.41}{6.78} \\
\addlinespace[0.6pt]
GPT-6 & SVR (\%) & \HtwoScenarioScore{8.46}{6.80} & \HtwoScenarioScore{53.83}{0.29} & \HtwoScenarioScore{5.56}{1.85} & \HtwoScenarioScore{21.68}{18.49} & \HtwoScenarioScore{35.12}{1.68} & \HtwoScenarioScore{3.71}{1.34} & \HtwoScenarioScore{21.39}{8.13} \\
 & FCT (s) & \HtwoScenarioScore{422.7}{167.0} & \HtwoScenarioScore{400.0}{0.0} & \HtwoScenarioScore{2900.0}{100.0} & \HtwoScenarioScore{2566.7}{1270.2} & \HtwoScenarioScore{500.0}{100.0} & \HtwoScenarioScore{1400.0}{200.0} & \HtwoScenarioScore{1364.9}{532.5} \\
 & HPR (\%) & \HtwoScenarioScore{2.31}{1.09} & \HtwoScenarioScore{41.34}{0.92} & \HtwoScenarioScore{15.04}{2.76} & \HtwoScenarioScore{7.58}{13.13} & \HtwoScenarioScore{72.49}{1.44} & \HtwoScenarioScore{33.47}{31.16} & \HtwoScenarioScore{28.70}{13.87} \\
 & CRR (\%) & \HtwoScenarioScore{16.67}{28.87} & \HtwoScenarioScore{50.00}{0.00} & \HtwoScenarioScore{16.67}{28.87} & \HtwoScenarioScore{16.67}{28.87} & \HtwoScenarioScore{50.00}{0.00} & \HtwoScenarioScore{50.00}{0.00} & \HtwoScenarioScore{33.33}{20.41} \\
 & EC (\%) & \HtwoScenarioScore{47.83}{26.45} & \HtwoScenarioScore{51.57}{7.82} & \HtwoScenarioScore{52.13}{6.97} & \HtwoScenarioScore{35.38}{3.65} & \HtwoScenarioScore{25.07}{2.15} & \HtwoScenarioScore{33.70}{12.86} & \HtwoScenarioScore{40.95}{12.86} \\
\addlinespace[0.6pt]
Claude & SVR (\%) & \HtwoScenarioScore{6.43}{8.21} & \HtwoScenarioScore{40.57}{21.28} & \HtwoScenarioScore{1.73}{1.25} & \HtwoScenarioScore{18.84}{15.85} & \HtwoScenarioScore{4.07}{1.36} & \HtwoScenarioScore{1.09}{0.78} & \HtwoScenarioScore{12.12}{11.37} \\
 & FCT (s) & \HtwoScenarioScore{566.7}{152.8} & \HtwoScenarioScore{900.0}{692.8} & \HtwoScenarioScore{2466.7}{152.8} & \HtwoScenarioScore{1866.7}{814.5} & \HtwoScenarioScore{2366.7}{57.7} & \HtwoScenarioScore{1600.0}{100.0} & \HtwoScenarioScore{1627.8}{447.8} \\
 & HPR (\%) & \HtwoScenarioScore{0.38}{1.10} & \HtwoScenarioScore{39.69}{1.52} & \HtwoScenarioScore{2.83}{1.89} & \HtwoScenarioScore{27.94}{3.06} & \HtwoScenarioScore{35.22}{35.24} & \HtwoScenarioScore{63.32}{6.80} & \HtwoScenarioScore{28.23}{14.74} \\
 & CRR (\%) & \HtwoScenarioScore{16.67}{28.87} & \HtwoScenarioScore{50.00}{0.00} & \HtwoScenarioScore{16.67}{28.87} & \HtwoScenarioScore{50.00}{0.00} & \HtwoScenarioScore{66.67}{28.87} & \HtwoScenarioScore{50.00}{0.00} & \HtwoScenarioScore{41.67}{20.41} \\
 & EC (\%) & \HtwoScenarioScore{40.32}{6.17} & \HtwoScenarioScore{50.60}{14.82} & \HtwoScenarioScore{45.17}{21.61} & \HtwoScenarioScore{32.60}{1.23} & \HtwoScenarioScore{35.18}{9.58} & \HtwoScenarioScore{27.84}{2.26} & \HtwoScenarioScore{38.62}{11.71} \\
\addlinespace[0.6pt]
DeepSeek & SVR (\%) & \HtwoScenarioScore{8.03}{8.21} & \HtwoScenarioScore{53.56}{0.47} & \HtwoScenarioScore{13.64}{12.86} & \HtwoScenarioScore{10.46}{14.45} & \HtwoScenarioScore{5.19}{2.71} & \HtwoScenarioScore{5.75}{3.98} & \HtwoScenarioScore{16.11}{8.80} \\
 & FCT (s) & \HtwoScenarioScore{533.4}{288.6} & \HtwoScenarioScore{400.0}{0.0} & \HtwoScenarioScore{2433.3}{896.3} & \HtwoScenarioScore{2733.3}{305.5} & \HtwoScenarioScore{2533.3}{115.5} & \HtwoScenarioScore{1566.7}{351.2} & \HtwoScenarioScore{1700.0}{431.4} \\
 & HPR (\%) & \HtwoScenarioScore{5.69}{4.88} & \HtwoScenarioScore{41.57}{0.28} & \HtwoScenarioScore{42.06}{35.81} & \HtwoScenarioScore{11.68}{20.23} & \HtwoScenarioScore{34.24}{35.57} & \HtwoScenarioScore{25.28}{43.78} & \HtwoScenarioScore{26.75}{28.57} \\
 & CRR (\%) & \HtwoScenarioScore{33.33}{57.74} & \HtwoScenarioScore{50.00}{0.00} & \HtwoScenarioScore{66.67}{28.87} & \HtwoScenarioScore{16.67}{28.87} & \HtwoScenarioScore{66.67}{28.87} & \HtwoScenarioScore{50.00}{0.00} & \HtwoScenarioScore{47.22}{31.18} \\
 & EC (\%) & \HtwoScenarioScore{33.63}{4.66} & \HtwoScenarioScore{37.67}{15.26} & \HtwoScenarioScore{39.63}{10.20} & \HtwoScenarioScore{32.55}{4.35} & \HtwoScenarioScore{35.62}{8.24} & \HtwoScenarioScore{43.06}{13.42} & \HtwoScenarioScore{37.02}{10.21} \\
\addlinespace[0.6pt]
GLM & SVR (\%) & \HtwoScenarioScore{2.41}{0.74} & \HtwoScenarioScore{53.60}{0.63} & \HtwoScenarioScore{4.77}{0.94} & \HtwoScenarioScore{21.57}{10.84} & \HtwoScenarioScore{3.73}{1.57} & \HtwoScenarioScore{2.52}{0.51} & \HtwoScenarioScore{14.77}{4.51} \\
 & FCT (s) & \HtwoScenarioScore{700.0}{0.0} & \HtwoScenarioScore{433.3}{57.7} & \HtwoScenarioScore{2733.3}{288.7} & \HtwoScenarioScore{2133.3}{650.6} & \HtwoScenarioScore{2400.0}{0.0} & \HtwoScenarioScore{1666.7}{57.7} & \HtwoScenarioScore{1677.8}{292.5} \\
 & HPR (\%) & \HtwoScenarioScore{1.98}{0.32} & \HtwoScenarioScore{40.78}{1.70} & \HtwoScenarioScore{12.49}{12.56} & \HtwoScenarioScore{16.66}{16.69} & \HtwoScenarioScore{80.25}{1.30} & \HtwoScenarioScore{40.76}{37.56} & \HtwoScenarioScore{32.15}{17.57} \\
 & CRR (\%) & \HtwoScenarioScore{0.00}{0.00} & \HtwoScenarioScore{50.00}{0.00} & \HtwoScenarioScore{33.33}{28.87} & \HtwoScenarioScore{33.33}{28.87} & \HtwoScenarioScore{100.00}{0.00} & \HtwoScenarioScore{33.33}{28.87} & \HtwoScenarioScore{41.67}{20.41} \\
 & EC (\%) & \HtwoScenarioScore{48.64}{10.04} & \HtwoScenarioScore{41.61}{18.90} & \HtwoScenarioScore{48.39}{16.64} & \HtwoScenarioScore{36.50}{5.69} & \HtwoScenarioScore{47.12}{0.44} & \HtwoScenarioScore{34.08}{12.54} & \HtwoScenarioScore{42.72}{12.42} \\
\bottomrule
\end{tabular*}\par
\newpage
{\fontsize{11}{12}\selectfont\bfseries H.2. Additional Visual Conditions and Single-Robot References (continued)\par}
\vspace{5pt}
{\fontsize{8.3}{9.5}\selectfont
\emph{Visual trial protocol.}
Table~\ref{tab:appendix_h2_single_robot_visual_references} covers 180 single-robot T5 trials
and Table~\ref{tab:appendix_h2_four_robot_visual_suppression} covers 72 four-robot T4 trials.
Each condition used three trials in each of six scenarios, eightfold suppression,
and the visual timing protocol in Appendix~\ref{appendix:visual_trial_protocols}.
For TG, we use the matched team and single-robot trials defined in
Appendix~\ref{appendix:evaluation_protocol}.
Throughout this section, statistics follow Appendix~\ref{appendix:complete_results}
and FCT uses the penalty-adjusted score in Appendix~\ref{appendix:metrics}.\par}
\vspace{3pt}
{\fontsize{8.3}{9.5}\selectfont
In visual T5, one FR uses the team's local RGB or RGB+DS interface for
suppression and rescue. DeepSeek's mean RGB SVR is 12.76\% with four FRs
and 2.33\% with one, yielding a TG of 10.43 percentage points (main-paper
Table~3). The DeepSeek visual T5 conditions in main-paper Table~4 use the
same single-robot references.\par}
\vspace{2pt}
\fontsize{8.2}{8.4}\selectfont
\captionof{table}{Single-robot visual references for joint fire suppression and rescue in T5.}
\label{tab:appendix_h2_single_robot_visual_references}
\begingroup
\renewcommand{\arraystretch}{0.985}
\noindent\begin{tabular*}{\dimexpr\linewidth-10pt\relax}{@{}ll@{\extracolsep{\fill}}*{7}{r}@{}}
\toprule
\textbf{Model} & \textbf{Metric} & \multicolumn{6}{c}{\textbf{Scenario}} & {} \\
\midrule
\multicolumn{2}{@{}l}{\textbf{RGB}} & \makebox[51.4pt][c]{\textbf{S1R}} & \makebox[51.4pt][c]{\textbf{S2R}} & \makebox[51.4pt][c]{\textbf{S3R}} & \makebox[51.4pt][c]{\textbf{S4R}} & \makebox[51.4pt][c]{\textbf{S5R}} & \makebox[51.4pt][c]{\textbf{S6R}} & \makebox[51.4pt][c]{\textbf{Overall}} \\
GPT-6.1 & SVR (\%) & \HtwoScenarioScore{4.38}{3.19} & \HtwoScenarioScore{7.54}{0.99} & \HtwoScenarioScore{9.50}{0.05} & \HtwoScenarioScore{12.40}{6.73} & \HtwoScenarioScore{14.64}{15.52} & \HtwoScenarioScore{7.99}{5.02} & \HtwoScenarioScore{9.41}{7.33} \\
 & FCT (s) & \HtwoScenarioScore{566.7}{152.8} & \HtwoScenarioScore{1133.3}{57.7} & \HtwoScenarioScore{2100.0}{0.0} & \HtwoScenarioScore{2233.3}{723.4} & \HtwoScenarioScore{1833.3}{1069.3} & \HtwoScenarioScore{1166.7}{208.2} & \HtwoScenarioScore{1505.6}{538.0} \\
 & HPR (\%) & \HtwoScenarioScore{1.78}{0.21} & \HtwoScenarioScore{27.09}{0.13} & \HtwoScenarioScore{19.20}{8.28} & \HtwoScenarioScore{0.00}{0.00} & \HtwoScenarioScore{53.41}{33.02} & \HtwoScenarioScore{13.09}{22.68} & \HtwoScenarioScore{19.10}{16.70} \\
 & CRR (\%) & \HtwoScenarioScore{0.00}{0.00} & \HtwoScenarioScore{50.00}{0.00} & \HtwoScenarioScore{100.00}{0.00} & \HtwoScenarioScore{0.00}{0.00} & \HtwoScenarioScore{66.67}{28.87} & \HtwoScenarioScore{16.67}{28.87} & \HtwoScenarioScore{38.89}{16.67} \\
 & EC (\%) & \HtwoScenarioScore{64.75}{8.41} & \HtwoScenarioScore{58.39}{0.28} & \HtwoScenarioScore{49.21}{2.70} & \HtwoScenarioScore{62.87}{1.99} & \HtwoScenarioScore{42.04}{12.50} & \HtwoScenarioScore{29.32}{7.10} & \HtwoScenarioScore{51.10}{6.94} \\
\addlinespace[0.6pt]
GPT-6 & SVR (\%) & \HtwoScenarioScore{0.65}{0.21} & \HtwoScenarioScore{18.72}{30.62} & \HtwoScenarioScore{3.45}{4.17} & \HtwoScenarioScore{22.40}{19.13} & \HtwoScenarioScore{14.68}{16.06} & \HtwoScenarioScore{1.92}{1.18} & \HtwoScenarioScore{10.30}{16.23} \\
 & FCT (s) & \HtwoScenarioScore{700.0}{0.0} & \HtwoScenarioScore{1233.3}{723.4} & \HtwoScenarioScore{2333.3}{378.6} & \HtwoScenarioScore{1766.7}{1106.0} & \HtwoScenarioScore{1600.0}{818.5} & \HtwoScenarioScore{1633.3}{57.7} & \HtwoScenarioScore{1544.4}{653.6} \\
 & HPR (\%) & \HtwoScenarioScore{0.82}{0.96} & \HtwoScenarioScore{15.24}{18.78} & \HtwoScenarioScore{6.29}{1.81} & \HtwoScenarioScore{12.17}{21.08} & \HtwoScenarioScore{11.52}{18.47} & \HtwoScenarioScore{0.00}{0.00} & \HtwoScenarioScore{7.67}{13.80} \\
 & CRR (\%) & \HtwoScenarioScore{0.00}{0.00} & \HtwoScenarioScore{0.00}{0.00} & \HtwoScenarioScore{0.00}{0.00} & \HtwoScenarioScore{16.67}{28.87} & \HtwoScenarioScore{0.00}{0.00} & \HtwoScenarioScore{0.00}{0.00} & \HtwoScenarioScore{2.78}{11.79} \\
 & EC (\%) & \HtwoScenarioScore{29.51}{13.64} & \HtwoScenarioScore{24.18}{1.19} & \HtwoScenarioScore{21.57}{1.57} & \HtwoScenarioScore{25.27}{4.25} & \HtwoScenarioScore{18.73}{0.84} & \HtwoScenarioScore{17.22}{1.67} & \HtwoScenarioScore{22.75}{5.94} \\
\addlinespace[0.6pt]
Claude & SVR (\%) & \HtwoScenarioScore{1.44}{1.22} & \HtwoScenarioScore{17.37}{30.79} & \HtwoScenarioScore{5.35}{2.41} & \HtwoScenarioScore{18.77}{20.17} & \HtwoScenarioScore{1.78}{1.50} & \HtwoScenarioScore{6.06}{6.17} & \HtwoScenarioScore{8.46}{15.29} \\
 & FCT (s) & \HtwoScenarioScore{633.3}{115.5} & \HtwoScenarioScore{1300.0}{692.8} & \HtwoScenarioScore{2200.0}{264.6} & \HtwoScenarioScore{2166.7}{1097.0} & \HtwoScenarioScore{2300.0}{0.0} & \HtwoScenarioScore{1500.0}{173.2} & \HtwoScenarioScore{1683.3}{547.2} \\
 & HPR (\%) & \HtwoScenarioScore{1.40}{0.26} & \HtwoScenarioScore{13.10}{15.86} & \HtwoScenarioScore{6.07}{0.95} & \HtwoScenarioScore{0.00}{0.00} & \HtwoScenarioScore{36.78}{32.70} & \HtwoScenarioScore{0.00}{0.00} & \HtwoScenarioScore{9.56}{14.84} \\
 & CRR (\%) & \HtwoScenarioScore{0.00}{0.00} & \HtwoScenarioScore{50.00}{0.00} & \HtwoScenarioScore{0.00}{0.00} & \HtwoScenarioScore{0.00}{0.00} & \HtwoScenarioScore{33.33}{28.87} & \HtwoScenarioScore{0.00}{0.00} & \HtwoScenarioScore{13.89}{11.79} \\
 & EC (\%) & \HtwoScenarioScore{31.71}{2.27} & \HtwoScenarioScore{20.64}{1.09} & \HtwoScenarioScore{21.27}{0.34} & \HtwoScenarioScore{27.96}{5.45} & \HtwoScenarioScore{18.81}{1.76} & \HtwoScenarioScore{16.63}{1.05} & \HtwoScenarioScore{22.84}{2.59} \\
\addlinespace[0.6pt]
DeepSeek & SVR (\%) & \HtwoScenarioScore{0.94}{0.13} & \HtwoScenarioScore{1.91}{1.07} & \HtwoScenarioScore{2.11}{1.03} & \HtwoScenarioScore{3.99}{2.74} & \HtwoScenarioScore{3.95}{0.12} & \HtwoScenarioScore{1.05}{0.63} & \HtwoScenarioScore{2.33}{1.30} \\
 & FCT (s) & \HtwoScenarioScore{700.0}{0.0} & \HtwoScenarioScore{1833.3}{115.5} & \HtwoScenarioScore{2600.0}{173.2} & \HtwoScenarioScore{2900.0}{173.2} & \HtwoScenarioScore{2300.0}{0.0} & \HtwoScenarioScore{1700.0}{0.0} & \HtwoScenarioScore{2005.6}{110.6} \\
 & HPR (\%) & \HtwoScenarioScore{1.97}{0.26} & \HtwoScenarioScore{3.97}{1.09} & \HtwoScenarioScore{7.08}{1.91} & \HtwoScenarioScore{0.00}{0.00} & \HtwoScenarioScore{0.67}{0.09} & \HtwoScenarioScore{0.00}{0.00} & \HtwoScenarioScore{2.28}{0.90} \\
 & CRR (\%) & \HtwoScenarioScore{0.00}{0.00} & \HtwoScenarioScore{0.00}{0.00} & \HtwoScenarioScore{0.00}{0.00} & \HtwoScenarioScore{0.00}{0.00} & \HtwoScenarioScore{0.00}{0.00} & \HtwoScenarioScore{0.00}{0.00} & \HtwoScenarioScore{0.00}{0.00} \\
 & EC (\%) & \HtwoScenarioScore{41.80}{2.63} & \HtwoScenarioScore{40.64}{5.72} & \HtwoScenarioScore{38.28}{3.19} & \HtwoScenarioScore{40.46}{2.55} & \HtwoScenarioScore{43.81}{3.66} & \HtwoScenarioScore{36.15}{1.23} & \HtwoScenarioScore{40.19}{3.45} \\
\addlinespace[0.6pt]
GLM & SVR (\%) & \HtwoScenarioScore{9.52}{9.25} & \HtwoScenarioScore{35.82}{30.04} & \HtwoScenarioScore{8.36}{6.07} & \HtwoScenarioScore{8.00}{0.23} & \HtwoScenarioScore{9.72}{2.80} & \HtwoScenarioScore{9.92}{5.96} & \HtwoScenarioScore{13.56}{13.34} \\
 & FCT (s) & \HtwoScenarioScore{358.5}{216.2} & \HtwoScenarioScore{933.3}{838.6} & \HtwoScenarioScore{2300.0}{300.0} & \HtwoScenarioScore{2833.3}{57.7} & \HtwoScenarioScore{1800.0}{0.0} & \HtwoScenarioScore{1200.0}{458.3} & \HtwoScenarioScore{1570.9}{419.0} \\
 & HPR (\%) & \HtwoScenarioScore{1.66}{0.18} & \HtwoScenarioScore{25.05}{19.08} & \HtwoScenarioScore{10.24}{3.96} & \HtwoScenarioScore{0.00}{0.00} & \HtwoScenarioScore{69.22}{0.71} & \HtwoScenarioScore{0.00}{0.00} & \HtwoScenarioScore{17.69}{7.96} \\
 & CRR (\%) & \HtwoScenarioScore{0.00}{0.00} & \HtwoScenarioScore{33.33}{28.87} & \HtwoScenarioScore{0.00}{0.00} & \HtwoScenarioScore{0.00}{0.00} & \HtwoScenarioScore{50.00}{0.00} & \HtwoScenarioScore{0.00}{0.00} & \HtwoScenarioScore{13.89}{11.79} \\
 & EC (\%) & \HtwoScenarioScore{30.10}{5.08} & \HtwoScenarioScore{32.05}{11.53} & \HtwoScenarioScore{29.66}{12.46} & \HtwoScenarioScore{26.67}{8.00} & \HtwoScenarioScore{17.99}{0.64} & \HtwoScenarioScore{21.01}{1.34} & \HtwoScenarioScore{26.25}{7.96} \\
\midrule
\multicolumn{2}{@{}l}{\textbf{RGB+DS}} & \makebox[51.4pt][c]{\textbf{S1R}} & \makebox[51.4pt][c]{\textbf{S2R}} & \makebox[51.4pt][c]{\textbf{S3R}} & \makebox[51.4pt][c]{\textbf{S4R}} & \makebox[51.4pt][c]{\textbf{S5R}} & \makebox[51.4pt][c]{\textbf{S6R}} & \makebox[51.4pt][c]{\textbf{Overall}} \\
GPT-6.1 & SVR (\%) & \HtwoScenarioScore{4.06}{0.45} & \HtwoScenarioScore{3.23}{0.26} & \HtwoScenarioScore{11.50}{4.99} & \HtwoScenarioScore{11.30}{7.10} & \HtwoScenarioScore{5.19}{1.21} & \HtwoScenarioScore{16.38}{1.30} & \HtwoScenarioScore{8.61}{3.62} \\
 & FCT (s) & \HtwoScenarioScore{366.0}{58.8} & \HtwoScenarioScore{1400.0}{173.2} & \HtwoScenarioScore{1866.7}{493.3} & \HtwoScenarioScore{2600.0}{519.6} & \HtwoScenarioScore{2400.0}{0.0} & \HtwoScenarioScore{633.3}{152.8} & \HtwoScenarioScore{1544.3}{308.3} \\
 & HPR (\%) & \HtwoScenarioScore{1.79}{0.17} & \HtwoScenarioScore{19.19}{0.63} & \HtwoScenarioScore{12.83}{0.72} & \HtwoScenarioScore{0.00}{0.00} & \HtwoScenarioScore{13.82}{1.42} & \HtwoScenarioScore{0.00}{0.00} & \HtwoScenarioScore{7.94}{0.70} \\
 & CRR (\%) & \HtwoScenarioScore{0.00}{0.00} & \HtwoScenarioScore{50.00}{0.00} & \HtwoScenarioScore{50.00}{0.00} & \HtwoScenarioScore{0.00}{0.00} & \HtwoScenarioScore{50.00}{0.00} & \HtwoScenarioScore{0.00}{0.00} & \HtwoScenarioScore{25.00}{0.00} \\
 & EC (\%) & \HtwoScenarioScore{58.43}{2.36} & \HtwoScenarioScore{58.47}{0.56} & \HtwoScenarioScore{45.32}{1.82} & \HtwoScenarioScore{60.03}{0.21} & \HtwoScenarioScore{33.92}{0.71} & \HtwoScenarioScore{21.46}{0.13} & \HtwoScenarioScore{46.27}{1.27} \\
\addlinespace[0.6pt]
GPT-6 & SVR (\%) & \HtwoScenarioScore{14.80}{7.98} & \HtwoScenarioScore{1.22}{0.24} & \HtwoScenarioScore{9.71}{5.98} & \HtwoScenarioScore{16.48}{22.40} & \HtwoScenarioScore{21.92}{17.42} & \HtwoScenarioScore{8.84}{6.07} & \HtwoScenarioScore{12.16}{12.53} \\
 & FCT (s) & \HtwoScenarioScore{288.9}{182.8} & \HtwoScenarioScore{1566.7}{152.8} & \HtwoScenarioScore{2566.7}{550.8} & \HtwoScenarioScore{2166.7}{1097.0} & \HtwoScenarioScore{1100.0}{871.8} & \HtwoScenarioScore{1133.3}{230.9} & \HtwoScenarioScore{1470.4}{629.4} \\
 & HPR (\%) & \HtwoScenarioScore{1.29}{0.10} & \HtwoScenarioScore{24.14}{19.45} & \HtwoScenarioScore{12.19}{15.08} & \HtwoScenarioScore{0.72}{1.25} & \HtwoScenarioScore{22.25}{18.87} & \HtwoScenarioScore{0.00}{0.00} & \HtwoScenarioScore{10.10}{12.67} \\
 & CRR (\%) & \HtwoScenarioScore{0.00}{0.00} & \HtwoScenarioScore{33.33}{28.87} & \HtwoScenarioScore{16.67}{28.87} & \HtwoScenarioScore{0.00}{0.00} & \HtwoScenarioScore{0.00}{0.00} & \HtwoScenarioScore{0.00}{0.00} & \HtwoScenarioScore{8.33}{16.67} \\
 & EC (\%) & \HtwoScenarioScore{29.14}{8.81} & \HtwoScenarioScore{32.05}{7.49} & \HtwoScenarioScore{27.64}{1.62} & \HtwoScenarioScore{25.85}{0.77} & \HtwoScenarioScore{20.50}{0.34} & \HtwoScenarioScore{20.12}{1.12} & \HtwoScenarioScore{25.88}{4.80} \\
\addlinespace[0.6pt]
Claude & SVR (\%) & \HtwoScenarioScore{1.40}{1.33} & \HtwoScenarioScore{0.01}{0.80} & \HtwoScenarioScore{3.37}{4.84} & \HtwoScenarioScore{3.74}{7.36} & \HtwoScenarioScore{6.07}{3.01} & \HtwoScenarioScore{5.47}{7.41} & \HtwoScenarioScore{3.35}{4.90} \\
 & FCT (s) & \HtwoScenarioScore{600.0}{173.2} & \HtwoScenarioScore{1733.3}{152.8} & \HtwoScenarioScore{2366.7}{404.1} & \HtwoScenarioScore{2600.0}{346.4} & \HtwoScenarioScore{1933.3}{230.9} & \HtwoScenarioScore{1466.7}{404.1} & \HtwoScenarioScore{1783.3}{303.7} \\
 & HPR (\%) & \HtwoScenarioScore{0.93}{0.13} & \HtwoScenarioScore{10.35}{22.14} & \HtwoScenarioScore{4.17}{2.27} & \HtwoScenarioScore{0.00}{0.00} & \HtwoScenarioScore{35.71}{31.23} & \HtwoScenarioScore{0.00}{0.00} & \HtwoScenarioScore{8.53}{15.66} \\
 & CRR (\%) & \HtwoScenarioScore{0.00}{0.00} & \HtwoScenarioScore{16.67}{28.87} & \HtwoScenarioScore{0.00}{0.00} & \HtwoScenarioScore{0.00}{0.00} & \HtwoScenarioScore{33.33}{28.87} & \HtwoScenarioScore{33.33}{28.87} & \HtwoScenarioScore{13.89}{20.41} \\
 & EC (\%) & \HtwoScenarioScore{32.45}{3.61} & \HtwoScenarioScore{25.38}{5.30} & \HtwoScenarioScore{30.79}{9.38} & \HtwoScenarioScore{26.54}{4.11} & \HtwoScenarioScore{20.50}{0.68} & \HtwoScenarioScore{27.99}{6.32} & \HtwoScenarioScore{27.27}{5.57} \\
\addlinespace[0.6pt]
DeepSeek & SVR (\%) & \HtwoScenarioScore{2.07}{2.19} & \HtwoScenarioScore{18.22}{30.04} & \HtwoScenarioScore{3.12}{1.27} & \HtwoScenarioScore{20.66}{18.38} & \HtwoScenarioScore{17.24}{16.70} & \HtwoScenarioScore{1.77}{1.00} & \HtwoScenarioScore{10.51}{15.95} \\
 & FCT (s) & \HtwoScenarioScore{600.0}{173.2} & \HtwoScenarioScore{1366.7}{757.2} & \HtwoScenarioScore{2700.0}{264.6} & \HtwoScenarioScore{2266.7}{1193.0} & \HtwoScenarioScore{1566.7}{1006.6} & \HtwoScenarioScore{1700.0}{0.0} & \HtwoScenarioScore{1700.0}{720.0} \\
 & HPR (\%) & \HtwoScenarioScore{1.71}{0.09} & \HtwoScenarioScore{17.93}{19.50} & \HtwoScenarioScore{7.48}{2.11} & \HtwoScenarioScore{0.00}{0.00} & \HtwoScenarioScore{11.98}{20.75} & \HtwoScenarioScore{12.33}{21.35} & \HtwoScenarioScore{8.57}{14.56} \\
 & CRR (\%) & \HtwoScenarioScore{0.00}{0.00} & \HtwoScenarioScore{33.33}{28.87} & \HtwoScenarioScore{0.00}{0.00} & \HtwoScenarioScore{0.00}{0.00} & \HtwoScenarioScore{16.67}{28.87} & \HtwoScenarioScore{16.67}{28.87} & \HtwoScenarioScore{11.11}{20.41} \\
 & EC (\%) & \HtwoScenarioScore{33.19}{3.13} & \HtwoScenarioScore{34.38}{7.13} & \HtwoScenarioScore{31.16}{4.18} & \HtwoScenarioScore{30.29}{8.97} & \HtwoScenarioScore{27.06}{6.20} & \HtwoScenarioScore{25.17}{5.19} & \HtwoScenarioScore{30.21}{6.11} \\
\addlinespace[0.6pt]
GLM & SVR (\%) & \HtwoScenarioScore{4.83}{3.27} & \HtwoScenarioScore{5.22}{4.21} & \HtwoScenarioScore{12.02}{4.59} & \HtwoScenarioScore{0.27}{7.76} & \HtwoScenarioScore{6.42}{2.37} & \HtwoScenarioScore{3.28}{1.50} & \HtwoScenarioScore{5.34}{4.43} \\
 & FCT (s) & \HtwoScenarioScore{530.4}{213.1} & \HtwoScenarioScore{1433.3}{416.3} & \HtwoScenarioScore{2066.7}{115.5} & \HtwoScenarioScore{2633.3}{208.2} & \HtwoScenarioScore{2100.0}{300.0} & \HtwoScenarioScore{1433.3}{288.7} & \HtwoScenarioScore{1699.5}{273.5} \\
 & HPR (\%) & \HtwoScenarioScore{1.79}{0.38} & \HtwoScenarioScore{23.79}{18.30} & \HtwoScenarioScore{9.20}{4.10} & \HtwoScenarioScore{6.72}{11.64} & \HtwoScenarioScore{23.28}{38.56} & \HtwoScenarioScore{0.00}{0.00} & \HtwoScenarioScore{10.80}{18.14} \\
 & CRR (\%) & \HtwoScenarioScore{0.00}{0.00} & \HtwoScenarioScore{33.33}{28.87} & \HtwoScenarioScore{0.00}{0.00} & \HtwoScenarioScore{16.67}{28.87} & \HtwoScenarioScore{33.33}{28.87} & \HtwoScenarioScore{0.00}{0.00} & \HtwoScenarioScore{13.89}{20.41} \\
 & EC (\%) & \HtwoScenarioScore{29.07}{9.81} & \HtwoScenarioScore{31.89}{10.17} & \HtwoScenarioScore{24.72}{2.02} & \HtwoScenarioScore{26.50}{2.97} & \HtwoScenarioScore{24.41}{4.69} & \HtwoScenarioScore{18.34}{0.34} & \HtwoScenarioScore{25.82}{6.26} \\
\bottomrule
\end{tabular*}\par
\endgroup
\newpage
{\fontsize{11}{12}\selectfont\bfseries H.2. Additional Visual Conditions and Single-Robot References (continued)\par}
\vspace{5pt}
{\fontsize{8.3}{9.5}\selectfont
In the visual variant of T4, four FRs suppress fires in civilian-free
scenarios using local RGB or RGB+DS under DMR with communication.
Matched single-robot visual T3 runs provide the reference for TG.
With RGB+DS, DeepSeek's mean SVR rises from 9.02\% to 19.55\%,
with increases in all six scenarios. GLM's mean falls from 24.67\%
to 22.15\%, with higher SVR in two scenarios and lower SVR in four.\par}
\vspace{4pt}
\fontsize{8.2}{8.5}\selectfont
\captionof{table}{Visual variant of T4. Per-scenario results under RGB and RGB+DS.
Within each interface, bold marks each column's best mean per metric, including ties
(FCT $\downarrow$, others $\uparrow$).}
\label{tab:appendix_h2_four_robot_visual_suppression}
\noindent\begin{tabular*}{\dimexpr\linewidth-10pt\relax}{@{}ll@{\extracolsep{\fill}}*{7}{r}@{}}
\toprule
\textbf{Model} & \textbf{Metric} & \multicolumn{6}{c}{\textbf{Scenario}} & {} \\
\midrule
\multicolumn{2}{@{}l}{\textbf{RGB}} & \makebox[51.4pt][c]{\textbf{S1F}} & \makebox[51.4pt][c]{\textbf{S2F}} & \makebox[51.4pt][c]{\textbf{S3F}} & \makebox[51.4pt][c]{\textbf{S4F}} & \makebox[51.4pt][c]{\textbf{S5F}} & \makebox[51.4pt][c]{\textbf{S6F}} & \makebox[51.4pt][c]{\textbf{Overall}} \\
DeepSeek & SVR (\%) & \HtwoScenarioScore{4.98}{0.57} & \HtwoScenarioScore{6.42}{0.68} & \HtwoScenarioScore{16.72}{2.11} & \HtwoScenarioScore{13.25}{9.49} & \HtwoScenarioScore{7.82}{1.33} & \HtwoScenarioScore{4.94}{0.38} & \HtwoScenarioScore{9.02}{4.03} \\
 & FCT (s) & \HtwoScenarioScore{600.0}{0.0} & \HtwoScenarioScore{2000.0}{0.0} & \HtwoScenarioScore{3000.0}{360.6} & \HtwoScenarioScore{3866.7}{321.5} & \HtwoScenarioScore{2733.3}{57.7} & \HtwoScenarioScore{2033.3}{115.5} & \HtwoScenarioScore{2372.2}{204.1} \\
 & EC (\%) & \HtwoScenarioScore{\textbf{76.97}}{12.27} & \HtwoScenarioScore{\textbf{86.43}}{2.99} & \HtwoScenarioScore{\textbf{81.12}}{2.14} & \HtwoScenarioScore{\textbf{83.13}}{5.40} & \HtwoScenarioScore{\textbf{77.88}}{11.08} & \HtwoScenarioScore{\textbf{71.94}}{12.26} & \HtwoScenarioScore{\textbf{79.58}}{8.82} \\
 & TG (pp) & \HtwoScenarioScore{\textbf{4.39}}{1.04} & \HtwoScenarioScore{\textbf{4.91}}{1.31} & \HtwoScenarioScore{\textbf{14.33}}{2.27} & \HtwoScenarioScore{-0.62}{15.42} & \HtwoScenarioScore{3.99}{1.62} & \HtwoScenarioScore{\textbf{2.68}}{1.03} & \HtwoScenarioScore{4.95}{6.45} \\
\addlinespace[0.6pt]
GLM & SVR (\%) & \HtwoScenarioScore{\textbf{5.69}}{0.29} & \HtwoScenarioScore{\textbf{38.24}}{27.05} & \HtwoScenarioScore{\textbf{17.86}}{2.24} & \HtwoScenarioScore{\textbf{47.74}}{13.44} & \HtwoScenarioScore{\textbf{27.22}}{11.58} & \HtwoScenarioScore{\textbf{11.30}}{3.73} & \HtwoScenarioScore{\textbf{24.67}}{13.33} \\
 & FCT (s) & \HtwoScenarioScore{\textbf{500.0}}{0.0} & \HtwoScenarioScore{\textbf{798.8}}{782.2} & \HtwoScenarioScore{\textbf{2800.0}}{346.4} & \HtwoScenarioScore{\textbf{1233.3}}{503.3} & \HtwoScenarioScore{\textbf{1333.3}}{1274.1} & \HtwoScenarioScore{\textbf{1133.3}}{208.2} & \HtwoScenarioScore{\textbf{1299.8}}{664.8} \\
 & EC (\%) & \HtwoScenarioScore{74.98}{5.73} & \HtwoScenarioScore{67.31}{10.36} & \HtwoScenarioScore{67.19}{7.83} & \HtwoScenarioScore{63.13}{9.96} & \HtwoScenarioScore{59.88}{7.92} & \HtwoScenarioScore{59.24}{4.05} & \HtwoScenarioScore{65.29}{7.96} \\
 & TG (pp) & \HtwoScenarioScore{3.17}{1.49} & \HtwoScenarioScore{1.96}{50.10} & \HtwoScenarioScore{9.14}{1.96} & \HtwoScenarioScore{\textbf{39.66}}{15.50} & \HtwoScenarioScore{\textbf{11.67}}{10.55} & \HtwoScenarioScore{1.97}{5.28} & \HtwoScenarioScore{\textbf{11.26}}{21.97} \\
\midrule
\multicolumn{2}{@{}l}{\textbf{RGB+DS}} & \makebox[51.4pt][c]{\textbf{S1F}} & \makebox[51.4pt][c]{\textbf{S2F}} & \makebox[51.4pt][c]{\textbf{S3F}} & \makebox[51.4pt][c]{\textbf{S4F}} & \makebox[51.4pt][c]{\textbf{S5F}} & \makebox[51.4pt][c]{\textbf{S6F}} & \makebox[51.4pt][c]{\textbf{Overall}} \\
DeepSeek & SVR (\%) & \HtwoScenarioScore{\textbf{13.16}}{16.08} & \HtwoScenarioScore{7.58}{3.16} & \HtwoScenarioScore{\textbf{30.08}}{25.73} & \HtwoScenarioScore{\textbf{37.87}}{25.55} & \HtwoScenarioScore{20.10}{14.24} & \HtwoScenarioScore{\textbf{8.52}}{5.81} & \HtwoScenarioScore{19.55}{17.42} \\
 & FCT (s) & \HtwoScenarioScore{424.8}{222.6} & \HtwoScenarioScore{1966.7}{57.7} & \HtwoScenarioScore{1900.0}{1179.0} & \HtwoScenarioScore{2633.3}{808.3} & \HtwoScenarioScore{2000.0}{1135.8} & \HtwoScenarioScore{1400.0}{519.6} & \HtwoScenarioScore{1720.8}{780.6} \\
 & EC (\%) & \HtwoScenarioScore{61.81}{9.28} & \HtwoScenarioScore{63.69}{3.74} & \HtwoScenarioScore{\textbf{57.00}}{3.40} & \HtwoScenarioScore{55.63}{16.01} & \HtwoScenarioScore{\textbf{61.21}}{4.97} & \HtwoScenarioScore{\textbf{53.08}}{4.22} & \HtwoScenarioScore{\textbf{58.74}}{8.27} \\
 & TG (pp) & \HtwoScenarioScore{\textbf{9.68}}{17.67} & \HtwoScenarioScore{\textbf{6.52}}{3.75} & \HtwoScenarioScore{11.30}{33.83} & \HtwoScenarioScore{\textbf{20.88}}{47.74} & \HtwoScenarioScore{7.11}{29.00} & \HtwoScenarioScore{\textbf{6.96}}{6.05} & \HtwoScenarioScore{10.41}{27.77} \\
\addlinespace[0.6pt]
GLM & SVR (\%) & \HtwoScenarioScore{8.12}{3.75} & \HtwoScenarioScore{\textbf{37.02}}{28.09} & \HtwoScenarioScore{23.53}{6.48} & \HtwoScenarioScore{29.04}{28.59} & \HtwoScenarioScore{\textbf{27.15}}{9.91} & \HtwoScenarioScore{8.00}{5.86} & \HtwoScenarioScore{\textbf{22.15}}{17.30} \\
 & FCT (s) & \HtwoScenarioScore{\textbf{390.7}}{114.2} & \HtwoScenarioScore{\textbf{866.7}}{808.3} & \HtwoScenarioScore{\textbf{1800.0}}{458.3} & \HtwoScenarioScore{\textbf{2300.0}}{1609.3} & \HtwoScenarioScore{\textbf{1033.3}}{665.8} & \HtwoScenarioScore{\textbf{1333.3}}{568.6} & \HtwoScenarioScore{\textbf{1287.3}}{839.9} \\
 & EC (\%) & \HtwoScenarioScore{\textbf{66.23}}{4.65} & \HtwoScenarioScore{\textbf{63.94}}{10.76} & \HtwoScenarioScore{55.28}{5.99} & \HtwoScenarioScore{\textbf{56.17}}{3.69} & \HtwoScenarioScore{58.63}{5.06} & \HtwoScenarioScore{50.63}{5.56} & \HtwoScenarioScore{58.48}{6.37} \\
 & TG (pp) & \HtwoScenarioScore{5.77}{4.08} & \HtwoScenarioScore{1.04}{50.47} & \HtwoScenarioScore{\textbf{18.27}}{8.30} & \HtwoScenarioScore{14.43}{2.72} & \HtwoScenarioScore{\textbf{24.15}}{8.72} & \HtwoScenarioScore{6.75}{6.64} & \HtwoScenarioScore{\textbf{11.73}}{21.45} \\
\bottomrule
\end{tabular*}\par
\endgroup

\newpage

\subsection{Coordination Configurations, Results, and Execution Traces}
\label{appendix:coordination_details}

Tables~\ref{tab:appendix_h3_control_communication},
\ref{tab:appendix_h3_team_size}, and \ref{tab:appendix_h3_role_composition}
give per-scenario results for the control and communication conditions in
main-paper Table~4, the team sizes in Fig.~3(a), and the role compositions
in Fig.~3(b), respectively. On T5, HPR was generally lower with broadcasts
enabled. The tables also report FCT and, for the T5 team-size comparison, HPR.
\par\vspace{5pt}

\begingroup
\setlength{\parindent}{0pt}
\setlength{\parskip}{0pt}
\setlength{\tabcolsep}{2pt}
\setlength{\aboverulesep}{1.1pt}
\setlength{\belowrulesep}{1.1pt}
\hfuzz=0.1pt
\vfuzz=0.1pt
\captionsetup{hypcap=false,font=footnotesize,labelfont=bf,position=top,justification=raggedright,singlelinecheck=false,skip=2.2pt}
\newcommand{\HThreeScenarioScore}[2]{%
  \makebox[22pt][r]{#1}\kern0.8pt\textpm\kern0.8pt%
  \makebox[22pt][l]{\textcolor{black!76}{#2}}%
}
{\fontsize{8.3}{9.5}\selectfont Scenario and Overall statistics follow Appendix~\ref{appendix:complete_results}.\par}
\vspace{4pt}
\fontsize{8.2}{8.5}\selectfont
\captionof{table}{Per-scenario control and communication results (Table~4 of the main paper).}
\label{tab:appendix_h3_control_communication}
\noindent\begin{tabular*}{\dimexpr\linewidth-10pt\relax}{@{}ll@{\extracolsep{\fill}}*{7}{r}@{}}
\toprule
\textbf{Condition} & \textbf{Metric} & \multicolumn{6}{c}{\textbf{Scenario}} & {} \\
\midrule
\multicolumn{2}{@{}l}{\textbf{T4. DeepSeek, Symbolic}} & \makebox[51.4pt][c]{\textbf{S1F}} & \makebox[51.4pt][c]{\textbf{S2F}} & \makebox[51.4pt][c]{\textbf{S3F}} & \makebox[51.4pt][c]{\textbf{S4F}} & \makebox[51.4pt][c]{\textbf{S5F}} & \makebox[51.4pt][c]{\textbf{S6F}} & \makebox[51.4pt][c]{\textbf{Overall}} \\
CMR & SVR (\%) & \HThreeScenarioScore{11.40}{15.07} & \HThreeScenarioScore{57.81}{6.52} & \HThreeScenarioScore{24.70}{4.12} & \HThreeScenarioScore{46.59}{42.64} & \HThreeScenarioScore{17.19}{23.27} & \HThreeScenarioScore{10.79}{6.44} & \HThreeScenarioScore{28.08}{21.16} \\
 & FCT (s) & \HThreeScenarioScore{485.1}{290.0} & \HThreeScenarioScore{328.5}{148.5} & \HThreeScenarioScore{1600.0}{871.8} & \HThreeScenarioScore{1358.7}{1299.6} & \HThreeScenarioScore{1527.7}{1088.2} & \HThreeScenarioScore{1033.3}{230.9} & \HThreeScenarioScore{1055.6}{795.0} \\
 & TG (pp) & \HThreeScenarioScore{-19.18}{20.63} & \HThreeScenarioScore{7.64}{23.14} & \HThreeScenarioScore{-5.49}{3.09} & \HThreeScenarioScore{8.82}{70.06} & \HThreeScenarioScore{-22.12}{26.27} & \HThreeScenarioScore{3.78}{8.67} & \HThreeScenarioScore{-4.43}{33.28} \\
\addlinespace[0.6pt]
DMR & SVR (\%) & \HThreeScenarioScore{38.78}{1.86} & \HThreeScenarioScore{66.16}{2.63} & \HThreeScenarioScore{68.28}{1.48} & \HThreeScenarioScore{63.02}{14.07} & \HThreeScenarioScore{45.29}{0.86} & \HThreeScenarioScore{14.83}{9.28} & \HThreeScenarioScore{49.39}{7.04} \\
 & FCT (s) & \HThreeScenarioScore{130.2}{30.3} & \HThreeScenarioScore{205.5}{15.0} & \HThreeScenarioScore{356.7}{74.9} & \HThreeScenarioScore{500.0}{173.2} & \HThreeScenarioScore{275.9}{14.4} & \HThreeScenarioScore{633.3}{321.5} & \HThreeScenarioScore{350.3}{152.9} \\
 & TG (pp) & \HThreeScenarioScore{8.19}{6.75} & \HThreeScenarioScore{15.99}{28.47} & \HThreeScenarioScore{38.09}{4.04} & \HThreeScenarioScore{25.25}{32.41} & \HThreeScenarioScore{5.97}{4.94} & \HThreeScenarioScore{7.82}{6.50} & \HThreeScenarioScore{16.88}{18.21} \\
\addlinespace[0.6pt]
DMR w/o comm. & SVR (\%) & \HThreeScenarioScore{32.16}{9.18} & \HThreeScenarioScore{65.36}{3.49} & \HThreeScenarioScore{46.37}{19.56} & \HThreeScenarioScore{56.11}{9.31} & \HThreeScenarioScore{15.38}{22.33} & \HThreeScenarioScore{14.22}{5.28} & \HThreeScenarioScore{38.27}{13.49} \\
 & FCT (s) & \HThreeScenarioScore{152.2}{28.2} & \HThreeScenarioScore{267.5}{115.5} & \HThreeScenarioScore{711.1}{416.9} & \HThreeScenarioScore{666.7}{208.2} & \HThreeScenarioScore{1666.7}{1097.0} & \HThreeScenarioScore{600.0}{173.2} & \HThreeScenarioScore{677.3}{494.1} \\
 & TG (pp) & \HThreeScenarioScore{1.57}{14.61} & \HThreeScenarioScore{15.19}{29.23} & \HThreeScenarioScore{16.18}{21.80} & \HThreeScenarioScore{18.34}{29.56} & \HThreeScenarioScore{-23.94}{17.99} & \HThreeScenarioScore{7.21}{6.82} & \HThreeScenarioScore{5.76}{21.55} \\
\midrule
\multicolumn{2}{@{}l}{\textbf{T5. DeepSeek, Symbolic}} & \makebox[51.4pt][c]{\textbf{S1R}} & \makebox[51.4pt][c]{\textbf{S2R}} & \makebox[51.4pt][c]{\textbf{S3R}} & \makebox[51.4pt][c]{\textbf{S4R}} & \makebox[51.4pt][c]{\textbf{S5R}} & \makebox[51.4pt][c]{\textbf{S6R}} & \makebox[51.4pt][c]{\textbf{Overall}} \\
CMR & SVR (\%) & \HThreeScenarioScore{8.97}{1.49} & \HThreeScenarioScore{20.66}{2.92} & \HThreeScenarioScore{42.53}{4.08} & \HThreeScenarioScore{39.69}{17.67} & \HThreeScenarioScore{20.86}{3.48} & \HThreeScenarioScore{13.03}{3.87} & \HThreeScenarioScore{24.29}{7.82} \\
 & FCT (s) & \HThreeScenarioScore{633.3}{57.7} & \HThreeScenarioScore{2466.7}{115.5} & \HThreeScenarioScore{3033.3}{493.3} & \HThreeScenarioScore{2800.0}{1053.6} & \HThreeScenarioScore{2366.7}{115.5} & \HThreeScenarioScore{1600.0}{346.4} & \HThreeScenarioScore{2150.0}{500.6} \\
 & HPR (\%) & \HThreeScenarioScore{4.72}{0.79} & \HThreeScenarioScore{29.53}{7.40} & \HThreeScenarioScore{55.33}{14.88} & \HThreeScenarioScore{10.07}{14.00} & \HThreeScenarioScore{47.82}{10.31} & \HThreeScenarioScore{0.00}{0.00} & \HThreeScenarioScore{24.58}{9.82} \\
 & CRR (\%) & \HThreeScenarioScore{16.67}{28.87} & \HThreeScenarioScore{33.33}{28.87} & \HThreeScenarioScore{33.33}{28.87} & \HThreeScenarioScore{16.67}{28.87} & \HThreeScenarioScore{100.00}{0.00} & \HThreeScenarioScore{16.67}{28.87} & \HThreeScenarioScore{36.11}{26.35} \\
 & EC (\%) & \HThreeScenarioScore{78.07}{6.55} & \HThreeScenarioScore{62.65}{3.74} & \HThreeScenarioScore{59.33}{9.15} & \HThreeScenarioScore{59.06}{17.08} & \HThreeScenarioScore{52.58}{3.78} & \HThreeScenarioScore{57.76}{2.38} & \HThreeScenarioScore{61.58}{8.68} \\
 & TG (pp) & \HThreeScenarioScore{0.94}{8.88} & \HThreeScenarioScore{-32.90}{3.30} & \HThreeScenarioScore{28.89}{12.10} & \HThreeScenarioScore{29.23}{27.75} & \HThreeScenarioScore{15.67}{6.09} & \HThreeScenarioScore{7.27}{2.67} & \HThreeScenarioScore{8.18}{13.23} \\
\addlinespace[0.6pt]
DMR & SVR (\%) & \HThreeScenarioScore{13.28}{4.41} & \HThreeScenarioScore{30.97}{19.44} & \HThreeScenarioScore{36.92}{10.07} & \HThreeScenarioScore{45.07}{7.80} & \HThreeScenarioScore{17.28}{2.61} & \HThreeScenarioScore{13.34}{3.82} & \HThreeScenarioScore{26.15}{9.84} \\
 & FCT (s) & \HThreeScenarioScore{478.7}{232.6} & \HThreeScenarioScore{2766.7}{680.7} & \HThreeScenarioScore{2866.7}{152.8} & \HThreeScenarioScore{2500.0}{866.0} & \HThreeScenarioScore{2466.7}{493.3} & \HThreeScenarioScore{1166.7}{57.7} & \HThreeScenarioScore{2040.9}{506.2} \\
 & HPR (\%) & \HThreeScenarioScore{9.98}{5.50} & \HThreeScenarioScore{26.28}{12.28} & \HThreeScenarioScore{47.59}{5.95} & \HThreeScenarioScore{40.05}{7.80} & \HThreeScenarioScore{14.38}{2.14} & \HThreeScenarioScore{46.19}{8.49} & \HThreeScenarioScore{30.74}{7.68} \\
 & CRR (\%) & \HThreeScenarioScore{16.67}{28.87} & \HThreeScenarioScore{16.67}{28.87} & \HThreeScenarioScore{50.00}{0.00} & \HThreeScenarioScore{66.67}{28.87} & \HThreeScenarioScore{50.00}{0.00} & \HThreeScenarioScore{100.00}{0.00} & \HThreeScenarioScore{50.00}{20.41} \\
 & EC (\%) & \HThreeScenarioScore{68.73}{1.69} & \HThreeScenarioScore{51.24}{20.10} & \HThreeScenarioScore{69.89}{0.67} & \HThreeScenarioScore{76.62}{3.49} & \HThreeScenarioScore{33.19}{1.60} & \HThreeScenarioScore{62.36}{4.68} & \HThreeScenarioScore{60.34}{8.60} \\
 & TG (pp) & \HThreeScenarioScore{5.25}{10.51} & \HThreeScenarioScore{-22.59}{19.25} & \HThreeScenarioScore{23.28}{22.73} & \HThreeScenarioScore{34.61}{7.14} & \HThreeScenarioScore{12.09}{3.02} & \HThreeScenarioScore{7.59}{4.49} & \HThreeScenarioScore{10.04}{13.40} \\
\addlinespace[0.6pt]
DMR w/o comm. & SVR (\%) & \HThreeScenarioScore{7.90}{3.14} & \HThreeScenarioScore{28.17}{10.85} & \HThreeScenarioScore{34.48}{2.32} & \HThreeScenarioScore{28.04}{7.48} & \HThreeScenarioScore{32.00}{7.94} & \HThreeScenarioScore{10.24}{3.51} & \HThreeScenarioScore{23.47}{6.64} \\
 & FCT (s) & \HThreeScenarioScore{633.3}{57.7} & \HThreeScenarioScore{2866.7}{585.9} & \HThreeScenarioScore{3133.3}{461.9} & \HThreeScenarioScore{3433.3}{611.0} & \HThreeScenarioScore{1600.0}{781.0} & \HThreeScenarioScore{1266.7}{321.5} & \HThreeScenarioScore{2155.6}{523.9} \\
 & HPR (\%) & \HThreeScenarioScore{12.21}{13.98} & \HThreeScenarioScore{26.59}{9.55} & \HThreeScenarioScore{57.47}{15.85} & \HThreeScenarioScore{20.94}{18.28} & \HThreeScenarioScore{69.52}{32.18} & \HThreeScenarioScore{29.06}{39.90} & \HThreeScenarioScore{35.96}{24.15} \\
 & CRR (\%) & \HThreeScenarioScore{16.67}{28.87} & \HThreeScenarioScore{16.67}{28.87} & \HThreeScenarioScore{66.67}{28.87} & \HThreeScenarioScore{33.33}{28.87} & \HThreeScenarioScore{83.33}{28.87} & \HThreeScenarioScore{66.67}{57.74} & \HThreeScenarioScore{47.22}{35.36} \\
 & EC (\%) & \HThreeScenarioScore{70.20}{2.87} & \HThreeScenarioScore{51.49}{18.02} & \HThreeScenarioScore{65.39}{15.46} & \HThreeScenarioScore{67.37}{13.36} & \HThreeScenarioScore{51.92}{4.23} & \HThreeScenarioScore{58.72}{1.56} & \HThreeScenarioScore{60.85}{11.33} \\
 & TG (pp) & \HThreeScenarioScore{-0.13}{5.17} & \HThreeScenarioScore{-25.39}{10.58} & \HThreeScenarioScore{20.84}{15.18} & \HThreeScenarioScore{17.58}{7.62} & \HThreeScenarioScore{26.82}{8.56} & \HThreeScenarioScore{4.48}{6.65} & \HThreeScenarioScore{7.37}{9.53} \\
\midrule
\multicolumn{2}{@{}l}{\textbf{T5. GLM, Symbolic}} & \makebox[51.4pt][c]{\textbf{S1R}} & \makebox[51.4pt][c]{\textbf{S2R}} & \makebox[51.4pt][c]{\textbf{S3R}} & \makebox[51.4pt][c]{\textbf{S4R}} & \makebox[51.4pt][c]{\textbf{S5R}} & \makebox[51.4pt][c]{\textbf{S6R}} & \makebox[51.4pt][c]{\textbf{Overall}} \\
CMR & SVR (\%) & \HThreeScenarioScore{14.14}{4.75} & \HThreeScenarioScore{54.11}{0.25} & \HThreeScenarioScore{47.17}{4.01} & \HThreeScenarioScore{45.54}{5.61} & \HThreeScenarioScore{33.03}{8.51} & \HThreeScenarioScore{17.00}{4.11} & \HThreeScenarioScore{35.17}{5.16} \\
 & FCT (s) & \HThreeScenarioScore{500.0}{100.0} & \HThreeScenarioScore{666.7}{57.7} & \HThreeScenarioScore{2500.0}{458.3} & \HThreeScenarioScore{2533.3}{550.8} & \HThreeScenarioScore{2133.3}{723.4} & \HThreeScenarioScore{1833.3}{305.5} & \HThreeScenarioScore{1694.4}{436.5} \\
 & HPR (\%) & \HThreeScenarioScore{5.66}{0.89} & \HThreeScenarioScore{36.90}{1.40} & \HThreeScenarioScore{66.11}{10.92} & \HThreeScenarioScore{26.02}{6.73} & \HThreeScenarioScore{83.94}{1.88} & \HThreeScenarioScore{26.76}{23.61} & \HThreeScenarioScore{40.90}{11.02} \\
 & CRR (\%) & \HThreeScenarioScore{16.67}{28.87} & \HThreeScenarioScore{50.00}{0.00} & \HThreeScenarioScore{50.00}{0.00} & \HThreeScenarioScore{33.33}{28.87} & \HThreeScenarioScore{100.00}{0.00} & \HThreeScenarioScore{50.00}{0.00} & \HThreeScenarioScore{50.00}{16.67} \\
 & EC (\%) & \HThreeScenarioScore{82.41}{5.24} & \HThreeScenarioScore{82.33}{4.94} & \HThreeScenarioScore{82.17}{0.47} & \HThreeScenarioScore{77.45}{5.36} & \HThreeScenarioScore{58.85}{6.40} & \HThreeScenarioScore{54.12}{23.54} & \HThreeScenarioScore{72.89}{10.61} \\
 & TG (pp) & \HThreeScenarioScore{11.72}{5.29} & \HThreeScenarioScore{0.52}{0.40} & \HThreeScenarioScore{42.40}{4.87} & \HThreeScenarioScore{23.97}{13.21} & \HThreeScenarioScore{29.30}{9.83} & \HThreeScenarioScore{14.48}{4.20} & \HThreeScenarioScore{20.40}{7.54} \\
\addlinespace[0.6pt]
DMR & SVR (\%) & \HThreeScenarioScore{10.72}{5.50} & \HThreeScenarioScore{53.95}{0.55} & \HThreeScenarioScore{33.59}{9.89} & \HThreeScenarioScore{40.63}{8.55} & \HThreeScenarioScore{30.18}{6.54} & \HThreeScenarioScore{13.43}{4.87} & \HThreeScenarioScore{30.42}{6.68} \\
 & FCT (s) & \HThreeScenarioScore{403.5}{167.2} & \HThreeScenarioScore{355.9}{124.9} & \HThreeScenarioScore{2566.7}{208.2} & \HThreeScenarioScore{2600.0}{529.2} & \HThreeScenarioScore{1833.3}{1078.6} & \HThreeScenarioScore{1100.0}{300.0} & \HThreeScenarioScore{1476.6}{519.6} \\
 & HPR (\%) & \HThreeScenarioScore{6.23}{0.19} & \HThreeScenarioScore{32.03}{10.73} & \HThreeScenarioScore{73.96}{6.53} & \HThreeScenarioScore{33.20}{5.77} & \HThreeScenarioScore{69.60}{26.66} & \HThreeScenarioScore{53.72}{21.58} & \HThreeScenarioScore{44.79}{15.10} \\
 & CRR (\%) & \HThreeScenarioScore{0.00}{0.00} & \HThreeScenarioScore{50.00}{0.00} & \HThreeScenarioScore{66.67}{28.87} & \HThreeScenarioScore{66.67}{28.87} & \HThreeScenarioScore{83.33}{28.87} & \HThreeScenarioScore{66.67}{28.87} & \HThreeScenarioScore{55.56}{23.57} \\
 & EC (\%) & \HThreeScenarioScore{81.75}{4.80} & \HThreeScenarioScore{75.66}{13.41} & \HThreeScenarioScore{61.50}{9.43} & \HThreeScenarioScore{67.42}{2.77} & \HThreeScenarioScore{59.73}{8.05} & \HThreeScenarioScore{51.60}{19.75} & \HThreeScenarioScore{66.28}{11.21} \\
 & TG (pp) & \HThreeScenarioScore{8.31}{4.95} & \HThreeScenarioScore{0.35}{1.09} & \HThreeScenarioScore{28.82}{9.12} & \HThreeScenarioScore{19.06}{19.28} & \HThreeScenarioScore{26.45}{5.18} & \HThreeScenarioScore{10.91}{4.47} & \HThreeScenarioScore{15.65}{9.38} \\
\addlinespace[0.6pt]
DMR w/o comm. & SVR (\%) & \HThreeScenarioScore{11.73}{0.34} & \HThreeScenarioScore{54.23}{0.33} & \HThreeScenarioScore{51.31}{1.33} & \HThreeScenarioScore{48.28}{8.00} & \HThreeScenarioScore{27.12}{5.06} & \HThreeScenarioScore{11.09}{3.65} & \HThreeScenarioScore{33.96}{4.18} \\
 & FCT (s) & \HThreeScenarioScore{600.0}{100.0} & \HThreeScenarioScore{528.7}{258.1} & \HThreeScenarioScore{2133.3}{251.7} & \HThreeScenarioScore{2300.0}{655.7} & \HThreeScenarioScore{2366.7}{208.2} & \HThreeScenarioScore{2000.0}{173.2} & \HThreeScenarioScore{1654.8}{327.4} \\
 & HPR (\%) & \HThreeScenarioScore{7.56}{0.09} & \HThreeScenarioScore{39.36}{0.04} & \HThreeScenarioScore{82.15}{5.99} & \HThreeScenarioScore{31.30}{0.69} & \HThreeScenarioScore{86.95}{0.28} & \HThreeScenarioScore{65.62}{8.36} & \HThreeScenarioScore{52.16}{4.21} \\
 & CRR (\%) & \HThreeScenarioScore{0.00}{0.00} & \HThreeScenarioScore{50.00}{0.00} & \HThreeScenarioScore{50.00}{0.00} & \HThreeScenarioScore{50.00}{0.00} & \HThreeScenarioScore{100.00}{0.00} & \HThreeScenarioScore{83.33}{28.87} & \HThreeScenarioScore{55.56}{11.79} \\
 & EC (\%) & \HThreeScenarioScore{76.67}{7.37} & \HThreeScenarioScore{59.28}{17.76} & \HThreeScenarioScore{65.47}{9.10} & \HThreeScenarioScore{71.53}{5.92} & \HThreeScenarioScore{54.42}{0.80} & \HThreeScenarioScore{51.45}{20.00} & \HThreeScenarioScore{63.14}{12.17} \\
 & TG (pp) & \HThreeScenarioScore{9.32}{1.07} & \HThreeScenarioScore{0.63}{0.55} & \HThreeScenarioScore{46.54}{2.26} & \HThreeScenarioScore{26.71}{8.23} & \HThreeScenarioScore{23.39}{4.07} & \HThreeScenarioScore{8.57}{3.29} & \HThreeScenarioScore{19.19}{4.12} \\
\bottomrule
\end{tabular*}\par
\newpage
{\fontsize{11}{12}\selectfont\bfseries H.3. Coordination Configurations, Results, and Execution Traces (continued)\par}
\vspace{5pt}
\fontsize{8.2}{8.5}\selectfont
\captionof*{table}{\textbf{Table~\ref{tab:appendix_h3_control_communication} (continued).} Per-scenario control and communication results.}
\noindent\begin{tabular*}{\dimexpr\linewidth-10pt\relax}{@{}ll@{\extracolsep{\fill}}*{7}{r}@{}}
\toprule
\textbf{Condition} & \textbf{Metric} & \multicolumn{6}{c}{\textbf{Scenario}} & {} \\
\midrule
\multicolumn{2}{@{}l}{\textbf{T5. DeepSeek, RGB}} & \makebox[51.4pt][c]{\textbf{S1R}} & \makebox[51.4pt][c]{\textbf{S2R}} & \makebox[51.4pt][c]{\textbf{S3R}} & \makebox[51.4pt][c]{\textbf{S4R}} & \makebox[51.4pt][c]{\textbf{S5R}} & \makebox[51.4pt][c]{\textbf{S6R}} & \makebox[51.4pt][c]{\textbf{Overall}} \\
CMR & SVR (\%) & \HThreeScenarioScore{4.95}{0.36} & \HThreeScenarioScore{9.28}{3.15} & \HThreeScenarioScore{17.97}{2.97} & \HThreeScenarioScore{24.07}{3.90} & \HThreeScenarioScore{11.87}{4.95} & \HThreeScenarioScore{3.99}{0.15} & \HThreeScenarioScore{12.02}{3.12} \\
 & FCT (s) & \HThreeScenarioScore{600.0}{0.0} & \HThreeScenarioScore{2000.0}{0.0} & \HThreeScenarioScore{2933.3}{251.7} & \HThreeScenarioScore{3700.0}{458.3} & \HThreeScenarioScore{2700.0}{100.0} & \HThreeScenarioScore{2100.0}{0.0} & \HThreeScenarioScore{2338.9}{217.3} \\
 & HPR (\%) & \HThreeScenarioScore{6.27}{0.57} & \HThreeScenarioScore{15.26}{3.09} & \HThreeScenarioScore{35.88}{2.21} & \HThreeScenarioScore{0.00}{0.00} & \HThreeScenarioScore{3.35}{0.24} & \HThreeScenarioScore{0.00}{0.00} & \HThreeScenarioScore{10.12}{1.57} \\
 & CRR (\%) & \HThreeScenarioScore{0.00}{0.00} & \HThreeScenarioScore{0.00}{0.00} & \HThreeScenarioScore{0.00}{0.00} & \HThreeScenarioScore{0.00}{0.00} & \HThreeScenarioScore{0.00}{0.00} & \HThreeScenarioScore{0.00}{0.00} & \HThreeScenarioScore{0.00}{0.00} \\
 & EC (\%) & \HThreeScenarioScore{91.17}{0.22} & \HThreeScenarioScore{90.76}{2.23} & \HThreeScenarioScore{85.09}{4.32} & \HThreeScenarioScore{82.10}{14.06} & \HThreeScenarioScore{84.00}{6.32} & \HThreeScenarioScore{76.99}{7.75} & \HThreeScenarioScore{85.02}{7.32} \\
 & TG (pp) & \HThreeScenarioScore{4.01}{0.43} & \HThreeScenarioScore{7.36}{3.28} & \HThreeScenarioScore{15.87}{3.93} & \HThreeScenarioScore{20.07}{6.63} & \HThreeScenarioScore{7.91}{4.87} & \HThreeScenarioScore{2.94}{0.78} & \HThreeScenarioScore{9.70}{3.97} \\
\addlinespace[0.6pt]
DMR & SVR (\%) & \HThreeScenarioScore{5.06}{0.66} & \HThreeScenarioScore{9.94}{2.67} & \HThreeScenarioScore{17.75}{1.82} & \HThreeScenarioScore{19.65}{1.21} & \HThreeScenarioScore{8.65}{2.59} & \HThreeScenarioScore{15.51}{20.03} & \HThreeScenarioScore{12.76}{8.37} \\
 & FCT (s) & \HThreeScenarioScore{600.0}{0.0} & \HThreeScenarioScore{2000.0}{0.0} & \HThreeScenarioScore{2833.3}{115.5} & \HThreeScenarioScore{3666.7}{115.5} & \HThreeScenarioScore{2666.7}{57.7} & \HThreeScenarioScore{2000.0}{173.2} & \HThreeScenarioScore{2294.4}{100.0} \\
 & HPR (\%) & \HThreeScenarioScore{5.56}{0.91} & \HThreeScenarioScore{14.36}{2.12} & \HThreeScenarioScore{32.49}{2.59} & \HThreeScenarioScore{0.00}{0.00} & \HThreeScenarioScore{2.69}{0.19} & \HThreeScenarioScore{0.00}{0.00} & \HThreeScenarioScore{9.18}{1.42} \\
 & CRR (\%) & \HThreeScenarioScore{0.00}{0.00} & \HThreeScenarioScore{0.00}{0.00} & \HThreeScenarioScore{0.00}{0.00} & \HThreeScenarioScore{0.00}{0.00} & \HThreeScenarioScore{0.00}{0.00} & \HThreeScenarioScore{0.00}{0.00} & \HThreeScenarioScore{0.00}{0.00} \\
 & EC (\%) & \HThreeScenarioScore{82.93}{3.89} & \HThreeScenarioScore{76.47}{4.98} & \HThreeScenarioScore{78.80}{9.82} & \HThreeScenarioScore{92.56}{2.19} & \HThreeScenarioScore{74.63}{3.22} & \HThreeScenarioScore{82.93}{3.57} & \HThreeScenarioScore{81.39}{5.23} \\
 & TG (pp) & \HThreeScenarioScore{4.12}{0.70} & \HThreeScenarioScore{8.02}{3.71} & \HThreeScenarioScore{15.65}{1.98} & \HThreeScenarioScore{15.65}{3.81} & \HThreeScenarioScore{4.69}{2.59} & \HThreeScenarioScore{14.46}{20.66} & \HThreeScenarioScore{10.43}{8.81} \\
\addlinespace[0.6pt]
DMR w/o comm. & SVR (\%) & \HThreeScenarioScore{5.12}{0.63} & \HThreeScenarioScore{10.11}{3.06} & \HThreeScenarioScore{15.62}{1.86} & \HThreeScenarioScore{20.84}{9.34} & \HThreeScenarioScore{12.29}{2.18} & \HThreeScenarioScore{3.00}{1.77} & \HThreeScenarioScore{11.16}{4.25} \\
 & FCT (s) & \HThreeScenarioScore{600.0}{0.0} & \HThreeScenarioScore{2000.0}{0.0} & \HThreeScenarioScore{3266.7}{230.9} & \HThreeScenarioScore{3600.0}{1100.0} & \HThreeScenarioScore{2800.0}{0.0} & \HThreeScenarioScore{2033.3}{57.7} & \HThreeScenarioScore{2383.3}{459.5} \\
 & HPR (\%) & \HThreeScenarioScore{5.37}{0.50} & \HThreeScenarioScore{15.06}{3.15} & \HThreeScenarioScore{29.39}{2.75} & \HThreeScenarioScore{0.00}{0.00} & \HThreeScenarioScore{3.09}{0.20} & \HThreeScenarioScore{0.00}{0.00} & \HThreeScenarioScore{8.82}{1.72} \\
 & CRR (\%) & \HThreeScenarioScore{0.00}{0.00} & \HThreeScenarioScore{0.00}{0.00} & \HThreeScenarioScore{0.00}{0.00} & \HThreeScenarioScore{0.00}{0.00} & \HThreeScenarioScore{0.00}{0.00} & \HThreeScenarioScore{0.00}{0.00} & \HThreeScenarioScore{0.00}{0.00} \\
 & EC (\%) & \HThreeScenarioScore{85.87}{8.41} & \HThreeScenarioScore{84.42}{12.88} & \HThreeScenarioScore{85.32}{5.77} & \HThreeScenarioScore{84.16}{5.50} & \HThreeScenarioScore{81.78}{5.26} & \HThreeScenarioScore{81.14}{4.28} & \HThreeScenarioScore{83.78}{7.60} \\
 & TG (pp) & \HThreeScenarioScore{4.18}{0.53} & \HThreeScenarioScore{8.20}{2.04} & \HThreeScenarioScore{13.51}{2.52} & \HThreeScenarioScore{16.85}{10.24} & \HThreeScenarioScore{8.34}{2.07} & \HThreeScenarioScore{1.95}{1.52} & \HThreeScenarioScore{8.84}{4.51} \\
\midrule
\multicolumn{2}{@{}l}{\textbf{T5. DeepSeek, RGB+DS}} & \makebox[51.4pt][c]{\textbf{S1R}} & \makebox[51.4pt][c]{\textbf{S2R}} & \makebox[51.4pt][c]{\textbf{S3R}} & \makebox[51.4pt][c]{\textbf{S4R}} & \makebox[51.4pt][c]{\textbf{S5R}} & \makebox[51.4pt][c]{\textbf{S6R}} & \makebox[51.4pt][c]{\textbf{Overall}} \\
CMR & SVR (\%) & \HThreeScenarioScore{9.75}{10.60} & \HThreeScenarioScore{6.05}{1.90} & \HThreeScenarioScore{14.50}{3.65} & \HThreeScenarioScore{11.89}{4.14} & \HThreeScenarioScore{8.81}{2.95} & \HThreeScenarioScore{6.10}{7.99} & \HThreeScenarioScore{9.52}{6.04} \\
 & FCT (s) & \HThreeScenarioScore{457.4}{247.0} & \HThreeScenarioScore{1966.7}{57.7} & \HThreeScenarioScore{2900.0}{264.6} & \HThreeScenarioScore{3933.3}{665.8} & \HThreeScenarioScore{2566.7}{208.2} & \HThreeScenarioScore{1633.3}{461.9} & \HThreeScenarioScore{2242.9}{372.9} \\
 & HPR (\%) & \HThreeScenarioScore{5.43}{0.33} & \HThreeScenarioScore{12.02}{1.87} & \HThreeScenarioScore{30.07}{3.55} & \HThreeScenarioScore{0.00}{0.00} & \HThreeScenarioScore{3.33}{0.15} & \HThreeScenarioScore{0.00}{0.00} & \HThreeScenarioScore{8.48}{1.65} \\
 & CRR (\%) & \HThreeScenarioScore{0.00}{0.00} & \HThreeScenarioScore{0.00}{0.00} & \HThreeScenarioScore{0.00}{0.00} & \HThreeScenarioScore{0.00}{0.00} & \HThreeScenarioScore{0.00}{0.00} & \HThreeScenarioScore{0.00}{0.00} & \HThreeScenarioScore{0.00}{0.00} \\
 & EC (\%) & \HThreeScenarioScore{80.50}{5.30} & \HThreeScenarioScore{72.45}{9.28} & \HThreeScenarioScore{68.61}{6.42} & \HThreeScenarioScore{65.76}{7.88} & \HThreeScenarioScore{54.50}{18.37} & \HThreeScenarioScore{53.97}{22.70} & \HThreeScenarioScore{65.97}{13.36} \\
 & TG (pp) & \HThreeScenarioScore{7.68}{8.41} & \HThreeScenarioScore{-12.17}{28.52} & \HThreeScenarioScore{11.39}{2.88} & \HThreeScenarioScore{-8.78}{17.22} & \HThreeScenarioScore{-8.43}{14.83} & \HThreeScenarioScore{4.32}{7.51} & \HThreeScenarioScore{-1.00}{15.63} \\
\addlinespace[0.6pt]
DMR & SVR (\%) & \HThreeScenarioScore{5.22}{1.36} & \HThreeScenarioScore{22.39}{27.26} & \HThreeScenarioScore{24.97}{18.26} & \HThreeScenarioScore{50.74}{9.13} & \HThreeScenarioScore{26.72}{9.07} & \HThreeScenarioScore{15.69}{1.19} & \HThreeScenarioScore{24.29}{14.41} \\
 & FCT (s) & \HThreeScenarioScore{533.3}{57.7} & \HThreeScenarioScore{1400.0}{866.0} & \HThreeScenarioScore{2666.7}{702.4} & \HThreeScenarioScore{1100.0}{360.6} & \HThreeScenarioScore{1866.7}{1101.5} & \HThreeScenarioScore{1033.3}{57.7} & \HThreeScenarioScore{1433.3}{657.4} \\
 & HPR (\%) & \HThreeScenarioScore{4.33}{0.38} & \HThreeScenarioScore{31.12}{13.80} & \HThreeScenarioScore{22.15}{7.14} & \HThreeScenarioScore{8.55}{8.09} & \HThreeScenarioScore{20.36}{21.51} & \HThreeScenarioScore{0.00}{0.00} & \HThreeScenarioScore{14.42}{11.33} \\
 & CRR (\%) & \HThreeScenarioScore{0.00}{0.00} & \HThreeScenarioScore{33.33}{28.87} & \HThreeScenarioScore{16.67}{28.87} & \HThreeScenarioScore{16.67}{28.87} & \HThreeScenarioScore{16.67}{28.87} & \HThreeScenarioScore{16.67}{28.87} & \HThreeScenarioScore{16.67}{26.35} \\
 & EC (\%) & \HThreeScenarioScore{71.52}{3.07} & \HThreeScenarioScore{65.94}{7.88} & \HThreeScenarioScore{63.22}{4.73} & \HThreeScenarioScore{69.45}{8.84} & \HThreeScenarioScore{55.24}{4.85} & \HThreeScenarioScore{53.90}{4.25} & \HThreeScenarioScore{63.21}{5.97} \\
 & TG (pp) & \HThreeScenarioScore{3.14}{3.36} & \HThreeScenarioScore{4.17}{4.39} & \HThreeScenarioScore{21.85}{18.60} & \HThreeScenarioScore{30.08}{9.31} & \HThreeScenarioScore{9.47}{18.42} & \HThreeScenarioScore{13.92}{1.51} & \HThreeScenarioScore{13.77}{11.58} \\
\addlinespace[0.6pt]
DMR w/o comm. & SVR (\%) & \HThreeScenarioScore{6.06}{1.80} & \HThreeScenarioScore{9.33}{4.13} & \HThreeScenarioScore{38.21}{6.23} & \HThreeScenarioScore{28.16}{15.49} & \HThreeScenarioScore{-2.82}{27.97} & \HThreeScenarioScore{12.75}{6.43} & \HThreeScenarioScore{15.28}{13.68} \\
 & FCT (s) & \HThreeScenarioScore{600.0}{0.0} & \HThreeScenarioScore{1933.3}{57.7} & \HThreeScenarioScore{1833.3}{351.2} & \HThreeScenarioScore{2500.0}{1228.8} & \HThreeScenarioScore{2766.7}{650.6} & \HThreeScenarioScore{1366.7}{550.8} & \HThreeScenarioScore{1833.3}{627.6} \\
 & HPR (\%) & \HThreeScenarioScore{4.98}{0.38} & \HThreeScenarioScore{25.67}{14.83} & \HThreeScenarioScore{37.39}{17.24} & \HThreeScenarioScore{9.27}{16.05} & \HThreeScenarioScore{31.49}{41.91} & \HThreeScenarioScore{0.00}{0.00} & \HThreeScenarioScore{18.13}{20.54} \\
 & CRR (\%) & \HThreeScenarioScore{0.00}{0.00} & \HThreeScenarioScore{33.33}{28.87} & \HThreeScenarioScore{33.33}{28.87} & \HThreeScenarioScore{16.67}{28.87} & \HThreeScenarioScore{50.00}{50.00} & \HThreeScenarioScore{16.67}{28.87} & \HThreeScenarioScore{25.00}{31.18} \\
 & EC (\%) & \HThreeScenarioScore{72.99}{5.21} & \HThreeScenarioScore{70.60}{5.03} & \HThreeScenarioScore{74.38}{9.98} & \HThreeScenarioScore{70.49}{8.66} & \HThreeScenarioScore{56.78}{2.80} & \HThreeScenarioScore{62.58}{4.50} & \HThreeScenarioScore{67.97}{6.52} \\
 & TG (pp) & \HThreeScenarioScore{3.99}{3.02} & \HThreeScenarioScore{-8.89}{34.08} & \HThreeScenarioScore{35.09}{6.41} & \HThreeScenarioScore{7.50}{26.46} & \HThreeScenarioScore{-20.07}{20.19} & \HThreeScenarioScore{10.97}{6.88} & \HThreeScenarioScore{4.77}{19.86} \\
\bottomrule
\end{tabular*}\par
\vspace{7pt}
\captionof{table}{Per-scenario team-size results for DeepSeek (Fig.~3(a) of the main paper).}
\label{tab:appendix_h3_team_size}
{\fontsize{7.8}{8.8}\selectfont For the four-robot conditions, see the CMR and DMR rows in the T4 DeepSeek symbolic block of Table~\ref{tab:appendix_h3_control_communication}.\par}\vspace{2.5pt}
\noindent\begin{tabular*}{\dimexpr\linewidth-10pt\relax}{@{}ll@{\extracolsep{\fill}}*{7}{r}@{}}
\toprule
\textbf{Condition} & \textbf{Metric} & \multicolumn{6}{c}{\textbf{Scenario}} & {} \\
\midrule
\multicolumn{2}{@{}l}{\textbf{T4. Team size}} & \makebox[51.4pt][c]{\textbf{S1F}} & \makebox[51.4pt][c]{\textbf{S2F}} & \makebox[51.4pt][c]{\textbf{S3F}} & \makebox[51.4pt][c]{\textbf{S4F}} & \makebox[51.4pt][c]{\textbf{S5F}} & \makebox[51.4pt][c]{\textbf{S6F}} & \makebox[51.4pt][c]{\textbf{Overall}} \\
SR, N=1 & SVR (\%) & \HThreeScenarioScore{30.59}{5.57} & \HThreeScenarioScore{50.17}{26.05} & \HThreeScenarioScore{30.19}{4.91} & \HThreeScenarioScore{37.77}{27.46} & \HThreeScenarioScore{39.32}{4.48} & \HThreeScenarioScore{7.02}{8.15} & \HThreeScenarioScore{32.51}{16.20} \\
 & FCT (s) & \HThreeScenarioScore{182.2}{0.0} & \HThreeScenarioScore{1100.0}{1126.9} & \HThreeScenarioScore{1100.0}{0.0} & \HThreeScenarioScore{1466.7}{1159.0} & \HThreeScenarioScore{363.1}{64.0} & \HThreeScenarioScore{1433.3}{378.6} & \HThreeScenarioScore{940.9}{678.3} \\
\addlinespace[0.6pt]
DMR, N=2 & SVR (\%) & \HThreeScenarioScore{36.22}{4.30} & \HThreeScenarioScore{67.70}{0.58} & \HThreeScenarioScore{46.31}{17.89} & \HThreeScenarioScore{43.03}{1.23} & \HThreeScenarioScore{45.21}{1.07} & \HThreeScenarioScore{13.12}{6.51} & \HThreeScenarioScore{41.93}{8.00} \\
 & FCT (s) & \HThreeScenarioScore{134.5}{19.1} & \HThreeScenarioScore{338.1}{107.3} & \HThreeScenarioScore{1066.7}{503.3} & \HThreeScenarioScore{866.7}{57.7} & \HThreeScenarioScore{322.8}{66.9} & \HThreeScenarioScore{900.0}{608.3} & \HThreeScenarioScore{604.8}{327.4} \\
 & TG (pp) & \HThreeScenarioScore{5.64}{8.46} & \HThreeScenarioScore{17.53}{26.57} & \HThreeScenarioScore{16.12}{19.40} & \HThreeScenarioScore{5.26}{27.03} & \HThreeScenarioScore{5.89}{3.43} & \HThreeScenarioScore{6.10}{6.92} & \HThreeScenarioScore{9.42}{18.00} \\
\addlinespace[0.6pt]
CMR, N=2 & SVR (\%) & \HThreeScenarioScore{20.00}{15.51} & \HThreeScenarioScore{27.68}{23.78} & \HThreeScenarioScore{17.20}{8.38} & \HThreeScenarioScore{27.93}{23.84} & \HThreeScenarioScore{14.25}{17.68} & \HThreeScenarioScore{10.40}{8.04} & \HThreeScenarioScore{19.58}{17.43} \\
 & FCT (s) & \HThreeScenarioScore{299.8}{260.0} & \HThreeScenarioScore{1266.7}{776.7} & \HThreeScenarioScore{2333.3}{1069.3} & \HThreeScenarioScore{1566.7}{1101.5} & \HThreeScenarioScore{1700.0}{1039.2} & \HThreeScenarioScore{1166.7}{461.9} & \HThreeScenarioScore{1388.9}{848.6} \\
 & TG (pp) & \HThreeScenarioScore{-10.59}{18.70} & \HThreeScenarioScore{-22.49}{49.51} & \HThreeScenarioScore{-13.00}{12.13} & \HThreeScenarioScore{-9.84}{18.84} & \HThreeScenarioScore{-25.06}{20.49} & \HThreeScenarioScore{3.39}{9.27} & \HThreeScenarioScore{-12.93}{25.20} \\
\bottomrule
\end{tabular*}\par
\newpage
{\fontsize{11}{12}\selectfont\bfseries H.3. Coordination Configurations, Results, and Execution Traces (continued)\par}
\vspace{5pt}
\fontsize{8.2}{8.5}\selectfont
\captionof*{table}{\textbf{Table~\ref{tab:appendix_h3_team_size} (continued).} Per-scenario team-size results for DeepSeek.}
{\fontsize{7.8}{8.8}\selectfont For the four-robot conditions, see the CMR and DMR rows in the T5 DeepSeek symbolic block of Table~\ref{tab:appendix_h3_control_communication}.\par}\vspace{2.5pt}
\noindent\begin{tabular*}{\dimexpr\linewidth-10pt\relax}{@{}ll@{\extracolsep{\fill}}*{7}{r}@{}}
\toprule
\textbf{Condition} & \textbf{Metric} & \multicolumn{6}{c}{\textbf{Scenario}} & {} \\
\midrule
\multicolumn{2}{@{}l}{\textbf{T5. Team size}} & \makebox[51.4pt][c]{\textbf{S1R}} & \makebox[51.4pt][c]{\textbf{S2R}} & \makebox[51.4pt][c]{\textbf{S3R}} & \makebox[51.4pt][c]{\textbf{S4R}} & \makebox[51.4pt][c]{\textbf{S5R}} & \makebox[51.4pt][c]{\textbf{S6R}} & \makebox[51.4pt][c]{\textbf{Overall}} \\
SR, N=1 & SVR (\%) & \HThreeScenarioScore{8.03}{8.21} & \HThreeScenarioScore{53.56}{0.47} & \HThreeScenarioScore{13.64}{12.86} & \HThreeScenarioScore{10.46}{14.45} & \HThreeScenarioScore{5.19}{2.71} & \HThreeScenarioScore{5.75}{3.98} & \HThreeScenarioScore{16.11}{8.80} \\
 & FCT (s) & \HThreeScenarioScore{533.4}{288.6} & \HThreeScenarioScore{400.0}{0.0} & \HThreeScenarioScore{2433.3}{896.3} & \HThreeScenarioScore{2733.3}{305.5} & \HThreeScenarioScore{2533.3}{115.5} & \HThreeScenarioScore{1566.7}{351.2} & \HThreeScenarioScore{1700.0}{431.4} \\
 & HPR (\%) & \HThreeScenarioScore{5.69}{4.88} & \HThreeScenarioScore{41.57}{0.28} & \HThreeScenarioScore{42.06}{35.81} & \HThreeScenarioScore{11.68}{20.23} & \HThreeScenarioScore{34.24}{35.57} & \HThreeScenarioScore{25.28}{43.78} & \HThreeScenarioScore{26.75}{28.57} \\
 & CRR (\%) & \HThreeScenarioScore{33.33}{57.74} & \HThreeScenarioScore{50.00}{0.00} & \HThreeScenarioScore{66.67}{28.87} & \HThreeScenarioScore{16.67}{28.87} & \HThreeScenarioScore{66.67}{28.87} & \HThreeScenarioScore{50.00}{0.00} & \HThreeScenarioScore{47.22}{31.18} \\
 & EC (\%) & \HThreeScenarioScore{33.63}{4.66} & \HThreeScenarioScore{37.67}{15.26} & \HThreeScenarioScore{39.63}{10.20} & \HThreeScenarioScore{32.55}{4.35} & \HThreeScenarioScore{35.62}{8.24} & \HThreeScenarioScore{43.06}{13.42} & \HThreeScenarioScore{37.02}{10.21} \\
\addlinespace[0.6pt]
DMR, N=2 & SVR (\%) & \HThreeScenarioScore{4.94}{0.07} & \HThreeScenarioScore{23.74}{25.40} & \HThreeScenarioScore{12.59}{4.40} & \HThreeScenarioScore{19.23}{5.01} & \HThreeScenarioScore{35.41}{2.81} & \HThreeScenarioScore{4.25}{2.30} & \HThreeScenarioScore{16.69}{10.82} \\
 & FCT (s) & \HThreeScenarioScore{566.7}{57.7} & \HThreeScenarioScore{1500.0}{866.0} & \HThreeScenarioScore{3200.0}{200.0} & \HThreeScenarioScore{3566.7}{305.5} & \HThreeScenarioScore{700.0}{0.0} & \HThreeScenarioScore{1900.0}{100.0} & \HThreeScenarioScore{1905.6}{386.6} \\
 & HPR (\%) & \HThreeScenarioScore{4.13}{0.53} & \HThreeScenarioScore{20.89}{13.94} & \HThreeScenarioScore{25.05}{6.06} & \HThreeScenarioScore{0.00}{0.00} & \HThreeScenarioScore{77.09}{10.34} & \HThreeScenarioScore{24.79}{42.94} & \HThreeScenarioScore{25.32}{19.07} \\
 & CRR (\%) & \HThreeScenarioScore{0.00}{0.00} & \HThreeScenarioScore{16.67}{28.87} & \HThreeScenarioScore{16.67}{28.87} & \HThreeScenarioScore{0.00}{0.00} & \HThreeScenarioScore{100.00}{0.00} & \HThreeScenarioScore{66.67}{28.87} & \HThreeScenarioScore{33.33}{20.41} \\
 & EC (\%) & \HThreeScenarioScore{53.79}{16.51} & \HThreeScenarioScore{51.49}{9.67} & \HThreeScenarioScore{50.79}{11.07} & \HThreeScenarioScore{51.59}{15.77} & \HThreeScenarioScore{47.79}{3.34} & \HThreeScenarioScore{59.17}{4.93} & \HThreeScenarioScore{52.43}{11.35} \\
 & TG (pp) & \HThreeScenarioScore{-3.09}{8.14} & \HThreeScenarioScore{-29.82}{24.98} & \HThreeScenarioScore{-1.05}{11.23} & \HThreeScenarioScore{8.77}{13.17} & \HThreeScenarioScore{30.22}{1.61} & \HThreeScenarioScore{-1.50}{4.42} & \HThreeScenarioScore{0.59}{12.99} \\
\addlinespace[0.6pt]
CMR, N=2 & SVR (\%) & \HThreeScenarioScore{4.35}{0.30} & \HThreeScenarioScore{7.62}{3.23} & \HThreeScenarioScore{10.44}{5.82} & \HThreeScenarioScore{24.36}{6.79} & \HThreeScenarioScore{8.92}{3.11} & \HThreeScenarioScore{7.63}{7.03} & \HThreeScenarioScore{10.55}{4.99} \\
 & FCT (s) & \HThreeScenarioScore{600.0}{0.0} & \HThreeScenarioScore{1966.7}{57.7} & \HThreeScenarioScore{2833.3}{115.5} & \HThreeScenarioScore{3533.3}{251.7} & \HThreeScenarioScore{2733.3}{57.7} & \HThreeScenarioScore{1733.3}{550.8} & \HThreeScenarioScore{2233.3}{253.9} \\
 & HPR (\%) & \HThreeScenarioScore{3.88}{0.43} & \HThreeScenarioScore{18.76}{12.50} & \HThreeScenarioScore{33.22}{21.44} & \HThreeScenarioScore{0.00}{0.00} & \HThreeScenarioScore{11.75}{4.26} & \HThreeScenarioScore{7.82}{13.55} & \HThreeScenarioScore{12.57}{11.67} \\
 & CRR (\%) & \HThreeScenarioScore{0.00}{0.00} & \HThreeScenarioScore{16.67}{28.87} & \HThreeScenarioScore{33.33}{28.87} & \HThreeScenarioScore{0.00}{0.00} & \HThreeScenarioScore{50.00}{0.00} & \HThreeScenarioScore{33.33}{28.87} & \HThreeScenarioScore{22.22}{20.41} \\
 & EC (\%) & \HThreeScenarioScore{62.91}{5.22} & \HThreeScenarioScore{55.26}{12.14} & \HThreeScenarioScore{57.38}{19.90} & \HThreeScenarioScore{34.44}{3.07} & \HThreeScenarioScore{27.73}{1.63} & \HThreeScenarioScore{49.44}{6.17} & \HThreeScenarioScore{47.86}{10.17} \\
 & TG (pp) & \HThreeScenarioScore{-3.69}{8.07} & \HThreeScenarioScore{-45.93}{3.37} & \HThreeScenarioScore{-3.20}{16.97} & \HThreeScenarioScore{13.90}{14.95} & \HThreeScenarioScore{3.73}{4.79} & \HThreeScenarioScore{1.88}{3.08} & \HThreeScenarioScore{-5.55}{10.17} \\
\bottomrule
\end{tabular*}\par
\vspace{7pt}
\captionof{table}{Per-scenario role-composition results for DeepSeek (Fig.~3(b) of the main paper).}
\label{tab:appendix_h3_role_composition}
{\fontsize{7.8}{8.8}\selectfont For the 4:0 composition, see the CMR and DMR rows in the T5 DeepSeek symbolic block of Table~\ref{tab:appendix_h3_control_communication}.\par}\vspace{2.5pt}
\noindent\begin{tabular*}{\dimexpr\linewidth-10pt\relax}{@{}ll@{\extracolsep{\fill}}*{7}{r}@{}}
\toprule
\textbf{Condition} & \textbf{Metric} & \multicolumn{6}{c}{\textbf{Scenario}} & {} \\
\midrule
\multicolumn{2}{@{}l}{\textbf{T5. FR:RR composition}} & \makebox[51.4pt][c]{\textbf{S1R}} & \makebox[51.4pt][c]{\textbf{S2R}} & \makebox[51.4pt][c]{\textbf{S3R}} & \makebox[51.4pt][c]{\textbf{S4R}} & \makebox[51.4pt][c]{\textbf{S5R}} & \makebox[51.4pt][c]{\textbf{S6R}} & \makebox[51.4pt][c]{\textbf{Overall}} \\
3:1, DMR & SVR (\%) & \HThreeScenarioScore{7.90}{2.35} & \HThreeScenarioScore{20.47}{8.76} & \HThreeScenarioScore{24.94}{8.76} & \HThreeScenarioScore{35.25}{15.36} & \HThreeScenarioScore{30.48}{11.08} & \HThreeScenarioScore{5.92}{0.37} & \HThreeScenarioScore{20.83}{9.29} \\
 & FCT (s) & \HThreeScenarioScore{566.7}{57.7} & \HThreeScenarioScore{2533.3}{550.8} & \HThreeScenarioScore{2933.3}{493.3} & \HThreeScenarioScore{2866.7}{1305.1} & \HThreeScenarioScore{1766.7}{513.2} & \HThreeScenarioScore{2066.7}{57.7} & \HThreeScenarioScore{2122.2}{648.1} \\
 & HPR (\%) & \HThreeScenarioScore{15.37}{16.07} & \HThreeScenarioScore{30.94}{8.90} & \HThreeScenarioScore{38.51}{8.83} & \HThreeScenarioScore{18.53}{9.12} & \HThreeScenarioScore{39.88}{38.78} & \HThreeScenarioScore{37.36}{33.27} & \HThreeScenarioScore{30.10}{22.76} \\
 & CRR (\%) & \HThreeScenarioScore{16.67}{28.87} & \HThreeScenarioScore{33.33}{28.87} & \HThreeScenarioScore{33.33}{28.87} & \HThreeScenarioScore{50.00}{0.00} & \HThreeScenarioScore{66.67}{28.87} & \HThreeScenarioScore{83.33}{28.87} & \HThreeScenarioScore{47.22}{26.35} \\
 & EC (\%) & \HThreeScenarioScore{67.55}{11.77} & \HThreeScenarioScore{53.41}{9.53} & \HThreeScenarioScore{70.79}{4.61} & \HThreeScenarioScore{69.43}{7.15} & \HThreeScenarioScore{55.90}{16.23} & \HThreeScenarioScore{68.97}{8.15} & \HThreeScenarioScore{64.34}{10.26} \\
 & TG (pp) & \HThreeScenarioScore{-0.13}{6.24} & \HThreeScenarioScore{-33.09}{8.29} & \HThreeScenarioScore{11.30}{16.21} & \HThreeScenarioScore{24.79}{9.43} & \HThreeScenarioScore{25.29}{13.62} & \HThreeScenarioScore{0.16}{3.61} & \HThreeScenarioScore{4.72}{10.47} \\
\addlinespace[0.6pt]
3:1, CMR & SVR (\%) & \HThreeScenarioScore{7.53}{1.33} & \HThreeScenarioScore{18.65}{2.43} & \HThreeScenarioScore{35.97}{9.38} & \HThreeScenarioScore{42.40}{10.91} & \HThreeScenarioScore{25.70}{1.22} & \HThreeScenarioScore{14.42}{5.11} & \HThreeScenarioScore{24.11}{6.36} \\
 & FCT (s) & \HThreeScenarioScore{566.7}{57.7} & \HThreeScenarioScore{2466.7}{305.5} & \HThreeScenarioScore{3200.0}{360.6} & \HThreeScenarioScore{3166.7}{351.2} & \HThreeScenarioScore{2533.3}{251.7} & \HThreeScenarioScore{1433.3}{152.8} & \HThreeScenarioScore{2227.8}{269.8} \\
 & HPR (\%) & \HThreeScenarioScore{7.52}{3.54} & \HThreeScenarioScore{36.30}{5.62} & \HThreeScenarioScore{47.73}{6.94} & \HThreeScenarioScore{28.04}{11.31} & \HThreeScenarioScore{44.07}{29.01} & \HThreeScenarioScore{19.65}{34.04} & \HThreeScenarioScore{30.55}{19.24} \\
 & CRR (\%) & \HThreeScenarioScore{0.00}{0.00} & \HThreeScenarioScore{16.67}{28.87} & \HThreeScenarioScore{0.00}{0.00} & \HThreeScenarioScore{33.33}{28.87} & \HThreeScenarioScore{50.00}{50.00} & \HThreeScenarioScore{16.67}{28.87} & \HThreeScenarioScore{19.44}{28.87} \\
 & EC (\%) & \HThreeScenarioScore{67.77}{7.93} & \HThreeScenarioScore{54.94}{3.01} & \HThreeScenarioScore{65.54}{13.95} & \HThreeScenarioScore{60.34}{4.86} & \HThreeScenarioScore{36.50}{12.48} & \HThreeScenarioScore{51.89}{17.63} & \HThreeScenarioScore{56.16}{11.23} \\
 & TG (pp) & \HThreeScenarioScore{-0.51}{7.11} & \HThreeScenarioScore{-34.91}{2.18} & \HThreeScenarioScore{22.33}{11.15} & \HThreeScenarioScore{31.94}{5.95} & \HThreeScenarioScore{20.51}{2.16} & \HThreeScenarioScore{8.67}{9.01} & \HThreeScenarioScore{8.01}{7.08} \\
\addlinespace[0.6pt]
2:2, DMR & SVR (\%) & \HThreeScenarioScore{8.40}{0.23} & \HThreeScenarioScore{24.56}{5.71} & \HThreeScenarioScore{20.14}{7.74} & \HThreeScenarioScore{32.80}{5.36} & \HThreeScenarioScore{27.62}{6.32} & \HThreeScenarioScore{9.91}{2.45} & \HThreeScenarioScore{20.57}{5.28} \\
 & FCT (s) & \HThreeScenarioScore{666.7}{57.7} & \HThreeScenarioScore{2800.0}{300.0} & \HThreeScenarioScore{3200.0}{435.9} & \HThreeScenarioScore{2733.3}{665.8} & \HThreeScenarioScore{2000.0}{624.5} & \HThreeScenarioScore{1933.3}{115.5} & \HThreeScenarioScore{2222.2}{434.0} \\
 & HPR (\%) & \HThreeScenarioScore{13.80}{11.40} & \HThreeScenarioScore{22.93}{1.55} & \HThreeScenarioScore{54.51}{8.06} & \HThreeScenarioScore{37.37}{3.61} & \HThreeScenarioScore{41.53}{18.92} & \HThreeScenarioScore{43.37}{40.62} & \HThreeScenarioScore{35.58}{19.23} \\
 & CRR (\%) & \HThreeScenarioScore{33.33}{57.74} & \HThreeScenarioScore{0.00}{0.00} & \HThreeScenarioScore{33.33}{28.87} & \HThreeScenarioScore{50.00}{0.00} & \HThreeScenarioScore{66.67}{28.87} & \HThreeScenarioScore{66.67}{28.87} & \HThreeScenarioScore{41.67}{31.18} \\
 & EC (\%) & \HThreeScenarioScore{76.09}{2.05} & \HThreeScenarioScore{51.57}{8.35} & \HThreeScenarioScore{66.29}{7.66} & \HThreeScenarioScore{67.89}{6.29} & \HThreeScenarioScore{53.24}{7.15} & \HThreeScenarioScore{65.26}{5.04} & \HThreeScenarioScore{63.39}{6.44} \\
 & TG (pp) & \HThreeScenarioScore{0.37}{8.37} & \HThreeScenarioScore{-28.99}{5.62} & \HThreeScenarioScore{6.50}{5.15} & \HThreeScenarioScore{22.34}{19.30} & \HThreeScenarioScore{22.43}{3.90} & \HThreeScenarioScore{4.16}{2.32} & \HThreeScenarioScore{4.47}{9.32} \\
\addlinespace[0.6pt]
2:2, CMR & SVR (\%) & \HThreeScenarioScore{7.64}{0.74} & \HThreeScenarioScore{21.79}{6.78} & \HThreeScenarioScore{46.54}{2.62} & \HThreeScenarioScore{31.08}{9.00} & \HThreeScenarioScore{23.07}{0.56} & \HThreeScenarioScore{13.16}{3.56} & \HThreeScenarioScore{23.88}{4.96} \\
 & FCT (s) & \HThreeScenarioScore{600.0}{0.0} & \HThreeScenarioScore{2466.7}{550.8} & \HThreeScenarioScore{2333.3}{550.8} & \HThreeScenarioScore{3200.0}{458.3} & \HThreeScenarioScore{2300.0}{0.0} & \HThreeScenarioScore{1800.0}{264.6} & \HThreeScenarioScore{2116.7}{384.4} \\
 & HPR (\%) & \HThreeScenarioScore{6.23}{2.15} & \HThreeScenarioScore{38.65}{3.00} & \HThreeScenarioScore{67.72}{14.10} & \HThreeScenarioScore{27.33}{7.76} & \HThreeScenarioScore{63.31}{17.16} & \HThreeScenarioScore{12.74}{22.07} & \HThreeScenarioScore{36.00}{13.25} \\
 & CRR (\%) & \HThreeScenarioScore{0.00}{0.00} & \HThreeScenarioScore{16.67}{28.87} & \HThreeScenarioScore{16.67}{28.87} & \HThreeScenarioScore{16.67}{28.87} & \HThreeScenarioScore{33.33}{28.87} & \HThreeScenarioScore{0.00}{0.00} & \HThreeScenarioScore{13.89}{23.57} \\
 & EC (\%) & \HThreeScenarioScore{75.94}{8.47} & \HThreeScenarioScore{53.98}{3.24} & \HThreeScenarioScore{48.46}{7.89} & \HThreeScenarioScore{53.42}{5.15} & \HThreeScenarioScore{48.60}{4.09} & \HThreeScenarioScore{52.56}{10.04} & \HThreeScenarioScore{55.49}{6.94} \\
 & TG (pp) & \HThreeScenarioScore{-0.39}{7.48} & \HThreeScenarioScore{-31.77}{7.20} & \HThreeScenarioScore{32.90}{15.19} & \HThreeScenarioScore{20.62}{10.01} & \HThreeScenarioScore{17.89}{3.00} & \HThreeScenarioScore{7.41}{3.38} & \HThreeScenarioScore{7.78}{8.75} \\
\addlinespace[0.6pt]
1:3, DMR & SVR (\%) & \HThreeScenarioScore{6.06}{1.59} & \HThreeScenarioScore{42.44}{19.35} & \HThreeScenarioScore{19.85}{5.37} & \HThreeScenarioScore{19.07}{13.74} & \HThreeScenarioScore{16.23}{1.77} & \HThreeScenarioScore{5.70}{2.67} & \HThreeScenarioScore{18.22}{10.04} \\
 & FCT (s) & \HThreeScenarioScore{600.0}{0.0} & \HThreeScenarioScore{966.7}{723.4} & \HThreeScenarioScore{3266.7}{115.5} & \HThreeScenarioScore{3366.7}{230.9} & \HThreeScenarioScore{2466.7}{288.7} & \HThreeScenarioScore{2033.3}{57.7} & \HThreeScenarioScore{2116.7}{335.8} \\
 & HPR (\%) & \HThreeScenarioScore{5.16}{1.31} & \HThreeScenarioScore{41.49}{2.03} & \HThreeScenarioScore{48.42}{16.60} & \HThreeScenarioScore{30.22}{10.86} & \HThreeScenarioScore{39.38}{43.88} & \HThreeScenarioScore{26.50}{45.90} & \HThreeScenarioScore{31.86}{27.18} \\
 & CRR (\%) & \HThreeScenarioScore{0.00}{0.00} & \HThreeScenarioScore{50.00}{0.00} & \HThreeScenarioScore{33.33}{28.87} & \HThreeScenarioScore{33.33}{28.87} & \HThreeScenarioScore{66.67}{28.87} & \HThreeScenarioScore{33.33}{28.87} & \HThreeScenarioScore{36.11}{23.57} \\
 & EC (\%) & \HThreeScenarioScore{70.93}{1.29} & \HThreeScenarioScore{64.18}{10.54} & \HThreeScenarioScore{51.84}{4.18} & \HThreeScenarioScore{68.83}{13.74} & \HThreeScenarioScore{55.09}{14.59} & \HThreeScenarioScore{61.62}{3.04} & \HThreeScenarioScore{62.08}{9.50} \\
 & TG (pp) & \HThreeScenarioScore{-1.98}{6.87} & \HThreeScenarioScore{-11.12}{19.33} & \HThreeScenarioScore{6.20}{11.03} & \HThreeScenarioScore{8.61}{9.91} & \HThreeScenarioScore{11.04}{4.47} & \HThreeScenarioScore{-0.06}{1.45} & \HThreeScenarioScore{2.12}{10.51} \\
\addlinespace[0.6pt]
1:3, CMR & SVR (\%) & \HThreeScenarioScore{7.92}{0.86} & \HThreeScenarioScore{33.95}{22.14} & \HThreeScenarioScore{38.82}{6.22} & \HThreeScenarioScore{32.40}{3.56} & \HThreeScenarioScore{23.35}{1.28} & \HThreeScenarioScore{13.34}{1.46} & \HThreeScenarioScore{24.97}{9.54} \\
 & FCT (s) & \HThreeScenarioScore{633.3}{57.7} & \HThreeScenarioScore{1966.7}{1250.3} & \HThreeScenarioScore{3300.0}{360.6} & \HThreeScenarioScore{3100.0}{200.0} & \HThreeScenarioScore{2600.0}{173.2} & \HThreeScenarioScore{1866.7}{230.9} & \HThreeScenarioScore{2244.4}{550.8} \\
 & HPR (\%) & \HThreeScenarioScore{5.41}{0.76} & \HThreeScenarioScore{41.47}{4.69} & \HThreeScenarioScore{79.18}{5.54} & \HThreeScenarioScore{3.42}{5.92} & \HThreeScenarioScore{29.75}{8.84} & \HThreeScenarioScore{10.97}{18.99} & \HThreeScenarioScore{28.37}{9.37} \\
 & CRR (\%) & \HThreeScenarioScore{0.00}{0.00} & \HThreeScenarioScore{16.67}{28.87} & \HThreeScenarioScore{0.00}{0.00} & \HThreeScenarioScore{0.00}{0.00} & \HThreeScenarioScore{16.67}{28.87} & \HThreeScenarioScore{0.00}{0.00} & \HThreeScenarioScore{5.56}{16.67} \\
 & EC (\%) & \HThreeScenarioScore{80.06}{7.81} & \HThreeScenarioScore{59.84}{11.58} & \HThreeScenarioScore{50.19}{4.85} & \HThreeScenarioScore{50.18}{12.04} & \HThreeScenarioScore{28.02}{4.87} & \HThreeScenarioScore{42.32}{12.18} & \HThreeScenarioScore{51.77}{9.45} \\
 & TG (pp) & \HThreeScenarioScore{-0.11}{8.58} & \HThreeScenarioScore{-19.61}{21.95} & \HThreeScenarioScore{25.18}{14.29} & \HThreeScenarioScore{21.94}{15.47} & \HThreeScenarioScore{18.17}{3.37} & \HThreeScenarioScore{7.59}{3.63} & \HThreeScenarioScore{8.86}{13.06} \\
\bottomrule
\end{tabular*}\par
\endgroup

\newpage
\noindent{\fontsize{11}{12}\selectfont\bfseries H.3. Coordination Configurations,
Results, and Execution Traces (continued)\par}
\vspace{6pt}

\noindent\textbf{Health protection and fire clearance.}
On T5 with symbolic observations, both four-robot DeepSeek teams increased
mean SVR relative to the matched single robot. Mean HPR was 26.75\% for
the single robot (Table~\ref{tab:appendix_h3_team_size}), 30.74\% for DMR,
and 24.58\% for CMR (Table~\ref{tab:appendix_h3_control_communication}).
Mean FCT rose from 1700.0\,s for the single robot to 2040.9\,s for four-robot
DMR. The single robot and the four-robot DMR team each recorded zero active
fires in one of 18 trials. Their remaining trials incurred the FCT penalty
for fires still burning at the deadline (Appendix~\ref{appendix:metrics}).
\par\vspace{4pt}

\noindent\textbf{Recorded execution trace.}
FR-03 was exploring Room~6 when it received FR-02's report of a fire in
Room~7. At its next decision (step~3), FR-03 cited the report and chose
to navigate to Room~7 to help suppress the fire. Action feedback confirmed
its arrival (Fig.~\ref{fig:appendix_h3_execution_trace}).

\par\vspace{4pt}
\noindent\begin{minipage}{\linewidth}
\captionsetup{type=figure}
\centering
\includegraphics[width=\linewidth]{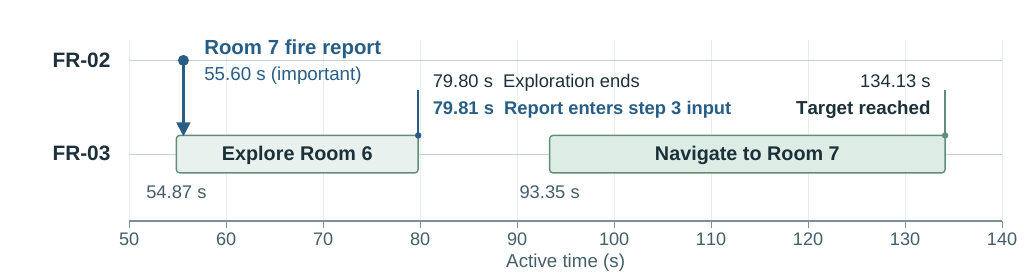}
\caption{\textbf{Message delivery and navigation in S5R trial~1.}
DeepSeek V4 Flash S2 controlled four FRs on T5 with RGB+DS and DMR
broadcasts enabled. Time is measured from the start of evaluation,
excluding confirmed simulator pauses
(Appendix~\ref{appendix:interaction_timing}).}
\label{fig:appendix_h3_execution_trace}
\end{minipage}
\par

\newpage
\subsection{Decision-Time Protocols and Detailed Results}
\label{appendix:decision_time}

We used DeepSeek V4 Flash S2 with symbolic observations in three
independent comparisons. On T1, we compared real-time execution with
pausing the simulator during model inference, requesting thinking off
in both conditions. We separately compared requested thinking on and
off under real-time execution on T1 and T5. Every condition used three
trials per scenario. The two comparisons on T1 used independently
collected real-time controls.

The evaluation used 300 seconds after preburn. For the inference-paused
condition, active time excluded simulator-confirmed pauses during model
calls (Appendix~\ref{appendix:interaction_timing}). The latency score is the mean wall-clock duration of completed calls
in each trial. The logs for the 108 runs contain 2,097 completed calls
and 108 cancellations. The provider-effective thinking setting was
unverified in the archived calls.

On T1, the inference-paused condition achieved a mean SVR 27.38
percentage points higher than the real-time condition, with higher SVR
in all six scenarios (Table~\ref{tab:appendix_h4_decision_time}).
Mean wall-clock call latency also differed (21.08\,s in real time and
16.25\,s with pauses), so the observed gain characterizes the two
timing protocols as tested. In the separate T5 comparison, the
condition requesting thinking on had higher HPR (+6.31 percentage
points) and lower SVR and CRR.
\par\vspace{5pt}

\begingroup
\setlength{\parindent}{0pt}
\setlength{\parskip}{0pt}
\setlength{\tabcolsep}{2pt}
\setlength{\aboverulesep}{1.1pt}
\setlength{\belowrulesep}{1.1pt}
\captionsetup{font=footnotesize,labelfont=bf,position=top,justification=raggedright,
  singlelinecheck=false,skip=2.2pt,hypcap=false}
\newcommand{\HDetailScore}[2]{%
  \makebox[22pt][r]{#1}\kern0.8pt\textpm\kern0.8pt%
  \makebox[22pt][l]{\textcolor{black!76}{#2}}%
}
\newcommand{\HDetailReference}[1]{\makebox[51.4pt][c]{#1}}
\fontsize{8.2}{8.5}\selectfont

\captionof{table}{Per-scenario decision-time results for DeepSeek V4 Flash S2.}
\label{tab:appendix_h4_decision_time}
\noindent\begin{tabular*}{\linewidth}{@{}ll@{\extracolsep{\fill}}*{7}{r}@{}}
\toprule
\textbf{Condition} & \textbf{Metric} & \multicolumn{6}{c}{\textbf{Scenario}} & {} \\
\midrule
\multicolumn{2}{@{}l}{\textbf{T1. Inference pause}} & \makebox[51.4pt][c]{\textbf{S1F}} & \makebox[51.4pt][c]{\textbf{S2F}} & \makebox[51.4pt][c]{\textbf{S3F}} & \makebox[51.4pt][c]{\textbf{S4F}} & \makebox[51.4pt][c]{\textbf{S5F}} & \makebox[51.4pt][c]{\textbf{S6F}} & \makebox[51.4pt][c]{\textbf{Overall}} \\
Real-time & SVR (\%) & \HDetailScore{2.42}{0.52} & \HDetailScore{26.14}{35.07} & \HDetailScore{0.88}{0.56} & \HDetailScore{49.04}{11.23} & \HDetailScore{27.90}{14.01} & \HDetailScore{1.12}{1.41} & \HDetailScore{17.91}{16.10} \\
 & FCT (s) & \HDetailScore{666.7}{57.7} & \HDetailScore{1200.0}{608.3} & \HDetailScore{2433.3}{57.7} & \HDetailScore{800.0}{264.6} & \HDetailScore{1033.3}{923.8} & \HDetailScore{1533.3}{115.5} & \HDetailScore{1277.8}{467.9} \\
 & Latency (s) & \HDetailScore{26.43}{11.05} & \HDetailScore{21.85}{5.58} & \HDetailScore{22.57}{7.15} & \HDetailScore{17.70}{3.95} & \HDetailScore{18.90}{4.34} & \HDetailScore{19.00}{1.17} & \HDetailScore{21.08}{6.33} \\
\addlinespace[1pt]
Paused & SVR (\%) & \HDetailScore{35.35}{17.14} & \HDetailScore{67.51}{3.52} & \HDetailScore{45.50}{14.43} & \HDetailScore{69.56}{34.72} & \HDetailScore{41.62}{4.97} & \HDetailScore{12.24}{7.09} & \HDetailScore{45.30}{17.29} \\
 & FCT (s) & \HDetailScore{284.9}{273.0} & \HDetailScore{347.4}{91.1} & \HDetailScore{800.0}{100.0} & \HDetailScore{698.8}{780.5} & \HDetailScore{400.0}{0.0} & \HDetailScore{1300.0}{700.0} & \HDetailScore{638.5}{445.7} \\
 & Latency (s) & \HDetailScore{20.01}{3.32} & \HDetailScore{17.28}{2.76} & \HDetailScore{15.63}{1.82} & \HDetailScore{15.29}{1.40} & \HDetailScore{18.58}{3.89} & \HDetailScore{10.68}{0.93} & \HDetailScore{16.25}{2.58} \\
\midrule
\multicolumn{2}{@{}l}{\textbf{T1. Requested thinking}} & \makebox[51.4pt][c]{\textbf{S1F}} & \makebox[51.4pt][c]{\textbf{S2F}} & \makebox[51.4pt][c]{\textbf{S3F}} & \makebox[51.4pt][c]{\textbf{S4F}} & \makebox[51.4pt][c]{\textbf{S5F}} & \makebox[51.4pt][c]{\textbf{S6F}} & \makebox[51.4pt][c]{\textbf{Overall}} \\
Off & SVR (\%) & \HDetailScore{18.28}{15.10} & \HDetailScore{19.96}{29.49} & \HDetailScore{21.24}{11.45} & \HDetailScore{59.93}{24.72} & \HDetailScore{4.56}{6.39} & \HDetailScore{3.59}{10.06} & \HDetailScore{21.26}{18.17} \\
 & FCT (s) & \HDetailScore{371.5}{285.6} & \HDetailScore{1266.7}{750.6} & \HDetailScore{1733.3}{1097.0} & \HDetailScore{633.3}{251.7} & \HDetailScore{2200.0}{173.2} & \HDetailScore{1400.0}{360.6} & \HDetailScore{1267.5}{587.6} \\
 & Latency (s) & \HDetailScore{18.11}{2.52} & \HDetailScore{18.96}{2.97} & \HDetailScore{16.80}{0.85} & \HDetailScore{13.29}{1.84} & \HDetailScore{15.90}{0.56} & \HDetailScore{15.56}{4.61} & \HDetailScore{16.44}{2.61} \\
\addlinespace[1pt]
On & SVR (\%) & \HDetailScore{9.55}{10.43} & \HDetailScore{18.94}{29.56} & \HDetailScore{18.58}{16.64} & \HDetailScore{59.66}{24.72} & \HDetailScore{14.56}{18.62} & \HDetailScore{1.04}{2.17} & \HDetailScore{20.39}{19.25} \\
 & FCT (s) & \HDetailScore{566.7}{152.8} & \HDetailScore{1233.3}{642.9} & \HDetailScore{1700.0}{721.1} & \HDetailScore{666.7}{305.5} & \HDetailScore{1700.0}{1039.2} & \HDetailScore{1633.3}{57.7} & \HDetailScore{1250.0}{596.3} \\
 & Latency (s) & \HDetailScore{26.44}{9.78} & \HDetailScore{19.14}{3.55} & \HDetailScore{18.78}{1.32} & \HDetailScore{14.66}{1.58} & \HDetailScore{13.84}{1.20} & \HDetailScore{14.35}{2.44} & \HDetailScore{17.87}{4.47} \\
\midrule
\multicolumn{2}{@{}l}{\textbf{T5. Requested thinking}} & \makebox[51.4pt][c]{\textbf{S1R}} & \makebox[51.4pt][c]{\textbf{S2R}} & \makebox[51.4pt][c]{\textbf{S3R}} & \makebox[51.4pt][c]{\textbf{S4R}} & \makebox[51.4pt][c]{\textbf{S5R}} & \makebox[51.4pt][c]{\textbf{S6R}} & \makebox[51.4pt][c]{\textbf{Overall}} \\
Off & SVR (\%) & \HDetailScore{8.19}{0.92} & \HDetailScore{34.63}{13.50} & \HDetailScore{43.74}{2.90} & \HDetailScore{35.87}{5.46} & \HDetailScore{25.24}{6.42} & \HDetailScore{15.11}{5.04} & \HDetailScore{27.13}{6.93} \\
 & FCT (s) & \HDetailScore{666.7}{57.7} & \HDetailScore{3033.3}{305.5} & \HDetailScore{3000.0}{519.6} & \HDetailScore{3033.3}{152.8} & \HDetailScore{2033.3}{230.9} & \HDetailScore{1433.3}{577.4} & \HDetailScore{2200.0}{359.8} \\
 & Latency (s) & \HDetailScore{23.96}{1.86} & \HDetailScore{19.15}{4.11} & \HDetailScore{14.56}{1.20} & \HDetailScore{13.35}{0.34} & \HDetailScore{22.56}{2.07} & \HDetailScore{20.36}{8.17} & \HDetailScore{18.99}{3.93} \\
 & HPR (\%) & \HDetailScore{5.93}{0.34} & \HDetailScore{25.34}{3.14} & \HDetailScore{61.95}{11.25} & \HDetailScore{19.61}{16.98} & \HDetailScore{44.64}{20.04} & \HDetailScore{6.23}{9.70} & \HDetailScore{27.28}{12.39} \\
 & CRR (\%) & \HDetailScore{0.00}{0.00} & \HDetailScore{0.00}{0.00} & \HDetailScore{66.67}{28.87} & \HDetailScore{33.33}{28.87} & \HDetailScore{83.33}{28.87} & \HDetailScore{50.00}{50.00} & \HDetailScore{38.89}{28.87} \\
\addlinespace[1pt]
On & SVR (\%) & \HDetailScore{9.42}{1.31} & \HDetailScore{26.86}{4.02} & \HDetailScore{38.34}{7.95} & \HDetailScore{38.16}{2.54} & \HDetailScore{25.21}{8.49} & \HDetailScore{11.52}{3.61} & \HDetailScore{24.92}{5.36} \\
 & FCT (s) & \HDetailScore{700.0}{0.0} & \HDetailScore{3033.3}{115.5} & \HDetailScore{3200.0}{458.3} & \HDetailScore{3033.3}{305.5} & \HDetailScore{2066.7}{776.7} & \HDetailScore{1566.7}{585.9} & \HDetailScore{2266.7}{458.9} \\
 & Latency (s) & \HDetailScore{23.52}{6.10} & \HDetailScore{20.49}{1.78} & \HDetailScore{16.50}{0.62} & \HDetailScore{14.59}{1.29} & \HDetailScore{18.89}{3.70} & \HDetailScore{21.60}{5.82} & \HDetailScore{19.26}{3.87} \\
 & HPR (\%) & \HDetailScore{6.26}{0.83} & \HDetailScore{23.21}{1.31} & \HDetailScore{55.40}{14.44} & \HDetailScore{15.69}{14.04} & \HDetailScore{80.74}{3.13} & \HDetailScore{20.25}{35.08} & \HDetailScore{33.59}{16.58} \\
 & CRR (\%) & \HDetailScore{0.00}{0.00} & \HDetailScore{0.00}{0.00} & \HDetailScore{33.33}{28.87} & \HDetailScore{33.33}{28.87} & \HDetailScore{100.00}{0.00} & \HDetailScore{33.33}{28.87} & \HDetailScore{33.33}{20.41} \\
\bottomrule
\end{tabular*}\par
\vspace{3pt}
{\footnotesize Summary statistics follow Appendix~\ref{appendix:complete_results}.
In 6 of 108 trials, the active-fire counts immediately before and after
the 300-s deadline differed, and the FCT penalty used the last valid count
at or before the deadline (Appendix~\ref{appendix:metrics}).\par}
\endgroup

\newpage
\subsection{Fire-Dynamics Settings and Extended Results}
\label{appendix:fire_sensitivity}

To complement the decision-time comparisons in Appendix~\ref{appendix:decision_time},
we examined how fire dynamics affected suppression and rescue over a 300-s
evaluation horizon. The default burning-rate multiplier was 20 during
preburn (\texttt{FireMultiply1}) and 1.5 during evaluation
(\texttt{FireMultiply2}). The default spread scale (\texttt{FireSpreadScale})
was 1.5 (Table~\ref{tab:appendix_scene_fields}). We varied one parameter
at a time. On T1, we tested preburn multipliers of 5, 10, 15, 20, and 25
(Table~\ref{tab:appendix_h5_fire_t1_initial}). Growth rate and spread scale
each took values of 1, 1.25, 1.5, 1.75, and 2
(Tables~\ref{tab:appendix_h5_fire_t1_speed} and~\ref{tab:appendix_h5_fire_t1_spread}).
On T5, the local symbolic trials used the same five settings for spread and
growth (Tables~\ref{tab:appendix_h5_fire_t5_spread} and~\ref{tab:appendix_h5_fire_t5_speed}).
The RGB and RGB+DS trials used growth rates of 1, 1.5, and 2
(Table~\ref{tab:appendix_h5_fire_v_speed}).
All sweeps used DeepSeek and a 5~s preburn. Symbolic trials used fourfold
suppression, and visual trials used eightfold suppression with the timing
protocol in Appendix~\ref{appendix:visual_trial_protocols}. For T1, we
collected separate robot trials at the default setting for each sweep.
The T5 symbolic spread sweep reused the main-experiment trials at its default.

In the tables, robot loss and free loss give the property lost between
the pre-ignition snapshot and the end of evaluation, in thousands of
durability units, with robot intervention and during free burning,
respectively. Each setting used three robot trials and one matched
free-burning reference per scenario (Appendix~\ref{appendix:evaluation_protocol}).
Overall weights the six scenarios equally.
Figure~\ref{fig:appendix_h5_free_burning} shows the six default reference trajectories.

On T1, mean property loss with robot intervention increased at each successive
spread-scale setting (Table~\ref{tab:appendix_h5_fire_t1_spread}). On T5 with
local symbolic observations, mean HPR declined at each successive growth-rate
setting (Table~\ref{tab:appendix_h5_fire_t5_speed}).
\par\vspace{5pt}

\noindent\begin{minipage}{\linewidth}
\captionsetup{type=figure}
\centering
\includegraphics[width=\linewidth]{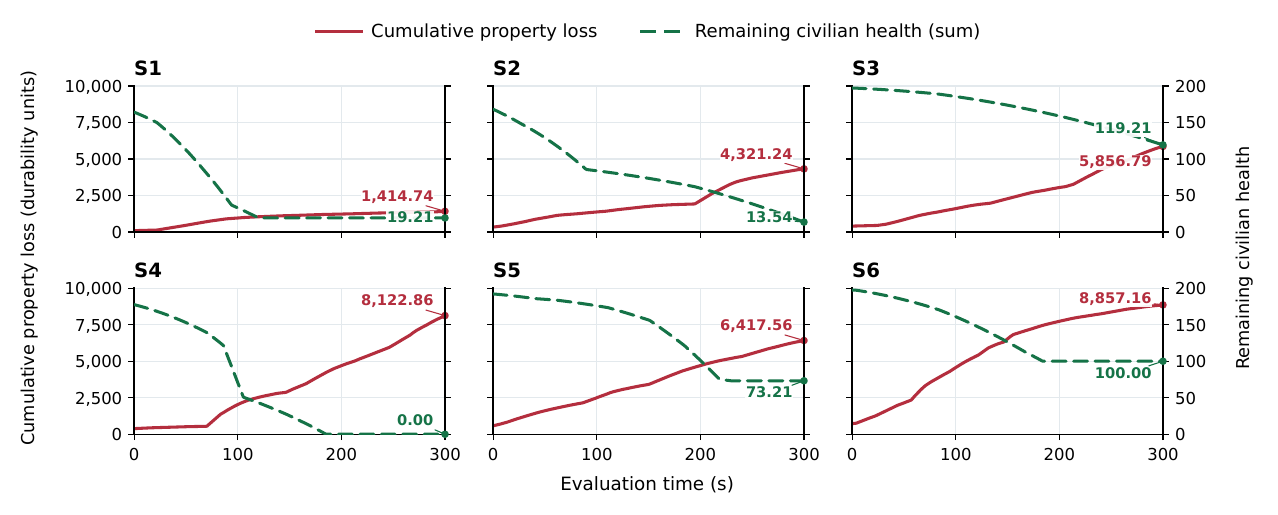}
\caption{\textbf{Free-burning trajectories under the default fire settings.}
Each panel shows the no-robot reference for one of S1--S6 over 300~s
of evaluation following a 5~s preburn. The solid red curve shows cumulative property loss
in durability units relative to the pre-ignition snapshot. The dashed green
curve shows the combined remaining health of the two civilians. All panels
use the same axis limits. Colored labels show the values at 300~s.}
\label{fig:appendix_h5_free_burning}
\end{minipage}
\par

\newpage
\noindent{\fontsize{11}{12}\selectfont\bfseries H.5. Fire-Dynamics Settings
and Extended Results (continued)\par}
\vspace{5pt}
\begingroup
\setlength{\parindent}{0pt}
\setlength{\parskip}{0pt}
\setlength{\tabcolsep}{2pt}
\setlength{\aboverulesep}{1.1pt}
\setlength{\belowrulesep}{1.1pt}
\captionsetup{font=footnotesize,labelfont=bf,position=top,justification=raggedright,
  singlelinecheck=false,skip=2.2pt,hypcap=false}
\newcommand{\HDetailScore}[2]{%
  \makebox[22pt][r]{#1}\kern0.8pt\textpm\kern0.8pt%
  \makebox[22pt][l]{\textcolor{black!76}{#2}}%
}
\newcommand{\HDetailReference}[1]{\makebox[51.4pt][c]{#1}}
\fontsize{8.2}{8.5}\selectfont

\captionof{table}{T1 sensitivity to the burning-rate multiplier during preburn under global symbolic observations.}
\label{tab:appendix_h5_fire_t1_initial}
\noindent\begin{tabular*}{\linewidth}{@{}ll@{\extracolsep{\fill}}*{7}{r}@{}}
\toprule
\textbf{Level} & \textbf{Metric} & \makebox[51.4pt][c]{\textbf{S1F}} & \makebox[51.4pt][c]{\textbf{S2F}} & \makebox[51.4pt][c]{\textbf{S3F}} & \makebox[51.4pt][c]{\textbf{S4F}} & \makebox[51.4pt][c]{\textbf{S5F}} & \makebox[51.4pt][c]{\textbf{S6F}} & \makebox[51.4pt][c]{\textbf{Overall}} \\
\midrule
5 & SVR (\%) & \HDetailScore{38.07}{38.11} & \HDetailScore{16.28}{15.27} & \HDetailScore{14.85}{5.25} & \HDetailScore{47.96}{31.30} & \HDetailScore{29.83}{10.39} & \HDetailScore{4.15}{4.52} & \HDetailScore{25.19}{21.68} \\
 & FCT (s) & \HDetailScore{353.5}{221.7} & \HDetailScore{2000.0}{264.6} & \HDetailScore{2366.7}{723.4} & \HDetailScore{1399.4}{955.0} & \HDetailScore{666.7}{57.7} & \HDetailScore{2066.7}{115.5} & \HDetailScore{1475.5}{511.7} \\
 & Robot loss & \HDetailScore{0.82}{0.50} & \HDetailScore{3.10}{0.57} & \HDetailScore{3.87}{0.24} & \HDetailScore{2.97}{1.78} & \HDetailScore{3.74}{0.55} & \HDetailScore{7.93}{0.37} & \HDetailScore{3.74}{0.84} \\
 & Free loss & \HDetailReference{1.32} & \HDetailReference{3.70} & \HDetailReference{4.55} & \HDetailReference{5.70} & \HDetailReference{5.33} & \HDetailReference{8.27} & \HDetailReference{4.81} \\
\addlinespace[1pt]
10 & SVR (\%) & \HDetailScore{44.63}{34.39} & \HDetailScore{9.75}{9.86} & \HDetailScore{28.33}{41.04} & \HDetailScore{54.12}{18.18} & \HDetailScore{30.04}{0.92} & \HDetailScore{8.11}{7.37} & \HDetailScore{29.16}{23.63} \\
 & FCT (s) & \HDetailScore{275.8}{194.4} & \HDetailScore{1800.0}{346.4} & \HDetailScore{1833.3}{1154.7} & \HDetailScore{1294.0}{882.0} & \HDetailScore{600.0}{0.0} & \HDetailScore{1466.7}{513.2} & \HDetailScore{1211.6}{649.7} \\
 & Robot loss & \HDetailScore{0.74}{0.46} & \HDetailScore{3.54}{0.39} & \HDetailScore{3.55}{2.03} & \HDetailScore{2.99}{1.19} & \HDetailScore{4.11}{0.05} & \HDetailScore{7.97}{0.64} & \HDetailScore{3.82}{1.02} \\
 & Free loss & \HDetailReference{1.34} & \HDetailReference{3.92} & \HDetailReference{4.95} & \HDetailReference{6.52} & \HDetailReference{5.87} & \HDetailReference{8.68} & \HDetailReference{5.21} \\
\addlinespace[1pt]
15 & SVR (\%) & \HDetailScore{16.62}{10.75} & \HDetailScore{37.28}{24.66} & \HDetailScore{9.84}{11.86} & \HDetailScore{35.34}{0.94} & \HDetailScore{8.46}{17.59} & \HDetailScore{2.10}{0.98} & \HDetailScore{18.27}{13.99} \\
 & FCT (s) & \HDetailScore{337.8}{227.2} & \HDetailScore{933.3}{838.6} & \HDetailScore{2200.0}{854.4} & \HDetailScore{1200.0}{100.0} & \HDetailScore{1966.7}{1184.6} & \HDetailScore{1666.7}{57.7} & \HDetailScore{1384.1}{695.4} \\
 & Robot loss & \HDetailScore{1.14}{0.15} & \HDetailScore{2.59}{1.02} & \HDetailScore{4.92}{0.65} & \HDetailScore{4.56}{0.07} & \HDetailScore{5.46}{1.05} & \HDetailScore{8.35}{0.08} & \HDetailScore{4.50}{0.66} \\
 & Free loss & \HDetailReference{1.36} & \HDetailReference{4.13} & \HDetailReference{5.46} & \HDetailReference{7.05} & \HDetailReference{5.97} & \HDetailReference{8.53} & \HDetailReference{5.42} \\
\addlinespace[1pt]
20 & SVR (\%) & \HDetailScore{13.57}{10.67} & \HDetailScore{53.87}{1.76} & \HDetailScore{18.32}{15.12} & \HDetailScore{48.11}{10.77} & \HDetailScore{27.65}{14.75} & \HDetailScore{2.98}{1.51} & \HDetailScore{27.42}{10.66} \\
 & FCT (s) & \HDetailScore{370.4}{286.4} & \HDetailScore{466.7}{57.7} & \HDetailScore{1600.0}{781.0} & \HDetailScore{866.7}{152.8} & \HDetailScore{1066.7}{981.5} & \HDetailScore{1600.0}{100.0} & \HDetailScore{995.1}{531.0} \\
 & Robot loss & \HDetailScore{1.22}{0.15} & \HDetailScore{1.99}{0.08} & \HDetailScore{4.78}{0.89} & \HDetailScore{4.21}{0.87} & \HDetailScore{4.64}{0.95} & \HDetailScore{8.59}{0.13} & \HDetailScore{4.24}{0.64} \\
 & Free loss & \HDetailReference{1.41} & \HDetailReference{4.32} & \HDetailReference{5.86} & \HDetailReference{8.12} & \HDetailReference{6.42} & \HDetailReference{8.86} & \HDetailReference{5.83} \\
\addlinespace[1pt]
25 & SVR (\%) & \HDetailScore{9.90}{13.13} & \HDetailScore{24.94}{25.49} & \HDetailScore{10.29}{16.69} & \HDetailScore{43.70}{0.06} & \HDetailScore{7.70}{3.95} & \HDetailScore{1.00}{0.85} & \HDetailScore{16.25}{13.64} \\
 & FCT (s) & \HDetailScore{523.7}{224.3} & \HDetailScore{533.3}{57.7} & \HDetailScore{1633.3}{635.1} & \HDetailScore{966.7}{57.7} & \HDetailScore{2033.3}{57.7} & \HDetailScore{1666.7}{57.7} & \HDetailScore{1226.2}{279.0} \\
 & Robot loss & \HDetailScore{1.33}{0.19} & \HDetailScore{3.35}{1.14} & \HDetailScore{5.51}{1.02} & \HDetailScore{4.77}{0.01} & \HDetailScore{6.12}{0.26} & \HDetailScore{8.79}{0.08} & \HDetailScore{4.98}{0.64} \\
 & Free loss & \HDetailReference{1.48} & \HDetailReference{4.47} & \HDetailReference{6.14} & \HDetailReference{8.47} & \HDetailReference{6.63} & \HDetailReference{8.87} & \HDetailReference{6.01} \\
\bottomrule
\end{tabular*}\par
\vspace{9pt}
\captionof{table}{T1 fire growth rate under global symbolic observations.}
\label{tab:appendix_h5_fire_t1_speed}
\noindent\begin{tabular*}{\linewidth}{@{}ll@{\extracolsep{\fill}}*{7}{r}@{}}
\toprule
\textbf{Level} & \textbf{Metric} & \makebox[51.4pt][c]{\textbf{S1F}} & \makebox[51.4pt][c]{\textbf{S2F}} & \makebox[51.4pt][c]{\textbf{S3F}} & \makebox[51.4pt][c]{\textbf{S4F}} & \makebox[51.4pt][c]{\textbf{S5F}} & \makebox[51.4pt][c]{\textbf{S6F}} & \makebox[51.4pt][c]{\textbf{Overall}} \\
\midrule
1 & SVR (\%) & \HDetailScore{14.21}{3.38} & \HDetailScore{8.62}{6.98} & \HDetailScore{4.29}{3.74} & \HDetailScore{26.08}{7.54} & \HDetailScore{19.11}{8.03} & \HDetailScore{-0.63}{0.73} & \HDetailScore{11.95}{5.71} \\
 & FCT (s) & \HDetailScore{458.1}{166.9} & \HDetailScore{1566.7}{1504.4} & \HDetailScore{1900.0}{173.2} & \HDetailScore{1433.3}{550.8} & \HDetailScore{2066.7}{757.2} & \HDetailScore{2233.3}{115.5} & \HDetailScore{1609.7}{731.6} \\
 & Robot loss & \HDetailScore{1.05}{0.04} & \HDetailScore{1.96}{0.15} & \HDetailScore{2.92}{0.11} & \HDetailScore{3.36}{0.34} & \HDetailScore{3.72}{0.37} & \HDetailScore{7.58}{0.06} & \HDetailScore{3.43}{0.22} \\
 & Free loss & \HDetailReference{1.23} & \HDetailReference{2.15} & \HDetailReference{3.05} & \HDetailReference{4.54} & \HDetailReference{4.60} & \HDetailReference{7.53} & \HDetailReference{3.85} \\
\addlinespace[1pt]
1.25 & SVR (\%) & \HDetailScore{13.19}{10.92} & \HDetailScore{19.78}{24.03} & \HDetailScore{7.41}{9.23} & \HDetailScore{32.50}{7.46} & \HDetailScore{7.79}{2.82} & \HDetailScore{1.90}{0.50} & \HDetailScore{13.76}{11.87} \\
 & FCT (s) & \HDetailScore{415.1}{177.9} & \HDetailScore{1566.7}{1011.6} & \HDetailScore{2633.3}{737.1} & \HDetailScore{1866.7}{680.7} & \HDetailScore{2700.0}{0.0} & \HDetailScore{1700.0}{360.6} & \HDetailScore{1813.6}{604.4} \\
 & Robot loss & \HDetailScore{1.14}{0.14} & \HDetailScore{2.97}{0.89} & \HDetailScore{4.21}{0.42} & \HDetailScore{4.05}{0.45} & \HDetailScore{4.92}{0.15} & \HDetailScore{8.12}{0.04} & \HDetailScore{4.23}{0.45} \\
 & Free loss & \HDetailReference{1.32} & \HDetailReference{3.70} & \HDetailReference{4.55} & \HDetailReference{6.00} & \HDetailReference{5.33} & \HDetailReference{8.27} & \HDetailReference{4.86} \\
\addlinespace[1pt]
1.5 & SVR (\%) & \HDetailScore{14.43}{11.75} & \HDetailScore{41.69}{20.63} & \HDetailScore{18.74}{15.66} & \HDetailScore{47.35}{9.47} & \HDetailScore{15.12}{18.09} & \HDetailScore{3.34}{1.49} & \HDetailScore{23.45}{14.31} \\
 & FCT (s) & \HDetailScore{368.1}{288.1} & \HDetailScore{700.0}{519.6} & \HDetailScore{1566.7}{808.3} & \HDetailScore{800.0}{264.6} & \HDetailScore{1666.7}{1011.6} & \HDetailScore{1266.7}{208.2} & \HDetailScore{1061.4}{597.6} \\
 & Robot loss & \HDetailScore{1.21}{0.17} & \HDetailScore{2.52}{0.89} & \HDetailScore{4.76}{0.92} & \HDetailScore{4.28}{0.77} & \HDetailScore{5.45}{1.16} & \HDetailScore{8.56}{0.13} & \HDetailScore{4.46}{0.78} \\
 & Free loss & \HDetailReference{1.41} & \HDetailReference{4.32} & \HDetailReference{5.86} & \HDetailReference{8.12} & \HDetailReference{6.42} & \HDetailReference{8.86} & \HDetailReference{5.83} \\
\addlinespace[1pt]
1.75 & SVR (\%) & \HDetailScore{10.88}{12.73} & \HDetailScore{37.04}{31.19} & \HDetailScore{32.85}{2.63} & \HDetailScore{42.28}{38.39} & \HDetailScore{5.92}{6.17} & \HDetailScore{1.15}{0.60} & \HDetailScore{21.69}{21.03} \\
 & FCT (s) & \HDetailScore{466.7}{115.5} & \HDetailScore{466.7}{115.5} & \HDetailScore{1033.3}{208.2} & \HDetailScore{966.4}{702.8} & \HDetailScore{1600.0}{100.0} & \HDetailScore{1400.0}{264.6} & \HDetailScore{988.8}{327.6} \\
 & Robot loss & \HDetailScore{1.36}{0.19} & \HDetailScore{2.83}{1.40} & \HDetailScore{4.29}{0.17} & \HDetailScore{5.27}{3.51} & \HDetailScore{6.29}{0.41} & \HDetailScore{8.92}{0.05} & \HDetailScore{4.83}{1.55} \\
 & Free loss & \HDetailReference{1.53} & \HDetailReference{4.49} & \HDetailReference{6.39} & \HDetailReference{9.13} & \HDetailReference{6.68} & \HDetailReference{9.02} & \HDetailReference{6.21} \\
\addlinespace[1pt]
2 & SVR (\%) & \HDetailScore{12.23}{13.30} & \HDetailScore{55.35}{0.78} & \HDetailScore{4.17}{1.34} & \HDetailScore{47.74}{0.82} & \HDetailScore{16.18}{22.58} & \HDetailScore{1.48}{0.36} & \HDetailScore{22.86}{10.72} \\
 & FCT (s) & \HDetailScore{346.1}{186.8} & \HDetailScore{237.5}{48.2} & \HDetailScore{1333.3}{152.8} & \HDetailScore{766.7}{57.7} & \HDetailScore{833.3}{378.6} & \HDetailScore{1066.7}{57.7} & \HDetailScore{763.9}{187.3} \\
 & Robot loss & \HDetailScore{1.39}{0.21} & \HDetailScore{2.01}{0.04} & \HDetailScore{6.40}{0.09} & \HDetailScore{4.84}{0.08} & \HDetailScore{5.93}{1.60} & \HDetailScore{9.06}{0.03} & \HDetailScore{4.94}{0.66} \\
 & Free loss & \HDetailReference{1.58} & \HDetailReference{4.51} & \HDetailReference{6.68} & \HDetailReference{9.26} & \HDetailReference{7.07} & \HDetailReference{9.19} & \HDetailReference{6.38} \\
\bottomrule
\end{tabular*}\par
\vspace{9pt}
\captionof{table}{T1 fire spread scale under global symbolic observations.}
\label{tab:appendix_h5_fire_t1_spread}
\noindent\begin{tabular*}{\linewidth}{@{}ll@{\extracolsep{\fill}}*{7}{r}@{}}
\toprule
\textbf{Level} & \textbf{Metric} & \makebox[51.4pt][c]{\textbf{S1F}} & \makebox[51.4pt][c]{\textbf{S2F}} & \makebox[51.4pt][c]{\textbf{S3F}} & \makebox[51.4pt][c]{\textbf{S4F}} & \makebox[51.4pt][c]{\textbf{S5F}} & \makebox[51.4pt][c]{\textbf{S6F}} & \makebox[51.4pt][c]{\textbf{Overall}} \\
\midrule
1 & SVR (\%) & \HDetailScore{11.39}{2.32} & \HDetailScore{66.17}{0.17} & \HDetailScore{9.51}{10.94} & \HDetailScore{53.07}{18.99} & \HDetailScore{45.19}{1.53} & \HDetailScore{16.16}{10.03} & \HDetailScore{33.58}{9.90} \\
 & FCT (s) & \HDetailScore{126.8}{11.6} & \HDetailScore{187.2}{0.0} & \HDetailScore{340.8}{137.9} & \HDetailScore{600.0}{346.4} & \HDetailScore{252.8}{127.5} & \HDetailScore{500.0}{0.0} & \HDetailScore{334.6}{160.9} \\
 & Robot loss & \HDetailScore{0.44}{0.01} & \HDetailScore{0.92}{0.00} & \HDetailScore{1.23}{0.15} & \HDetailScore{0.74}{0.30} & \HDetailScore{0.56}{0.02} & \HDetailScore{1.73}{0.21} & \HDetailScore{0.94}{0.16} \\
 & Free loss & \HDetailReference{0.50} & \HDetailReference{2.72} & \HDetailReference{1.36} & \HDetailReference{1.57} & \HDetailReference{1.03} & \HDetailReference{2.07} & \HDetailReference{1.54} \\
\addlinespace[1pt]
1.25 & SVR (\%) & \HDetailScore{4.29}{1.53} & \HDetailScore{66.58}{0.43} & \HDetailScore{16.05}{17.87} & \HDetailScore{30.36}{0.99} & \HDetailScore{15.10}{2.77} & \HDetailScore{0.94}{0.54} & \HDetailScore{22.22}{7.43} \\
 & FCT (s) & \HDetailScore{146.8}{8.1} & \HDetailScore{329.1}{122.9} & \HDetailScore{1266.7}{513.2} & \HDetailScore{666.7}{57.7} & \HDetailScore{533.3}{57.7} & \HDetailScore{633.3}{115.5} & \HDetailScore{596.0}{223.0} \\
 & Robot loss & \HDetailScore{0.77}{0.01} & \HDetailScore{1.05}{0.01} & \HDetailScore{3.11}{0.66} & \HDetailScore{2.98}{0.04} & \HDetailScore{3.71}{0.12} & \HDetailScore{2.96}{0.02} & \HDetailScore{2.43}{0.28} \\
 & Free loss & \HDetailReference{0.80} & \HDetailReference{3.14} & \HDetailReference{3.71} & \HDetailReference{4.28} & \HDetailReference{4.37} & \HDetailReference{2.98} & \HDetailReference{3.22} \\
\addlinespace[1pt]
1.5 & SVR (\%) & \HDetailScore{8.19}{9.16} & \HDetailScore{38.56}{26.88} & \HDetailScore{25.04}{1.98} & \HDetailScore{42.25}{0.88} & \HDetailScore{19.34}{21.55} & \HDetailScore{6.61}{8.49} & \HDetailScore{23.33}{14.99} \\
 & FCT (s) & \HDetailScore{532.1}{210.3} & \HDetailScore{866.7}{723.4} & \HDetailScore{1533.3}{665.8} & \HDetailScore{900.0}{100.0} & \HDetailScore{1600.0}{1039.2} & \HDetailScore{1400.0}{435.9} & \HDetailScore{1138.7}{617.9} \\
 & Robot loss & \HDetailScore{1.30}{0.13} & \HDetailScore{2.65}{1.16} & \HDetailScore{4.39}{0.12} & \HDetailScore{4.69}{0.07} & \HDetailScore{5.18}{1.38} & \HDetailScore{8.27}{0.75} & \HDetailScore{4.41}{0.80} \\
 & Free loss & \HDetailReference{1.41} & \HDetailReference{4.32} & \HDetailReference{5.86} & \HDetailReference{8.12} & \HDetailReference{6.42} & \HDetailReference{8.86} & \HDetailReference{5.83} \\
\addlinespace[1pt]
1.75 & SVR (\%) & \HDetailScore{10.01}{12.13} & \HDetailScore{25.63}{28.93} & \HDetailScore{0.74}{1.65} & \HDetailScore{40.66}{4.19} & \HDetailScore{0.01}{3.94} & \HDetailScore{7.13}{5.87} & \HDetailScore{14.03}{13.26} \\
 & FCT (s) & \HDetailScore{502.1}{261.1} & \HDetailScore{1633.3}{986.6} & \HDetailScore{2566.7}{288.7} & \HDetailScore{1566.7}{723.4} & \HDetailScore{3333.3}{57.7} & \HDetailScore{1400.0}{264.6} & \HDetailScore{1833.7}{535.6} \\
 & Robot loss & \HDetailScore{1.50}{0.20} & \HDetailScore{3.75}{1.46} & \HDetailScore{6.47}{0.11} & \HDetailScore{5.22}{0.37} & \HDetailScore{12.00}{0.47} & \HDetailScore{8.82}{0.56} & \HDetailScore{6.29}{0.69} \\
 & Free loss & \HDetailReference{1.66} & \HDetailReference{5.05} & \HDetailReference{6.51} & \HDetailReference{8.80} & \HDetailReference{12.00} & \HDetailReference{9.50} & \HDetailReference{7.26} \\
\addlinespace[1pt]
2 & SVR (\%) & \HDetailScore{16.67}{14.85} & \HDetailScore{13.78}{23.10} & \HDetailScore{19.06}{16.89} & \HDetailScore{52.04}{1.57} & \HDetailScore{1.41}{2.07} & \HDetailScore{6.27}{4.57} & \HDetailScore{18.20}{13.33} \\
 & FCT (s) & \HDetailScore{427.7}{259.5} & \HDetailScore{1466.7}{577.4} & \HDetailScore{2500.0}{866.0} & \HDetailScore{2266.7}{577.4} & \HDetailScore{3966.7}{404.1} & \HDetailScore{2033.3}{208.2} & \HDetailScore{2110.2}{530.8} \\
 & Robot loss & \HDetailScore{1.39}{0.25} & \HDetailScore{5.48}{1.47} & \HDetailScore{7.67}{1.60} & \HDetailScore{5.59}{0.18} & \HDetailScore{15.97}{0.34} & \HDetailScore{14.25}{0.70} & \HDetailScore{8.39}{0.95} \\
 & Free loss & \HDetailReference{1.66} & \HDetailReference{6.36} & \HDetailReference{9.48} & \HDetailReference{11.66} & \HDetailReference{16.20} & \HDetailReference{15.21} & \HDetailReference{10.09} \\
\bottomrule
\end{tabular*}\par
\newpage
{\fontsize{11}{12}\selectfont\bfseries H.5. Fire-Dynamics Settings and Extended Results (continued)\par}
\vspace{5pt}
\captionof{table}{T5 fire spread scale under local symbolic observations.}
\label{tab:appendix_h5_fire_t5_spread}
\noindent\begin{tabular*}{\linewidth}{@{}ll@{\extracolsep{\fill}}*{7}{r}@{}}
\toprule
\textbf{Level} & \textbf{Metric} & \makebox[51.4pt][c]{\textbf{S1R}} & \makebox[51.4pt][c]{\textbf{S2R}} & \makebox[51.4pt][c]{\textbf{S3R}} & \makebox[51.4pt][c]{\textbf{S4R}} & \makebox[51.4pt][c]{\textbf{S5R}} & \makebox[51.4pt][c]{\textbf{S6R}} & \makebox[51.4pt][c]{\textbf{Overall}} \\
\midrule
1 & SVR (\%) & \HDetailScore{0.00}{0.00} & \HDetailScore{19.22}{3.55} & \HDetailScore{8.36}{7.30} & \HDetailScore{53.57}{14.43} & \HDetailScore{2.42}{1.37} & \HDetailScore{16.46}{11.33} & \HDetailScore{16.67}{8.21} \\
 & FCT (s) & \HDetailScore{184.4}{6.8} & \HDetailScore{1800.0}{0.0} & \HDetailScore{277.9}{108.3} & \HDetailScore{364.0}{62.3} & \HDetailScore{500.0}{0.0} & \HDetailScore{433.3}{57.7} & \HDetailScore{593.3}{56.3} \\
 & Robot loss & \HDetailScore{0.50}{0.00} & \HDetailScore{2.20}{0.10} & \HDetailScore{1.24}{0.10} & \HDetailScore{0.73}{0.23} & \HDetailScore{1.00}{0.01} & \HDetailScore{1.73}{0.23} & \HDetailScore{1.23}{0.14} \\
 & Free loss & \HDetailReference{0.50} & \HDetailReference{2.72} & \HDetailReference{1.36} & \HDetailReference{1.57} & \HDetailReference{1.03} & \HDetailReference{2.07} & \HDetailReference{1.54} \\
\addlinespace[1pt]
1.25 & SVR (\%) & \HDetailScore{0.00}{0.00} & \HDetailScore{34.89}{27.97} & \HDetailScore{24.81}{10.55} & \HDetailScore{8.86}{20.47} & \HDetailScore{9.57}{9.20} & \HDetailScore{15.77}{12.88} & \HDetailScore{15.65}{16.14} \\
 & FCT (s) & \HDetailScore{184.4}{4.2} & \HDetailScore{1337.2}{974.7} & \HDetailScore{1366.7}{404.1} & \HDetailScore{1500.0}{755.0} & \HDetailScore{666.7}{152.8} & \HDetailScore{533.3}{152.8} & \HDetailScore{931.4}{537.0} \\
 & Robot loss & \HDetailScore{0.80}{0.00} & \HDetailScore{2.05}{0.88} & \HDetailScore{2.79}{0.39} & \HDetailScore{3.90}{0.88} & \HDetailScore{3.96}{0.40} & \HDetailScore{2.51}{0.38} & \HDetailScore{2.67}{0.58} \\
 & Free loss & \HDetailReference{0.80} & \HDetailReference{3.14} & \HDetailReference{3.71} & \HDetailReference{4.28} & \HDetailReference{4.37} & \HDetailReference{2.98} & \HDetailReference{3.22} \\
\addlinespace[1pt]
1.5 & SVR (\%) & \HDetailScore{13.28}{4.41} & \HDetailScore{30.97}{19.44} & \HDetailScore{36.92}{10.07} & \HDetailScore{45.07}{7.80} & \HDetailScore{17.28}{2.61} & \HDetailScore{13.34}{3.82} & \HDetailScore{26.15}{9.84} \\
 & FCT (s) & \HDetailScore{478.7}{232.6} & \HDetailScore{2766.7}{680.7} & \HDetailScore{2866.7}{152.8} & \HDetailScore{2500.0}{866.0} & \HDetailScore{2466.7}{493.3} & \HDetailScore{1166.7}{57.7} & \HDetailScore{2040.9}{506.2} \\
 & Robot loss & \HDetailScore{1.23}{0.06} & \HDetailScore{2.98}{0.84} & \HDetailScore{3.69}{0.59} & \HDetailScore{4.46}{0.63} & \HDetailScore{5.31}{0.17} & \HDetailScore{7.68}{0.34} & \HDetailScore{4.22}{0.52} \\
 & Free loss & \HDetailReference{1.41} & \HDetailReference{4.32} & \HDetailReference{5.86} & \HDetailReference{8.12} & \HDetailReference{6.42} & \HDetailReference{8.86} & \HDetailReference{5.83} \\
\addlinespace[1pt]
1.75 & SVR (\%) & \HDetailScore{5.52}{2.04} & \HDetailScore{27.30}{26.56} & \HDetailScore{20.66}{4.76} & \HDetailScore{36.28}{15.02} & \HDetailScore{20.06}{5.92} & \HDetailScore{7.02}{3.94} & \HDetailScore{19.47}{12.96} \\
 & FCT (s) & \HDetailScore{633.3}{115.5} & \HDetailScore{1600.0}{953.9} & \HDetailScore{2033.3}{585.9} & \HDetailScore{2433.3}{1530.8} & \HDetailScore{4033.3}{907.4} & \HDetailScore{1866.7}{378.6} & \HDetailScore{2100.0}{873.4} \\
 & Robot loss & \HDetailScore{1.57}{0.03} & \HDetailScore{3.67}{1.34} & \HDetailScore{5.17}{0.31} & \HDetailScore{5.61}{1.32} & \HDetailScore{9.59}{0.71} & \HDetailScore{8.83}{0.37} & \HDetailScore{5.74}{0.85} \\
 & Free loss & \HDetailReference{1.66} & \HDetailReference{5.05} & \HDetailReference{6.51} & \HDetailReference{8.80} & \HDetailReference{12.00} & \HDetailReference{9.50} & \HDetailReference{7.26} \\
\addlinespace[1pt]
2 & SVR (\%) & \HDetailScore{4.02}{0.16} & \HDetailScore{19.74}{20.84} & \HDetailScore{26.23}{11.06} & \HDetailScore{37.37}{8.76} & \HDetailScore{11.28}{3.47} & \HDetailScore{15.45}{1.05} & \HDetailScore{19.02}{10.38} \\
 & FCT (s) & \HDetailScore{700.0}{0.0} & \HDetailScore{1700.0}{793.7} & \HDetailScore{3333.3}{1266.2} & \HDetailScore{3466.7}{1069.3} & \HDetailScore{5200.0}{692.8} & \HDetailScore{3266.7}{1078.6} & \HDetailScore{2944.4}{914.7} \\
 & Robot loss & \HDetailScore{1.60}{0.00} & \HDetailScore{5.10}{1.33} & \HDetailScore{6.99}{1.05} & \HDetailScore{7.30}{1.02} & \HDetailScore{14.37}{0.56} & \HDetailScore{12.86}{0.16} & \HDetailScore{8.04}{0.84} \\
 & Free loss & \HDetailReference{1.66} & \HDetailReference{6.36} & \HDetailReference{9.48} & \HDetailReference{11.66} & \HDetailReference{16.20} & \HDetailReference{15.21} & \HDetailReference{10.09} \\
\bottomrule
\end{tabular*}\par
\vspace{9pt}
\captionof{table}{T5 fire growth rate under local symbolic observations.}
\label{tab:appendix_h5_fire_t5_speed}
\noindent\begin{tabular*}{\linewidth}{@{}ll@{\extracolsep{\fill}}*{7}{r}@{}}
\toprule
\textbf{Level} & \textbf{Metric} & \makebox[51.4pt][c]{\textbf{S1R}} & \makebox[51.4pt][c]{\textbf{S2R}} & \makebox[51.4pt][c]{\textbf{S3R}} & \makebox[51.4pt][c]{\textbf{S4R}} & \makebox[51.4pt][c]{\textbf{S5R}} & \makebox[51.4pt][c]{\textbf{S6R}} & \makebox[51.4pt][c]{\textbf{Overall}} \\
\midrule
1 & SVR (\%) & \HDetailScore{8.42}{1.21} & \HDetailScore{17.90}{0.78} & \HDetailScore{22.46}{3.76} & \HDetailScore{37.00}{0.59} & \HDetailScore{27.43}{8.74} & \HDetailScore{14.38}{2.50} & \HDetailScore{21.27}{4.07} \\
 & HPR (\%) & \HDetailScore{13.00}{9.50} & \HDetailScore{15.19}{0.48} & \HDetailScore{70.51}{25.09} & \HDetailScore{32.58}{2.32} & \HDetailScore{80.65}{8.23} & \HDetailScore{56.73}{20.77} & \HDetailScore{44.78}{14.28} \\
\addlinespace[1pt]
1.25 & SVR (\%) & \HDetailScore{8.15}{0.23} & \HDetailScore{44.52}{5.90} & \HDetailScore{34.06}{4.43} & \HDetailScore{30.29}{6.19} & \HDetailScore{20.11}{11.44} & \HDetailScore{12.11}{4.16} & \HDetailScore{24.87}{6.34} \\
 & HPR (\%) & \HDetailScore{7.99}{0.37} & \HDetailScore{27.63}{9.61} & \HDetailScore{56.75}{27.67} & \HDetailScore{29.04}{1.34} & \HDetailScore{83.37}{4.46} & \HDetailScore{33.87}{9.86} & \HDetailScore{39.77}{12.76} \\
\addlinespace[1pt]
1.5 & SVR (\%) & \HDetailScore{8.19}{0.60} & \HDetailScore{27.15}{11.22} & \HDetailScore{22.10}{5.78} & \HDetailScore{31.26}{7.33} & \HDetailScore{30.84}{6.88} & \HDetailScore{13.38}{3.36} & \HDetailScore{22.15}{6.73} \\
 & HPR (\%) & \HDetailScore{9.28}{5.20} & \HDetailScore{22.95}{3.57} & \HDetailScore{36.95}{5.58} & \HDetailScore{20.63}{18.22} & \HDetailScore{72.74}{21.97} & \HDetailScore{10.89}{18.86} & \HDetailScore{28.91}{14.39} \\
\addlinespace[1pt]
1.75 & SVR (\%) & \HDetailScore{13.10}{2.26} & \HDetailScore{27.51}{23.91} & \HDetailScore{13.66}{4.09} & \HDetailScore{32.24}{9.75} & \HDetailScore{20.34}{2.39} & \HDetailScore{13.42}{7.31} & \HDetailScore{20.05}{11.16} \\
 & HPR (\%) & \HDetailScore{6.37}{0.96} & \HDetailScore{31.05}{11.35} & \HDetailScore{49.56}{21.00} & \HDetailScore{23.74}{11.42} & \HDetailScore{35.41}{43.63} & \HDetailScore{18.52}{17.20} & \HDetailScore{27.44}{21.99} \\
\addlinespace[1pt]
2 & SVR (\%) & \HDetailScore{6.99}{2.95} & \HDetailScore{2.69}{1.51} & \HDetailScore{18.11}{6.11} & \HDetailScore{13.68}{9.89} & \HDetailScore{15.75}{6.33} & \HDetailScore{9.19}{7.67} & \HDetailScore{11.07}{6.39} \\
 & HPR (\%) & \HDetailScore{6.54}{1.19} & \HDetailScore{17.74}{2.21} & \HDetailScore{59.79}{20.89} & \HDetailScore{12.01}{11.34} & \HDetailScore{38.55}{37.95} & \HDetailScore{28.86}{29.88} & \HDetailScore{27.25}{22.00} \\
\bottomrule
\end{tabular*}\par
\vspace{9pt}
\captionof{table}{T5 fire growth rate with RGB and RGB+DS observations.}
\label{tab:appendix_h5_fire_v_speed}
\noindent\begin{tabular*}{\linewidth}{@{}ll@{\extracolsep{\fill}}*{7}{r}@{}}
\toprule
\textbf{Level} & \textbf{Metric} & \multicolumn{6}{c}{\textbf{Scenario}} & {} \\
\midrule
\multicolumn{2}{@{}l}{\textbf{RGB}} & \makebox[51.4pt][c]{\textbf{S1R}} & \makebox[51.4pt][c]{\textbf{S2R}} & \makebox[51.4pt][c]{\textbf{S3R}} & \makebox[51.4pt][c]{\textbf{S4R}} & \makebox[51.4pt][c]{\textbf{S5R}} & \makebox[51.4pt][c]{\textbf{S6R}} & \makebox[51.4pt][c]{\textbf{Overall}} \\
\midrule
1 & SVR (\%) & \HDetailScore{2.52}{0.44} & \HDetailScore{9.98}{2.66} & \HDetailScore{12.57}{2.20} & \HDetailScore{13.92}{3.23} & \HDetailScore{12.12}{2.19} & \HDetailScore{3.73}{0.17} & \HDetailScore{9.14}{2.13} \\
 & FCT (s) & \HDetailScore{666.7}{57.7} & \HDetailScore{1633.3}{1443.4} & \HDetailScore{2600.0}{0.0} & \HDetailScore{3066.7}{208.2} & \HDetailScore{3100.0}{0.0} & \HDetailScore{2600.0}{173.2} & \HDetailScore{2277.8}{600.0} \\
 & Robot loss & \HDetailScore{1.19}{0.01} & \HDetailScore{1.93}{0.06} & \HDetailScore{2.66}{0.07} & \HDetailScore{3.91}{0.15} & \HDetailScore{4.05}{0.10} & \HDetailScore{7.25}{0.01} & \HDetailScore{3.50}{0.08} \\
 & Free loss & \HDetailReference{1.23} & \HDetailReference{2.15} & \HDetailReference{3.05} & \HDetailReference{4.54} & \HDetailReference{4.60} & \HDetailReference{7.53} & \HDetailReference{3.85} \\
\addlinespace[1pt]
1.5 & SVR (\%) & \HDetailScore{5.16}{1.42} & \HDetailScore{9.21}{3.17} & \HDetailScore{16.58}{12.80} & \HDetailScore{15.88}{11.33} & \HDetailScore{11.97}{1.37} & \HDetailScore{2.80}{0.59} & \HDetailScore{10.27}{7.15} \\
 & FCT (s) & \HDetailScore{600.0}{0.0} & \HDetailScore{2000.0}{0.0} & \HDetailScore{2966.7}{378.6} & \HDetailScore{2966.7}{1446.8} & \HDetailScore{2666.7}{57.7} & \HDetailScore{1966.7}{57.7} & \HDetailScore{2194.4}{611.5} \\
 & Robot loss & \HDetailScore{1.34}{0.02} & \HDetailScore{3.92}{0.14} & \HDetailScore{4.89}{0.75} & \HDetailScore{6.83}{0.92} & \HDetailScore{5.65}{0.09} & \HDetailScore{8.61}{0.05} & \HDetailScore{5.21}{0.49} \\
 & Free loss & \HDetailReference{1.41} & \HDetailReference{4.32} & \HDetailReference{5.86} & \HDetailReference{8.12} & \HDetailReference{6.42} & \HDetailReference{8.86} & \HDetailReference{5.83} \\
\addlinespace[1pt]
2 & SVR (\%) & \HDetailScore{1.83}{0.48} & \HDetailScore{0.29}{0.15} & \HDetailScore{3.79}{7.88} & \HDetailScore{1.27}{0.80} & \HDetailScore{5.78}{0.99} & \HDetailScore{1.20}{1.31} & \HDetailScore{2.36}{3.31} \\
 & FCT (s) & \HDetailScore{600.0}{0.0} & \HDetailScore{566.7}{57.7} & \HDetailScore{1800.0}{608.3} & \HDetailScore{1400.0}{400.0} & \HDetailScore{1666.7}{57.7} & \HDetailScore{1500.0}{0.0} & \HDetailScore{1255.6}{299.1} \\
 & Robot loss & \HDetailScore{1.55}{0.01} & \HDetailScore{4.50}{0.01} & \HDetailScore{6.43}{0.53} & \HDetailScore{9.15}{0.07} & \HDetailScore{6.66}{0.07} & \HDetailScore{9.08}{0.12} & \HDetailScore{6.23}{0.22} \\
 & Free loss & \HDetailReference{1.58} & \HDetailReference{4.51} & \HDetailReference{6.68} & \HDetailReference{9.26} & \HDetailReference{7.07} & \HDetailReference{9.19} & \HDetailReference{6.38} \\
\midrule
\multicolumn{2}{@{}l}{\textbf{RGB+DS}} & \makebox[51.4pt][c]{\textbf{S1R}} & \makebox[51.4pt][c]{\textbf{S2R}} & \makebox[51.4pt][c]{\textbf{S3R}} & \makebox[51.4pt][c]{\textbf{S4R}} & \makebox[51.4pt][c]{\textbf{S5R}} & \makebox[51.4pt][c]{\textbf{S6R}} & \makebox[51.4pt][c]{\textbf{Overall}} \\
\midrule
1 & SVR (\%) & \HDetailScore{10.13}{3.99} & \HDetailScore{12.87}{1.70} & \HDetailScore{18.72}{14.88} & \HDetailScore{22.61}{15.71} & \HDetailScore{18.08}{6.65} & \HDetailScore{7.04}{0.62} & \HDetailScore{14.91}{9.41} \\
 & FCT (s) & \HDetailScore{342.2}{136.8} & \HDetailScore{800.0}{0.0} & \HDetailScore{1400.0}{519.6} & \HDetailScore{2233.3}{757.2} & \HDetailScore{2833.3}{152.8} & \HDetailScore{1833.3}{152.8} & \HDetailScore{1573.7}{389.2} \\
 & Robot loss & \HDetailScore{1.10}{0.05} & \HDetailScore{1.87}{0.04} & \HDetailScore{2.48}{0.45} & \HDetailScore{3.52}{0.71} & \HDetailScore{3.77}{0.31} & \HDetailScore{7.00}{0.05} & \HDetailScore{3.29}{0.37} \\
 & Free loss & \HDetailReference{1.23} & \HDetailReference{2.15} & \HDetailReference{3.05} & \HDetailReference{4.54} & \HDetailReference{4.60} & \HDetailReference{7.53} & \HDetailReference{3.85} \\
\addlinespace[1pt]
1.5 & SVR (\%) & \HDetailScore{4.65}{1.18} & \HDetailScore{6.84}{2.58} & \HDetailScore{16.71}{3.98} & \HDetailScore{47.46}{3.20} & \HDetailScore{13.06}{1.75} & \HDetailScore{11.34}{7.50} & \HDetailScore{16.68}{3.94} \\
 & FCT (s) & \HDetailScore{600.0}{0.0} & \HDetailScore{1900.0}{100.0} & \HDetailScore{2300.0}{100.0} & \HDetailScore{1466.7}{230.9} & \HDetailScore{2566.7}{57.7} & \HDetailScore{1333.3}{493.3} & \HDetailScore{1694.4}{230.9} \\
 & Robot loss & \HDetailScore{1.35}{0.02} & \HDetailScore{4.03}{0.11} & \HDetailScore{4.88}{0.23} & \HDetailScore{4.27}{0.26} & \HDetailScore{5.58}{0.11} & \HDetailScore{7.85}{0.66} & \HDetailScore{4.66}{0.31} \\
 & Free loss & \HDetailReference{1.41} & \HDetailReference{4.32} & \HDetailReference{5.86} & \HDetailReference{8.12} & \HDetailReference{6.42} & \HDetailReference{8.86} & \HDetailReference{5.83} \\
\addlinespace[1pt]
2 & SVR (\%) & \HDetailScore{3.69}{1.52} & \HDetailScore{18.58}{31.56} & \HDetailScore{22.98}{14.78} & \HDetailScore{48.11}{0.91} & \HDetailScore{5.41}{1.78} & \HDetailScore{13.11}{11.52} & \HDetailScore{18.65}{15.02} \\
 & FCT (s) & \HDetailScore{500.0}{100.0} & \HDetailScore{461.7}{160.9} & \HDetailScore{1333.3}{351.2} & \HDetailScore{833.3}{57.7} & \HDetailScore{1400.0}{264.6} & \HDetailScore{1666.7}{57.7} & \HDetailScore{1032.5}{198.3} \\
 & Robot loss & \HDetailScore{1.52}{0.02} & \HDetailScore{3.67}{1.42} & \HDetailScore{5.15}{0.99} & \HDetailScore{4.81}{0.08} & \HDetailScore{6.69}{0.13} & \HDetailScore{7.99}{1.06} & \HDetailScore{4.97}{0.83} \\
 & Free loss & \HDetailReference{1.58} & \HDetailReference{4.51} & \HDetailReference{6.68} & \HDetailReference{9.26} & \HDetailReference{7.07} & \HDetailReference{9.19} & \HDetailReference{6.38} \\
\bottomrule
\end{tabular*}\par
\endgroup

\end{document}